%% file: iclr2024_conference.tex
\documentclass[table]{article} %
\usepackage{iclr2024_conference,times}

\usepackage[utf8]{inputenc} %
\usepackage[T1]{fontenc}    %
\usepackage{hyperref}       %
\usepackage{url}            %
\usepackage{booktabs}       %
\usepackage{amsfonts}       %
\usepackage{nicefrac}       %
\usepackage{microtype}      %
\usepackage{xcolor}         %
\usepackage{mdframed}
\usepackage{amsmath}
\usepackage{graphicx}
\usepackage{xspace}
\usepackage{xparse}
\usepackage{dashrule}
\usepackage{wrapfig}
\usepackage{subcaption}
\usepackage{tikz}
\usepackage{enumitem}
\usepackage{multirow}
\usepackage{siunitx}

\usepackage{algorithm}
\usepackage{algpseudocode}

\usepackage{mathtools}
\usepackage{amssymb}%
\usepackage{pifont}%
\usepackage{makecell}
\usepackage{cancel}
\usepackage{authblk}

\usepackage{tikz}
\usetikzlibrary{patterns}
\usepackage{tablefootnote}

\usepackage{tcolorbox}
\definecolor{wkblue}{RGB}{179,229,226}
\definecolor{meta-color}{RGB}{16,177,168}
\usepackage{listings}

\usepackage{minitoc} %

\noptcrule

\definecolor{battleshipgrey}{rgb}{0.3, 0.3, 0.3}
\definecolor{brilliantrose}{rgb}{1.0, 0.33, 0.64}
\definecolor{americanrose}{rgb}{1.0, 0.01, 0.24}
\definecolor{jweigreen}{rgb}{0,0.45,0.24}
\definecolor{bluegray}{rgb}{0.1, 0.1, 0.4}
\definecolor{ao(english)}{rgb}{0.0, 0.5, 0.0}
\definecolor{blanchedalmond}{rgb}{1.0, 0.92, 0.8}
\definecolor{atomictangerine}{rgb}{1.0, 0.6, 0.4}
\definecolor{chocolate(web)}{rgb}{0.82, 0.41, 0.12}
\definecolor{bananayellow}{rgb}{1.0, 0.88, 0.21}
\definecolor{goldenbrown}{rgb}{0.6, 0.4, 0.08}
\definecolor{aliceblue}{rgb}{0.94, 0.97, 1.0}
\definecolor{beige}{rgb}{0.96, 0.96, 0.86}
\definecolor{babyblue}{rgb}{0.54, 0.81, 0.94}
\definecolor{camel}{rgb}{0.76, 0.6, 0.42}
\definecolor{cinnamon}{rgb}{0.82, 0.41, 0.12}
\definecolor{titlecolor}{HTML}{4c9cff}

\definecolor{mydarkblue}{rgb}{0,0.08,0.45}
\definecolor{citeblue}{HTML}{3b86d9} %
\hypersetup{
    colorlinks=true,
    linkcolor=mydarkblue,
    filecolor=blue,      
    urlcolor=mydarkblue,
    citecolor=citeblue,
}

\definecolor{proxGreen}{HTML}{097a54}
\definecolor{proxDarkBlue}{HTML}{012e59}
\definecolor{proxBlue}{HTML}{265ed4}
\definecolor{proxLightBlue}{HTML}{012e59}
\definecolor{proxTeal}{HTML}{00d5ff}
\definecolor{proxYellow}{HTML}{ffbb00}
\definecolor{proxOrange}{HTML}{e68e1a}
\definecolor{proxPink}{HTML}{f0539b}
\definecolor{proxYellow}{HTML}{ff9100}
\definecolor{proxDarkYellow}{HTML}{fdac15}
\definecolor{proxBlue}{HTML}{2E3168}
\definecolor{proxLightBlue}{HTML}{2a88ef}
\colorlet{lightProxPink}{proxPink!50}
\colorlet{lightProxYellow}{proxYellow!50}
\colorlet{lightProxLightBlue}{proxLightBlue!50}
\colorlet{lightProxGreen}{proxGreen!50}
\definecolor{cellHighlight}{HTML}{dbefff}
\definecolor{removered}{HTML}{FFB3BA}

\definecolor{quoteborder}{RGB}{214,227,240} %
\definecolor{quotebg}{RGB}{236,243,250} 
\newmdenv[
  backgroundcolor=quotebg,   %
  linecolor=quoteborder,
  skipabove=1em,
  skipbelow=0em,
  leftline=true,
  topline=false,
  bottomline=false,
  rightline=false,
  linecolor=blue!66,        %
  linewidth=4pt
]{githubquote}

\newcommand{\tlmxs}{\textsc{TLM-xs}\xspace}
\newcommand{\tlms}{\textsc{TLM-s}\xspace}
\newcommand{\tlmm}{\textsc{TLM-m}\xspace}

\newcommand{\tinylm}{\textsc{TinyLlama-1.1B}\xspace}

\newcommand{\llemma}{\textsc{Llemma}\xspace}
\newcommand{\tinyllama}{\textsc{TinyLlama}\xspace}
\newcommand{\codellama}{\textsc{CodeLlama}\xspace}
\newcommand{\llamaii}{\textsc{Llama-2}\xspace}
\newcommand{\llamaiii}{\textsc{Llama-3-70B-Instruct}\xspace}
\newcommand{\mistral}{\textsc{Mistral}\xspace}
\newcommand{\intern}{\textsc{InternLM2-Base}\xspace}
\newcommand{\internmath}{\textsc{InternLM2-Math}\xspace}
\newcommand{\pythia}{\textsc{Pythia}\xspace}

\newcommand{\mixtral}{\textsc{Mistral}\xspace}

\newcommand{\owm}{\text{OpenWebMath}\xspace}
\newcommand{\redpj}{\text{RedPajama-V2}\xspace}
\newcommand{\fineweb}{\text{FineWeb}\xspace}
\newcommand{\gopher}{\text{Gopher}\xspace}

\newcommand{\huggingface}{\raisebox{-1.5pt}{\includegraphics[height=1.05em]{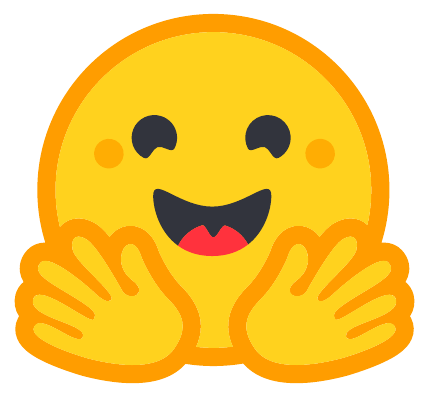}}\xspace}
\newcommand{\github}{\raisebox{-1.5pt}{\includegraphics[height=1.05em]{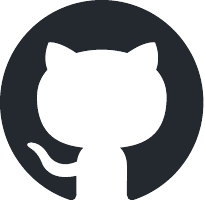}}\xspace}

\newcommand{\method}{\textsc{ProX}\xspace}
\newcommand{\m}[1]{$#1$}

\newcommand{\HL}{\cellcolor{cellHighlight}}

\definecolor{authorGreen}{HTML}{59C4B8}
\definecolor{authorBlue}{HTML}{4C9CFF}
\definecolor{authorPurple}{HTML}{AF73FA}
\newcommand{\bsjtu}{{\color{authorBlue}\boldsymbol{\alpha}}}
\newcommand{\bsea}{{\color{authorGreen}\boldsymbol{s}}}
\newcommand{\bailab}{{\color{authorPurple}\boldsymbol{\delta}}}
\newcommand{\bgair}{{\color{proxYellow}\boldsymbol{\mu}}}

\title{\textcolor{titlecolor}{Pro}gramming Every E\textcolor{titlecolor}{x}ample: Lifting Pre-training Data Quality Like Experts at Scale}
\title{\textcolor{titlecolor}{\textbf{Pro}}gramming Every E\textcolor{titlecolor}{\textbf{x}}ample: Lifting Pre-training Data Quality Like Experts at Scale}

\input{author}

\iclrfinalcopy
\begin{document}
\doparttoc
\faketableofcontents
\maketitle
\vspace{-5mm}
\begin{abstract}
Large language model pre-training has traditionally relied on human experts to craft heuristics for improving the corpora quality, resulting in numerous rules developed to date.
However, these rules lack the flexibility to address the unique characteristics of individual example effectively.
Meanwhile, applying tailored rules to every example is impractical for human experts.
In this paper, we demonstrate that even small language models, with as few as 0.3B parameters, can exhibit substantial
data refining capabilities comparable to those of human experts.
We introduce Programming Every Example~(\method), a novel framework that treats data refinement as a \emph{programming task}, enabling models to refine corpora by generating and executing fine-grained operations, such as string normalization, for each individual example at scale.
Experimental results show that models pre-trained on \method-curated data outperform either original data or data filtered by other selection methods by more than \m{2\%} across various downstream benchmarks.
Its effectiveness spans various model sizes and pre-training corpora, including C4, \redpj, \fineweb, \fineweb-Edu, and DCLM.
Furthermore, \method exhibits significant potential in domain-specific continual pre-training: without 
domain specific design, models trained on \owm refined by \method outperform human-crafted rule-based methods, improving average accuracy by \m{\mathbf{7.6\%}} over \mistral-7B, with \m{\mathbf{14.6\%}} for \llamaii-7B and \m{\mathbf{20.3\%}} for \codellama-7B, all within \m{\mathbf{10}}B tokens to be comparable to models like \llemma{}-7B trained on \m{\mathbf{200}}B tokens.
Further analysis highlights that \method significantly saves training FLOPs, offering a promising path for efficient LLM pre-training.
We are open-sourcing \method with \m{\mathbf{\ge500}}B corpus, models, and sharing all training and implementation details for reproducible research and future innovation.
\begin{itemize}
    \item \small{\huggingface \textbf{HF Repo}: \url{https://huggingface.co/gair-prox}}
    \item \small{\github \textbf{Code}: \url{https://github.com/GAIR-NLP/ProX}}
\end{itemize}
\end{abstract}
\vspace{-4mm}

\input{figures/intro_fig}

\newpage

\section{Introduction}
\vspace{5.0mm}

Large Language Models (LLMs) have made significant strides in capabilities~\citep{metallama3,achiam2023gpt,anthropic2024claude,reid2024gemini}, excelling in tasks such as creative writing~\citep{yuan2022wordcraft}, complex reasoning~\citep{wei2022chain,kojima2022large}, and agentic task planning and execution~\citep{fan2022minedojo,park2023generative}. Behind these, massive, high-quality pre-training corpora form the backbone of these models, equipping them with the essential knowledge and reasoning abilities crucial for a wide range of downstream tasks~\citep{together2023redpajama,penedo2024fineweb}.

The Internet offers vast amounts of data, but much of it is noisy and unrefined, requiring extensive cleaning and quality enhancement before being applied for pre-training. 
Previous works focus primarily on designing heuristic-based pipelines to lift data quality, such as 
document filtering~\citep{rae2021gopher, penedo2024fineweb, soldaini-etal-2024-dolma} and 
perplexity-based scoring methods~\citep{together2023redpajama}, relying heavily on human expertise and manual adjustments~\citep{zhang2024map}. While widely adopted, these labor-intensive solutions are inherently limited by rule coverage and their inability to address every specific case. Recently, some efforts have explored leveraging LLMs for high-quality data acquisition. On the one hand, language models have been applied for data filtering or selection~\citep{xie2023data,wettig2024qurating,yu2024mates,dubey2024llama3}, but their role is largely limited to identifying low-quality documents without enabling fine-grained refinements (e.g., string-level). On the other hand, LLMs are also being used directly generating high-quality data, \emph{i.e.}, data synthesis~\citep{gunasekar2023textbooks-phi1,li2023textbooks-phi1.5,benallal2024cosmopedia}. Unlike filtering, synthesis methods actively create or refine data to produce new documents, but they require substantial computational resources, limiting scalability. Despite their success, these methods can also inherit issues like hallucination~\citep{maini2024rephrasing}, and assessing their correctness and completeness in an interpretable manner remains a challenge~\citep{liu2024best}.

Standing at the intersection of data processing efficiency and data quality improvement, in this work, we propose \method, a model-based framework for pre-training level data refinement.
\method focuses on refining large-scale data with relatively smaller models, offering a more efficient alternative.
As shown in Figure~\ref{fig:tlm-intro}, in practice, \method first adapts 
a small base language model (less than $1$B) 
to data refining tasks via fine-tuning on seed data.
This \method's refining model then determines the appropriate operations for each example in the pre-training corpora
through versatile programs, including operations such as filtering, string normalization and noisy line removal.
Finally, the generated program is executed by a pre-defined 
executor, producing refined corpus ready for pre-training.
In this way, \method is empowered with language models to autonomously refine pre-training corpora, leveraging flexible function calls to enhance data quality.

Experimental results demonstrate that the proposed \method framework consistently lifts data quality for \textbf{pre-training}.
Specifically, \method achieves an average improvement of $2.1\%$ over $10$ downstream benchmarks and outperforms state-of-the-art data selection methods by over $2.0\%$.
Furthermore, \method shows broad applicability across model sizes from \m{0.3}B to \m{1.7}B and shows consistent performance gains across diverse pre-training corpora of varying quality, including \redpj~\citep{together2023redpajama}, C4~\citep{raffel2020exploring}, \fineweb,  \fineweb-Edu~\citep{penedo2024fineweb}, and DCLM~\citep{li2024datacomp}.
In domain-specific \textbf{continual pre-training}, \method yields an $11\%$ gain over \owm~\citep{paster2023openwebmath} for \tinylm and $7.6\%$ for \mistral-7B across \m{9} mathematical tasks, with similar improvements seen on \llamaii-7B and \codellama-7B. 
Beyond performance gains, results also suggest that pre-training on the refined corpus significantly boosts pre-training efficiency, achieving similar downstream performance with up to \m{\mathbf{20}\times} less computing.
We believe it is worthwhile to \textbf{scale up computing FLOPs for data refinement}, which enables similar performance with much less training cost and offers a promising path for efficient LLM pre-training.

\section{Approach: Programming Every Example}
\subsection{Data Refinement Task Formulation}
\label{subsec:task-formulation}

\begin{figure}[!t]
    \centering
    \includegraphics[width=0.92\textwidth]{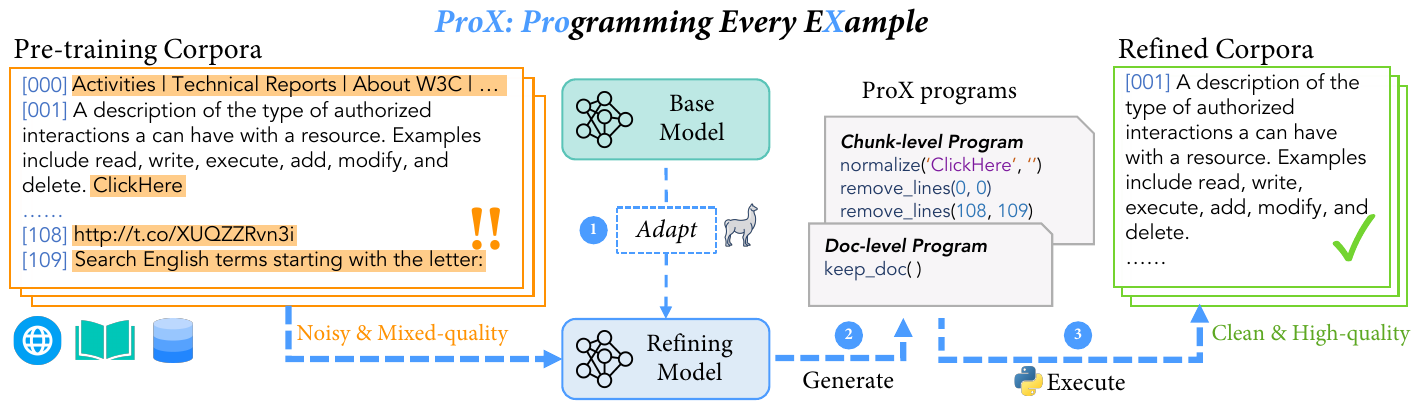}
    \vspace{-2.5mm}
    \caption{An overview of \method framework: (1) we adapt a base language model to perform data refinement; (2)~\method refining models are able to generate complex programs for each document, including document level filtering and more fine-grained chunk level refining; (3) A \includegraphics[height=0.8em]{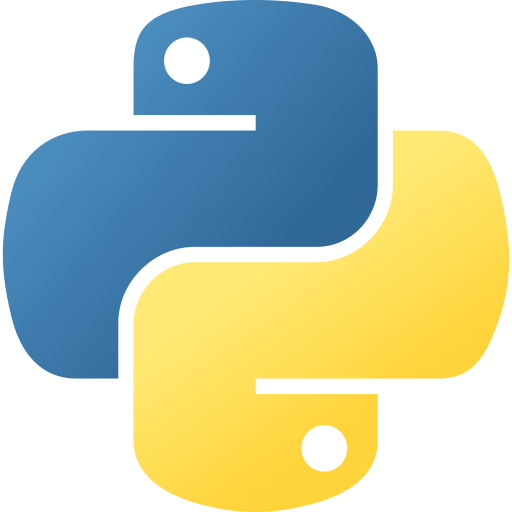}~Python executor will execute the programs with the docs, producing the refined high-quality corpora. 
    }
    \vspace{-3.5mm}
    \label{fig:tlm-intro}
\end{figure}

Given any document in the corpus $d\in\mathcal{D}$, such as an HTML extract or a textbook, we define data refinement as the process of transforming $d$ into $\hat{d}$, where $\hat{d}$ exhibits higher quality.
While it is challenging to formally define ``higher quality'' for pre-training data, we assume it can be described through qualitative improvements, such as the removal of advertisements, meaningless URL links, random code gibberish, and content lacking educational value, just as shown on the left side of Figure~\ref{fig:tlm-intro}.
Specifically, we formulate this refining process as the generation of a data processing program $\mathcal{Z}$, conditioned on $d$.
The refined document $\hat{d}$ is then produced by executing program $\mathcal{Z}$ on the original document $d$.
For instance, the ``string normalization'' can be a very fine-grained process transforming noisy strings into clean ones with executor \m{\mathcal{E}} and program \m{\mathcal{Z}_{\text{normalize}}} :  
\begin{equation}
\mathcal{E}(\mathcal{Z}_{\text{normalize}}, d) = (s'_i)_{i=1}^{|d|}, \text{ where } s'_i = 
\text{normalize}(s_i) \text{ if } s_i \text{ needs normalization else } s_i
\end{equation}
Here, 
$d = (s_1, s_2, ..., s_{|d|})$ is the original document represented as a sequence of strings, and
{\texttt{normalize()}}
is our normalization function that maps certain strings to their normalized versions.
Moreover, the document filtering process can be regarded as a special case of such refining transformation where executing on $\mathcal{Z}_{\text{filter}}$ will lead to removing the whole document, \emph{i.e.}, $\mathcal{E}(\mathcal{Z}_{\text{filter}}, d) = \varnothing$.

In this manner, data quality improvement operations, such as data cleaning or normalizing, can be unified into the standardized function that applies a specific transformation or cleaning process to the document.
These operations can be represented as various instantiations of the general executor $\mathcal{E}(\mathcal{Z}, d)$, where $\mathcal{Z}$ encodes the function calling snippets or heuristics for the specific task.

\subsection{\method Framework}
\paragraph{Overview}
As shown in Figure~\ref{fig:tlm-intro}, given any document $d$ as input, the \method framework utilizes the language model itself with parameter $\theta$ to generate the data refinement program $\mathcal{Z}=f(\theta, d)$.
The snippet is executed within the executor
$\mathcal{E}$, producing the refined document $\Hat{d} = \mathcal{E}(f(\theta, d), d)$.
We include two stages in the \method framework, aiming to refine the data progressively, from rough to fine-grained.
These two stages are referred to as 
\textit{{document-level programming}} and \textit{{chunk-level programming}}, as illustrated in Figure~\ref{fig:tlm-intro}.
In each stage, the \method refining model will generate programs $\mathcal{Z}_\text{doc}$ and $\mathcal{Z}_\text{chunk}$ that refine the corpora at varying levels of granularities.

\paragraph{\method Program Design}

The detailed program space design is also crucial for maximizing the capabilities of language models.
We believed designing such model-based operations should consider several realistic factors when scaling to large pre-training corpora:
(1) the model does not need to be very powerful or very large to handle these tasks, it only needs to recognize several patterns; 
(2) the solution, though requiring more computing budget compared to heuristic-rule-based pipelines, still needs to be simple and efficient.
Under such consideration, we simply let the language models generate function calls without detailed implementations.
These design choices aim to balance functionality with the limitations of small language models, enabling effective document manipulation while maintaining simplicity and coherence.

\input{tables/program-space}
The most fundamental operations we aim to perform on a document, are deletion and replacement.
We incorporate these types of operations across different programming stages aiming to refine the corpus with different granularities in \method:
(1) In the document-level programming stage, we simply define the function~\texttt{drop\_doc()}to delete a document and~\texttt{keep\_doc()}to retain it.
(2) In chunk-level programming, we split the lengthy documents into smaller chunks 
and apply fine-grained operations to these chunks.
These operations include deleting specific lines \texttt{remove\_lines()} and replacing strings \texttt{normalize()}, providing flexibility in modifying content rather than simply dropping the whole document.
Also for high-quality chunks that do not require any modifications, we use the \texttt{keep\_chunk()} function for flagging.  
We present the detailed function definition in Table~\ref{tab:program-space}, which is also the generation space of \method's refining models.
While the individual functions may seem straightforward, their design space is flexible and capable of expressing complex rules previously developed by human experts as shown in Table~\ref{tab:program-space}.
In fact, these rules can be projected into the program space of \method, showcasing that our approach not only simplifies but also enhances the rule-creation process, offering more systematic and scalable refinement capabilities.
\paragraph{\method Execution}
During the execution stage, the generated program snippets $\mathcal{Z}$ will be executed by the executor $\mathcal{E}$ to refine the document.
For simplicity and flexibility, \method integrates Pythonic grammars, wrapping all operations into different function calling with parameters and implements these function in Python for later execution.
For example, in Figure~\ref{fig:tlm-intro}, the document contains some noisy patterns including navigation bars, meaningless HTML links and page indexes.
The refining model will then generate programs to remove the corresponding lines and patterns.
In the document-level and chunk-level cleaning stage, \method utilizes an independent refining model to generate programs with various function calls described in Table~\ref{tab:program-space}.
We believe this sequential approach ensures a structured and effective refinement, addressing the larger document noise first, and then focusing on finer-grained cleaning.

\subsection{Model Adaptation for \method}

\begin{figure}[h]
    \centering
    \includegraphics[width=0.92\textwidth]{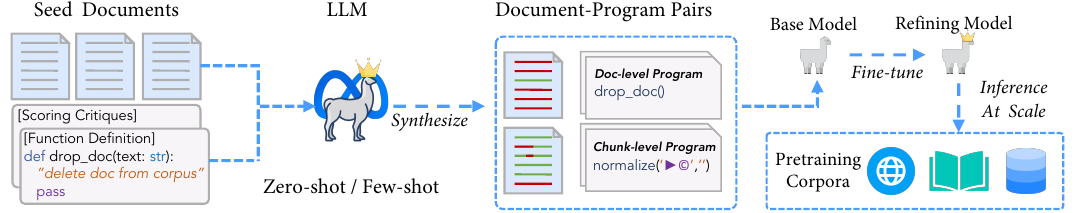}
    \caption{
    The illustration of the model adaptation in \method. We employ powerful LLMs~(\textsc{Llama-3}) to annotate random seed documents with valid programs, and use this \textit{doc-program} pairs to fine-tune a small base model, obtaining the refining model suitable for fine-grained data refining tasks. 
    }
    \label{fig:tlm-overview}
\end{figure}

It is generally difficult for base models to directly generate \method programs.
In fact, even for the most powerful post-trained LLMs, generating custom API calls is relatively challenging at the current stage~\citep{zhuo2024bigcodebench}.
Thus, it will be necessary that we curate some seed data to adapt the model for these scenarios.
Under such consideration, we employ strong LLMs to annotate these operations via zero-shot and few-shot prompting, and then adapt our base model to these tasks by supervised fine-tuning~(SFT).
We first use two additive scale scoring prompts~\citep{yuan2024self,penedo2024fineweb} to split the corpus into kept documents and dropped documents.
And then we use large models to annotate fine-grained programs based on kept documents.
Specifically, we leverage the \textsc{Llama-3} series of models~\citep{dubey2024llama3} for data collection and annotation.
In \method, this data collection is performed only once, and all base models are adapted with the same curated data.
To ensure the reliability of the collected data, we also conduct necessary checks for grammar correctness and control the removal ratio threshold. 
The detailed procedure for program synthesis and post-processing can be found in \S~\ref{app:subsec-seed-data-collection}.

For simplicity, we directly use a small language model (\emph{e.g.}, $0.3$B parameters) that we have trained on approximately $26$B tokens of original unrefined data as the base model, which also serves as the comparison baseline in subsequent experiments.
The adapted model's performance is then evaluated using the F1 score on the split validation dataset, ensuring a robust assessment.
We select the highest-performing model checkpoint and employ the model to generate programs $\mathcal{Z}$, for each document or chunk of the dataset. These programs together with the documents are then executed using the corresponding function implementation, resulting in the final processed corpus. 
Please see appendix for more training details~(\S~\ref{app-subsec:sft-training-details}), implementation for calculating the F1 score~(\S~\ref{app-subsec:sft-eval-metrics}), and large scale inference~(\S~\ref{app:subsec-inference-at-scale}).

\section{Experiments}\label{sec:exp}
In this section, we first describe our experimental setup, then verify the effectiveness of each \method stage and compare it with existing data selection methods tailored for pretraining corpus (\S~\ref{subsec:exp1-verify-walle}).
We then apply \method 
to various model sizes and corpora to demonstrate its broad applicability (\S~\ref{subsec:exp2-beyond-size-and-corpora}).
Finally, we apply \method to the mathematical domain, demonstrating its superiority and universality in domain-specific training (\S~\ref{subsec:exp3-math-cpt}).

\subsection{Experiment Setup}

\paragraph{Training Corpora}
We utilize various corpora for both general and specific domain data in our experiments. For general domain data, we begin with \redpj~\citep{together2023redpajama}, a preprocessed large-scale dataset of $30$ trillion tokens from diverse Internet sources, ready for pre-training. We further apply \method on the C4 corpus~\citep{raffel2020exploring} with $198$ billion tokens, the \fineweb dataset~\citep{penedo2024fineweb}(as well as \fineweb-Edu) containing $15$ trillion tokens, noted for high data quality, and DCLM-baseline-1.0~\citep{li2024datacomp}. For specific domain experiments, we use \owm~\citep{paster2023openwebmath}, a math-focused dataset with $15$ billion tokens.
Given the limitations in computational resources, we conduct experiments on a randomly sampled subset of the entire pre-training dataset.
See Table~\ref{app:tab-pt-corpora}~(\S~\ref{app:sub-sec-pt-corpora}) for sampling details.
\paragraph{Base Model Selection}
Our pre-training experiments are conducted using various sizes of decoder-only language models. Detailed specifications of these models and all training recipes are provided in \S~\ref{app:subsec-model-and-train-config}, especially in Table~\ref{app-tab:tlm-architecture} and Table~\ref{tab:app-training-params}.
\begin{enumerate}[leftmargin=1.25em,itemindent=0.25em,labelsep=0.4em,itemsep=0.15em]
\item To verify different stages' effectiveness of \method{}, we employ a $750$M sized model sharing \llamaii{} architecture~\citep{touvron2023llama}, denoted as \tlms{}, used for both pre-training from scratch and refining.
We also compare \method{} with data selection methods using \pythia-410M{}/1B's architecture~\citep{biderman2023pythia}, as those employed in MATES~\citep{yu2024mates}.
\item For further evaluation of \method{} using different refining and base model sizes, we scale the model sizes from $350$M~($0.5\times$ smaller, denoted as \tlmxs{}) and $1.7$B~($2\times$ larger, denoted as \tlmm), all based on the \llamaii{} architecture.
\item For domain-specific continual pre-training, we select \tinylm{}~\citep{zhang2024tinyllama}, \llamaii~\citep{touvron2023llama}, \codellama~\citep{DBLP:journals/corr/abs-2308-12950-codellama} and \mistral-7B{}~\citep{jiang2023mistral} as representative base models for their adequate training and solid performance.
\end{enumerate}

\paragraph{Baselines}
To ensure a fair comparison w.r.t. training cost, we keep all training hyperparameters, such as training steps and batch size, consistent across baselines, with only the data refining and selection pipelines differing.
We compare \method to a series of baselines:
\begin{enumerate}[leftmargin=1.25em,itemindent=0.25em,labelsep=0.4em,itemsep=0.15em]
\item In \S~\ref{subsec:exp1-verify-walle}, to verify \method's effectiveness, we first compare with \method with regular pre-training over the raw \redpj data.
We also introduce heuristic baselines used to curate the \fineweb corpora, which is the combination of three filtering strategies from C4~\citep{raffel2020exploring}, \gopher~\citep{rae2021gopher}, and newly crafted rules~(as \fineweb rules). 
Apart from rule-based baselines,
we also introduce existing data selection techniques proposed in previous works, including 
(1) importance resampling: DSIR~\citep{xie2023data};
(2) model-based selection:~DsDm~\citep{engstrom2024dsdm}, MATES~\citep{yu2024mates}, and QuRating~\citep{wettig2024qurating}.

\item In \S~\ref{subsec:exp2-beyond-size-and-corpora}, to test \method on different model sizes and training corpora, we finally scale the \tlmm's training tokens to $50$B over \redpj, C4, \fineweb (as well as \fineweb-Edu) and DCLM-baseline-1.0.
To show \method efficiency, we then directly compare with models covering a variety of pre-training approaches including (1) large-scale pre-training: \tinylm~\citep{zhang2024tinyllama} trained on $3$T tokens;
(2) model pruning from existing models:~(\textsc{SheadLlama}~\citep{xiasheared} pruned from \llamaii and trained on extra 50B tokens);
(3) LLM synthesis~(\textsc{InstructionLM}-1.3B~\citep{cheng2024instruction} trained on \mistral-7B generated data and \textsc{cosmo}-1.8B~\citep{benallal2024cosmopedia} trained on \mixtral-8x7B generated data).
\item In \S~\ref{subsec:exp3-math-cpt}'s specific domain continual pre-training, apart from standard continual pre-training on \tinylm, \textsc{Llama-2}-7B, \textsc{CodeLlama}-7B, and \mistral-7B, we additionally introduce with well-known and strong baselines trained on public~(or partially public) data, including \textsc{Rho}-1~\citep{lin2024rho}, \internmath~\citep{ying2024internlm-math}, \llemma~\citep{azerbayev2024llemma}, and an internal checkpoint reported in \textsc{DeepSeek-Math}~\citep{shao2024deepseekmath}. 
\end{enumerate}
\paragraph{Evaluation Setup}
We compare the base models' performance over a vast of datasets  for comprehensive evaluation:
(1) For general pre-training, we evaluate the performance across ten selected tasks using lighteval's implementation~\citep{lighteval}, and report the zero-shot
accuracy; we have also included LM-eval-harness~\citep{biderman2024lessons} for fair comparison with data selection methods.
(2) For domain-specific continual pre-training evaluation, \emph{i.e.}, mathematical related benchmarks, we use the same nine implementation and benchmarks used in \textsc{Rho}-1~\citep{lin2024rho} and evaluate all the base models with few-shot chain-of-thought (CoT) examples~\citep{wei2022chain}. 
The selected evaluation benchmarks, number of evaluation examples, and full details can be found in \S~\ref{app:sec-eval-details}.

\subsection{Verifying \method's effectiveness}
\label{subsec:exp1-verify-walle}
\paragraph{Verifying Effectiveness for Each \method Operation}
We first conduct a series of experiments to verify the effectiveness of each \method operation. We begin by training \tlms on the \redpj raw data for approximately $26$B tokens (or $12.5$K steps) as the initial baseline.
Following~\citet{wettig2024qurating} and for convenience, we then sequentially apply the doc-level and chunk-level refining pipelines by fine-tuning the \m{0.7}B model itself.
We then perform large-scale program synthesis and execution using the refining models, resulting in $\mathcal{D}_{\text{Doc}}$ and $\mathcal{D}_{\text{Doc+Chunk}}$.
Such $2$-stage synthesis requires approximately $192$ A100-80G GPU hours for processing
$60$B tokens of data.
The resulting zero-shot downstream performance is presented in Table~\ref{tab:exp1-zero-shot-avg-performance}, including base models trained on the data produced by \method refinement methods and different rule-based filtering methods.
Moreover, we visualize the dynamic benchmark performance in Figure~\ref{fig:exp1-bench}, implying the consistent improvement of \method over all baselines. 
See \S~\ref{subsec:full-lighteval-performance} for full detailed results of all intermediate checkpoints.

These results show that \method is highly effective, outperforming the raw corpus with an average boost of $2.5\%$, including significant improvements such as $7.6\%$ on ARC-E, $3.3\%$ on HellaSwag, and $2.1\%$ on MMLU.
We believe such consistent performance is significant given that these improvements were achieved even on benchmarks that are typically prone to performance instability, such as SIQA, WinoGrande, and CSQA.
By contrast, rule-based methods demonstrate relatively marginal overall improvement.
For instance, \gopher rules achieve only a $0.2\%$ boost, while C4 shows a modest $0.5\%$ improvement.
Furthermore, combining all three rules~(as is done in constructing the official \fineweb corpus), does not lead to any larger enhancement in overall performance.

\input{tables/exp-1}
\input{figures/exp1-curve}

\paragraph{Comparing with Data Selection Methods}
Apart from comparing with heuristic methods, we also include 
existing representative model-based data selection methods tailored for pertaining corpus to verify \method's effectiveness in Table~\ref{tab:exp1-data-selection-baseline}, where we report both \m{0}-shot and \m{2}-shot performance under the same settings used in MATES~\citep{yu2024mates}.
While we merely apply document-level stage~(\emph{i.e.}, \method-D) which is indeed similar to data selection methods, we can see that \method outperforms the strongest data selection method MATES, by $2.2\%$ and $2.5\%$ in \m{0}-shot and \m{2}-shot average performance for $410$M model, and by $1.0\%$ and $2.0\%$ for $1$B model.
Additionally, \method achieves the best performance on $7$ out of $8$ benchmarks tested, demonstrating its superiority over existing data selection methods. Full evaluation results are provided in Table~\ref{tab:exp-1-full-results-mates}~(\S~\ref{app:subsec-full-lmeval-performance}).

\subsection{Applying \method across model sizes and pretraining corpora}
\label{subsec:exp2-beyond-size-and-corpora}
\vspace{2.0mm}

In this section, we demonstrate that \method can effectively benefit models beyond scale and across different corpora, showing potential for iterative pre-training improvements.

\paragraph{\method works well across different scales.}
We train a family of models from $350$M to $1.7$B~(\emph{i.e.}, \tlmxs, \tlms, and \tlmm) on the same $26$B tokens used in \S~\ref{subsec:exp1-verify-walle}, and then fine-tune these models on doc-level and chunk-level tasks, obtaining refining models with different sizes.
We then apply these models in doc-level refining and chunk-level refining stages, and use the curated data for from-scratch pre-training.
We report in Table~\ref{tab:walle-adaptation-metrics} the adaptation performance on refining tasks of different refining model sizes.
According to the validation performance, adapting \method works well across all model sizes, all achieving $80\%$ F1 on doc-level 
refinement, and $75\%$ F1
on chunk-level refinement.
We further train these models of different sizes from scratch using data produced by refining models of varying sizes.
In Figure~\ref{fig:exp2-walle-on-different-model-size}, the results indicate that refining models of all sizes help improve performance over raw data, with a consistent absolute gap of $2\%$ over all base model sizes.
While in Figure~\ref{fig:exp2-walle-on-different-model-size}, \tlmxs curated data shows slightly better downstream performance, it has a significantly lower token-level retention ratio (\m{\mathbf{23.2\%}} vs. \m{\mathbf{28.8\%}}) compared to larger models as reflected in Table~\ref{tab:walle-adaptation-metrics}.
This implies that moderately larger models suggest a favorable balance between data quality and quantity.
These additional tokens likely provide more knowledge during pre-training without compromising downstream benchmark performance, showcasing an effective trade-off between data refinement and information preservation.

\input{figures/exp2-beyond}

\paragraph{\method works well across pre-training corpora.}

To assess the applicability of \method across various pre-training corpora, we extend our experiments beyond \redpj to include C4 and the recently released top-quality corpus including \fineweb, \fineweb-Edu, and DCLM. For consistency, we apply exactly the same \method-xs refining models detailed in Table~\ref{tab:walle-adaptation-metrics} to these corpora without constructing new SFT data for each corpus. 
We conducted larger-scale experiments by training our model on approximately \m{50} billion tokens, again achieving notable improvements. On ten downstream benchmarks, models trained on our method's curated data showed improvements of \m{+2.0\%} on \redpj, \m{+3.1\%} on C4, \m{+2.4\%} on \fineweb, \m{\mathbf{+0.9\%}} on \fineweb-Edu, and \m{\mathbf{+1.7\%}} on DCLM.

\paragraph{ProX trains language models with much greater efficiency.}
To demonstrate the non-trivial nature of these results, we compared models trained on \method curated data against various models trained by different approaches.
These include models like
\tinyllama-1.1B-3T (trained directly on $3$ trillion tokens, about $\mathbf{60}\times$ of our training tokens and $\mathbf{40}\times$ training FLOPs),
\textsc{SheadLlama}-1.3B (denoted as S-Llama, a pruned version of \llamaii{}-7B, with extra training on $50$ billion tokens),
and models using LLM data synthesis, such as \textsc{InstructionLM}-1.3B~(denoted as Inst-LM) and \textsc{Cosmo}-1.8B.
Our results, including \tlmm~(\method) and \tlmm~(Raw), are presented alongside all these baselines in Figure~\ref{fig:across-corpora-bar-chart}.
On \fineweb, which is recognized for its high-quality data, \tlmm using \method-refined data performs comparably to pruned models like \textsc{SheadLlama}-1.3B and \tinyllama-1.1B, despite their reliance on additional pruning techniques or much larger datasets.
Moreover, using much less inference-time computing overhead, our model surprisingly outperforms models that rely heavily on LLM data synthesis, underscoring \method's efficiency.
Notably, models like \textsc{Instruct-LM}-1.3B, trained on $100$ billion tokens leveraging a fine-tuned \mistral-7B synthesizer, and \textsc{Cosmo}-1.8B, trained on $180$ billion tokens (including $25$ billion tokens synthesized by \mistral-8x7B), require significantly more computational resources than \method.

\subsection{Applying \method to Domain-Specific Contiual Preraining}~\label{subsec:exp3-math-cpt}

We also demonstrate the potential of \method in the continual pre-training scenario, specifically, in the mathematical domain.
We apply the very same pipeline as in general domains to the already cleaned \owm corpus~\citep{paster2023openwebmath}, aiming to further refine and mine the high quality and clean data from the vast web pages crawled in it. 
We then adapt and apply \method-xs series, which was initially trained on general text as described in \S~\ref{subsec:exp2-beyond-size-and-corpora}, and further adapted on math text for the doc-level and chunk-level refining tasks.
Finally, we obtain about $5.5$B tokens left after the document-level cleaning stage and about $4.7$B tokens left after the chunk-level refining stage.
We present the final mathematical evaluation results of models trained on the refined OpenWebMath in Table~\ref{tab:exp3-owm-ct}, with full evaluation results presented in \S~\ref{app:subsec:owm}.

\input{tables/exp-3}

\paragraph{\method boosts math continual pre-training efficiency vastly.} 
Without any domain-specific design, Table~\ref{tab:exp3-owm-ct} shows that pre-training on OpenWebMath refined by \method brings $11.0\%$ average performance improvements for base \tinylm, $14.6\%$ for base \llamaii, $20.3\%$ for base \codellama, $7.6\%$ for base \mistral, which clearly exceed the improvements of all baselines, including their counterparts pre-trained on the original corpus, under the same settings. It is also worth noticing that,  applying the rule-based filtering method does not bring improvements; instead, it leads to a $3.1\%$ performance degradation compared to continual pre-training on the original corpus. This finding implies that there are no universal workable heuristics for all domains, highlighting the demands for automated pipelines just like \method.
Moreover, compared with some existing state-of-the-art math continual pre-training models like \llemma and \internmath typically requiring hundreds of billions of tokens continual pre-training, our \method demonstrates remarkable efficiency gains. 
A more controlled comparison further highlights this efficiency: \llemma{}-7B, based on \codellama-7B, was trained on $200$B tokens, whereas our \method, also starting from \codellama-7B, reaches similar performance levels with just $10$B tokens of training, indicating a $\mathbf{20}$ times reduction in training computes. 
These results suggest that our approach may contribute to more efficient and accessible development of LLMs and could offer a new perspective in domain-specific model adaptation, potentially enhancing how to address specialized LLM in resource-constrained settings.

\section{Analysis}

\subsection{Impact on the original data}

What changes occur in the corpora after applying \method? We compare the document length distribution of the original corpus with that of the \method-refined corpus in Figure~\ref{fig:token-distribution}.
In the general domain corpora~(\redpj, C4, and \fineweb), the data refined by \method exhibits a noticeable shift in the average number of tokens per document.
For instance, in \redpj, we observe that documents with fewer than \m{100} tokens make up a significant portion of the corpus. After applying the \method, the majority of documents contain more than \m{200} tokens, with an average number of tokens per document increasing from \m{1217} to over \m{2000}.
This suggests that very short documents may be noisy and lack sufficient meaningful information to be suitable for pre-training. This shift, however, is not observed in \owm, where the average number of tokens per document is already larger. 
One possible reason for this outlier is that the \owm corpus is collected mostly from sources different from the general domain, \emph{e.g.,} online forums like Stack Exchange, and academic publisher websites such as arXiv.
The noises of these sources can be quite different from general domains.
Further case studies on these documents are provided in \S~\ref{app:subsec-case-study}.

\input{figures/tok_analysis}

\subsection{Computing Overhead Analysis}~\label{subsec:analyze-compute-flops}
\vspace{-5.0mm}

Although \method demonstrates promising results in downstream tasks, it is important to acknowledge that large-scale model inference still requires a substantial computing budget. 
For example, as mentioned in \S~\ref{subsec:exp1-verify-walle}, and in Table~\ref{app:tab-pt-corpora}, the \redpj corpus used for training \tlms was refined from about $60$B raw tokens.
As calculated in \S~\ref{app:subsec-flops-analysis},
if we utilize \method-XS for both two refining stages, the additional computational overhead will amount to approximately $C = 5 \times 10^{19}$ FLOPs, 
which is equivalent to training an additional $12$B tokens on \tlms and $5$B tokens on \tlmm. It is noteworthy that this overhead ratio keeps decreasing as model size increases, meaning that the relative computational cost diminishes for larger models.
\input{figures/flops_analysis}

In Figure~\ref{fig:analyze-perf-vs-flops}, we compare the FLOPs consumed by checkpoints with similar performance, both with and without applying \method, across three different model sizes. 
As the model size increases, the proportion of inference FLOPs required for applying \method decreases.
For the \m{0.7}B model, the total FLOPs when using \method are already lower than without it ($6.3\times1e19$ vs. $6.7\times1e19$).
Notably, for the largest \m{1.7}B model, we achieve performance comparable to a model pre-trained on the original data, but with only $58\%$ of the total FLOPs. This demonstrates that refining methods like \method not only enhances data quality but also becomes more computationally efficient as model sizes grow, reinforcing the value of allocating additional resources to refining pre-training data.

\section{Related Works}

\paragraph{Pre-training Data Processing}
Raw data collected from public sources (\emph{e.g.}, CommonCrawl) are noisy, and directly using these data can greatly hurt model performance; thus, it has been a common practice to execute extensive pre-processing before pre-training~\citep{touvron2023llama1, together2023redpajama,penedo2024fineweb}.
The pipeline usually starts with document preparation, which includes URL filtering, text extraction, language-based filtering~\citep{smith2022using}.
The remaining document will then undergo several quality checks with heuristic rules like overall length, symbol-to-word ratio, and other
criteria to determine whether it is kept, partially or fully aborted~\citep{zhang2024map,dou2024sailor,qiu2024wanjuan}.
Finally, these documents are deduplicated using different matching methods, \emph{e.g.}, fuzzy match like MinHash~\citep{broder1997resemblance}, or exact sequences matches~\citep{penedo2023refinedweb}.
In \method, we uses the language model for further data refining, outperforming heuristic rules with acceptable computational overhead.
\paragraph{Data Selection Methods}
Data selection, slightly distinct from data processing, is more commonly applied in the later stages of large-scale data pre-processing.
In supervised fine-tuning~(SFT), it typically involves selecting a much smaller subset of samples to minimize tuning overhead while maintaining performance~\citep{liu2024what}.
Recent efforts have extended these selection strategies to the pre-training stage~\citep{engstrom2024dsdm, xie2023data, ankner2024perplexed,sachdeva2024train,liu2024regmix}.
For instance, \citet{wettig2024qurating} train a rater model to score documents on four quality criteria in SlimPajama~\citep{cerebras2023slimpajama} and conduct pre-training on a resampled subset based on scores.
MATES~\citep{yu2024mates} apply a smaller model for estimating data influence during pre-training, enabling dynamic data selection schema.
Moreover, as mentioned in \textsc{Llama-3}~\citep{metallama3}, \llamaii models~\citep{touvron2023llama} was used as text-quality classifiers that underpin \textsc{Llama-3}'s training data.
Instead of merely selecting documents, \method enables more fine-grained operations within documents, contributing to further performance improvements.
\vspace{-2.0mm}
\paragraph{Model-based Data Synthesizing}
Another branch of research focuses on editing or rephrasing existing data with models to improve the data quality.
\citet{fan2024reformatted} use ChatGPT to rephrase several instruction-tuning datasets for a clear format based on massive scenario-based criteria. 
\citet{yue2024mammoth2} use LLMs to extract and refine $5$M QA pairs from web documents,
obtaining \m{10}M instruction-response pairs.
Synthesis techniques have also been applied in the pre-training phase such as the \textsc{Phi} series~\citep{gunasekar2023textbooks-phi1,li2023textbooks-phi1.5}.
Recently, \citet{maini2024rephrasing} and \citet{cheng2024instruction} utilize off-the-shelf instruction-tuned models to paraphrase web documents in specific styles such as QA, and mix these synthetic rephrases with real data in pre-training.
\citet{benallal2024cosmopedia} further synthesize from mere seed topics, by prompting LLMs to generate pre-training samples in a cleaner format like textbooks. However, despite its success, it typically requires substantial computation to synthesize a pre-training-scale corpus, and more critically, it inevitably inherits flaws from the advanced model, also suffering from hallucination issues~\citep{liu2024best}.
In this work, we focus on leveraging language models to lift data quality through the synthesis of executable and interpretable programs, rather than directly generating data. We demonstrate that \method could clearly improve data quality at scale only with acceptable extra computing.

\paragraph{Inference Time Scaling} Recent trends in language models have begun to explore the potential of allocating additional computing at inference time, complementing the extensive computations already deviated to the pre-training and post-training phases. Several studies have demonstrated the potential of this approach, showing that smaller language models equipped with additional inference-time computing can perform comparably to, or even outperform, significantly larger models, evidenced across various domains, including code generation~\citep{hassid2024the, DBLP:journals/corr/abs-2407-21787-large-language-monkeys-scaling-inference-compute}, and math problem-solving~\citep{DBLP:journals/corr/abs-2408-03314-scaling-llm-test-time-compute, DBLP:journals/corr/abs-2408-00724-an-empirical-analysis-compute-optimal-inference}. The significance of this approach has been further corroborated by OpenAI's latest o1 model release~\citep{openaio1}. While these studies focus on scaling computing on test time, our work demonstrates \textbf{an alternative perspective on inference computing scaling}. We advocate for allocating computing to refine pre-training corpora, particularly given that Internet-based corpora have been extensively utilized in language model pre-training. Our proposed \method demonstrates remarkable gains in pre-training efficiency by investing moderately additional compute in the corpus refinement, facilitating more efficient and accessible development of LLMs.

\section{Conclusion}

We introduced \method, a framework that uses language models to refine pre-training data at scale through program generation.
Our extensive experiments show that \method curated data improves model performance by over \m{2\%} on various downstream benchmarks and is effective across different model sizes and pre-training datasets. 
For domain-specific continual pre-training, models trained on \method curated tokens also yield significant improvements in $20\times$ fewer tokens, and comparable to state-of-the-art models trained on $200$B tokens.
Further analysis also implies applying \method can achieve similar results with less computing power for large-scale LLM pre-training. 
In summary, these results demonstrate \method's potential for greatly improving data quality and reducing costs in language model training. 

\section{Implications and Future Directions}
The strong results from \method highlight the potential of automated data refinement to significantly improve model performance while reducing computational costs. By refining data more effectively, \method opens new possibilities for improving training efficiency and achieving better results across a range of benchmarks.
Looking ahead, these results suggest several future directions.
First, incorporating additional refining operations like reformatting and rephrasing could further enhance data quality. 
Second, improving efficiency by reducing model size and applying inference acceleration techniques is a key goal. Expanding \method to domains like code and multilingual data is also promising. 
Scaling up with more computational resources will allow for a thorough evaluation of its potential. Finally, we believe that prioritizing data refinement before pre-training can greatly improve training efficiency, and we encourage continued exploration in this area.

\input{ack}

\newpage
\bibliographystyle{unsrtnat} %
\bibliography{iclr2024_conference}

\clearpage
\newpage
\input{appendix}

\end{document}

%% file: author.tex
\author{
    {\textbf{Fan Zhou}}\thanks{Equal contribution.~$^\ddagger$Corresponding author.}~~$^\bsjtu$$^\bgair$ \ \
    {\textbf{Zengzhi Wang}}$^*$$^\bsjtu$$^\bgair$ \ \
    {\textbf{Qian Liu}}$^\bsea$ \ \
    {\textbf{Junlong Li}}$^\bsjtu$ \ \
    {\textbf{Pengfei Liu}}$^\ddagger$$^\bsjtu$$^\bgair$$^\bailab$ \\
    $^\bsjtu${\rm{Shanghai Jiao Tong University}} \ \
    $^\bailab${\rm{Shanghai Artificial Intelligence Laboratory}} \ \
    $^\bsea${\rm{Sea AI Lab}} \ \
    $^\bgair${\rm{Generative AI Research Lab (GAIR)}} \\
    {\normalfont\texttt{\{zhoufan98,pengfei\}@sjtu.edu.cn}}
}

%% file: figures/intro_fig.tex
\vspace{-3.0mm}
\begin{figure}[htbp]
    \centering
    \begin{minipage}[t]{0.43\textwidth}
        \centering
        \includegraphics[width=\textwidth]{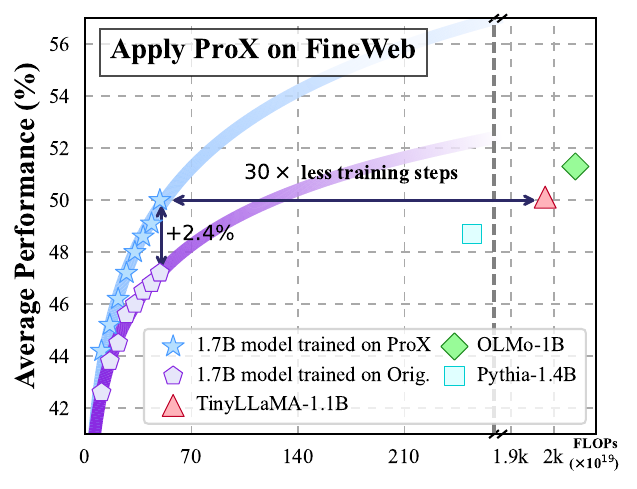}
    \end{minipage}
    \hspace{1.0mm}
    \begin{minipage}[t]{0.43\textwidth}
        \centering
        \vspace{-4.61cm}
        \includegraphics[width=\textwidth]{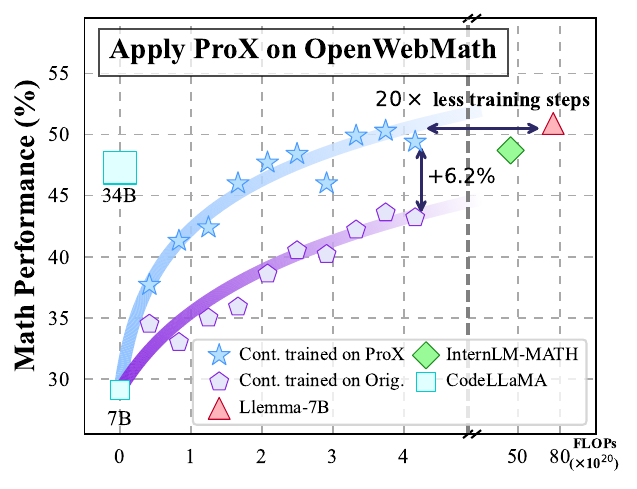}
    \end{minipage}
    \vspace{-3.0mm}
    \caption{Training FLOPs v.s. average downstream performance. Although these corpora have gone through expert-crafted rules, applying \method still yields significant improvements over these baseline models trained with original data corpus. Moreover, with much less training FLOPs, 
    model trained on \method curated data show comparable performance with existing models.
     }
    \label{fig:combined_plots}
\end{figure}
\vspace{-7.0mm}

%% file: tables/program-space.tex
\begin{table}[!t]
  \centering
  \small
  \caption{\method program design of document-level and chunk-level refining stage. For input, \texttt{doc} and \texttt{chunk} will be sent into the corresponding function as string-type inputs for execution.}
  \vspace{-2.5mm}
  \setlength{\tabcolsep}{3.5pt}
  \label{tab:program-space}
  \resizebox{0.97\linewidth}{!}{%
    \begin{tabular}{clp{15.em}}
    \toprule
\multicolumn{1}{c}{\textbf{Stage}} & \multicolumn{1}{l}{\textbf{Function Interface}} & \multicolumn{1}{l}{\textbf{Description}} \\
    \midrule
    \multicolumn{1}{c}{\multirow{2}[2]{*}{\begin{tabular}[c]{@{}c@{}}Document\\ Level\end{tabular}}} &
    \multicolumn{1}{l}{\multirow{1}{*}{\begin{tabular}[c]{@{}l@{}}\texttt{drop\_doc()}$\rightarrow$ \texttt{<None>}\end{tabular}}}& 
    Delete the whole doc. \\
\cmidrule{2-3}          
    & \multicolumn{1}{l}{\multirow{1}{*}{\begin{tabular}[c]{@{}l@{}}\texttt{keep\_doc()}$\rightarrow$ \texttt{<str>}\end{tabular}}}& Return the orignal doc.\\
    \midrule
    \multicolumn{1}{c}{\multirow{7}[7]{*}{\begin{tabular}[c]{@{}c@{}}Chunk\\ Level\end{tabular}}} & \multicolumn{1}{p{25.5em}}{\texttt{remove\_lines(line\_start, line\_end)}$\rightarrow$ \texttt{<str>}\newline{}$~~\triangleright$ \texttt{line\_start<int>}, index of the first line to be removed\newline{}$~~\triangleright$ \texttt{line\_end<int>}, index of the last line to be removed} & Delete noisy lines from chunk;\newline{}Return chunk after removal. \\
\cmidrule{2-3}          & \multicolumn{1}{p{25.5em}}{\texttt{normalize(source\_str, target\_str)}$\rightarrow$ \texttt{<str>}\newline{}$~~\triangleright$ \texttt{source\_str<str>}, the noisy string pattern\newline{}$~~\triangleright$ \texttt{target\_str<str>}, the string for replacement} & Replace strings with normalized ones;\newline{}Return chunk after replacement. \\
\cmidrule{2-3}          & \multicolumn{1}{l}{\multirow{1}{*}{\begin{tabular}[c]{@{}l@{}}\texttt{keep\_chunk()}$\rightarrow$ \texttt{<str>}\end{tabular}}} & Return the orignal chunk. \\
    \bottomrule
    \end{tabular}%
    }
  \label{tab:addlabel}%
\end{table}%

%% file: tables/exp-1.tex
\begin{table}[!t]
  \centering
  \caption{Zero-shot performance on $10$ selected tasks. All models use the same \tlms architecture and are trained on \redpj. The doc-level~(\method-D) and chunk-level~(\method-C) refining are done by fine-tuning the raw data pre-trained model as a refining model.
  \textbf{Bolded} entries represent the best results.
  \textbf{\#Win} represents the number of tasks where the method achieved the best performance.
  }
  \newcommand{\B}[1]{\textbf{#1}}
  \newcommand{\degrad}{\color[rgb]{0.8,0,0}{\raisebox{0.3ex}{\scalebox{1.0}[0.8]{$\bm{\downarrow}$}}}}
  \setlength{\tabcolsep}{3.0pt}
  \resizebox{0.95\textwidth}{!}{%
 \begin{tabular}{l|cccccccccc|ll}
\toprule    \textbf{Method} & \textbf{ARC-C} & \multicolumn{1}{c}{\textbf{ARC-E}} & \multicolumn{1}{c}{\textbf{CSQA}} & \multicolumn{1}{c}{\textbf{HellaS}} & \multicolumn{1}{c}{\textbf{MMLU}} & \multicolumn{1}{c}{\textbf{OBQA}} & \multicolumn{1}{c}{\textbf{PIQA}} & \multicolumn{1}{c}{\textbf{SIQA}} & \multicolumn{1}{c}{\textbf{WinoG}} & \multicolumn{1}{c|}{\textbf{SciQ}} & \multicolumn{1}{c}{\textbf{AVG}} & \multicolumn{1}{c}{\textbf{\#Win}} \\
    \midrule
    Raw   & 26.1  &            44.3  &            29.7  &            39.1  &            27.3  &            29.2  &            66.9  &            39.0  &            52.0  &            67.4  &            42.1  & 0 / 10 \\
    \midrule
    \multicolumn{13}{c}{{Rule-based filtering}: \textsc{Go} = \gopher rules, \textsc{C4} = C4 rules, \textsc{Fw} = \fineweb rules.} \\
    \midrule
    \textsc{Go}  & 25.7  & 44.0  & 31.3 & 40.2  & 27.3  & 29.0  & 66.3  & 39.0 & 51.2  & 68.9  & 42.3  & 0 / 10 \\
    \textsc{C4}  & 25.0 & 46.0 & 31.0 & {40.5} & 27.1 & 29.2 & \textbf{68.5} & \textbf{40.5} & 51.7 & 66.6 & 42.6  & 2 / 10 \\
    \textsc{Fw}  & 25.2 & 46.8 & \textbf{32.6} & 39.6 & 27.2 & 29.0 & 66.5 & 39.4 & \textbf{52.4} & 69.2 & 42.8   & 2 / 10 \\
    \textsc{Go}+\textsc{C4}+\textsc{Fw}  & 25.2  & 43.9  & 30.0 & 41.9  & 27.5  & 31.0  & 67.0  & 39.9 & 51.9  & 65.3  & 42.3  & 0 / 10 \\
    \midrule
    \multicolumn{13}{c}{\method~(ours): \textsc{D} = Doc-level Programming, \textsc{C} = Chunk-level Programming.}\\
    \midrule
    \rowcolor{cellHighlight}
    \method-D    & \textbf{26.6} & 49.7  & 30.1  & 40.5  & \textbf{29.4} & 30.4  & 66.3  & 39.0  & 51.2  & 71.6  & 43.5  & 2 / 10 \\
    \rowcolor{cellHighlight}
    \method-D+C & {26.4}  & \textbf{51.9} & 30.9  & \textbf{42.4} & \textbf{29.4} & \textbf{31.6} & {67.9} & 40.0  & {52.2} & \textbf{73.5} & \textbf{44.6} & \textbf{5} / \textbf{10} \\
    \bottomrule
    \end{tabular}%
 }%
  \label{tab:exp1-zero-shot-avg-performance}%
\end{table}

%% file: figures/exp1-curve.tex
\begin{figure}[!t]
    \centering
    \begin{minipage}[]{0.37\textwidth}
        \centering
        \includegraphics[width=\textwidth]{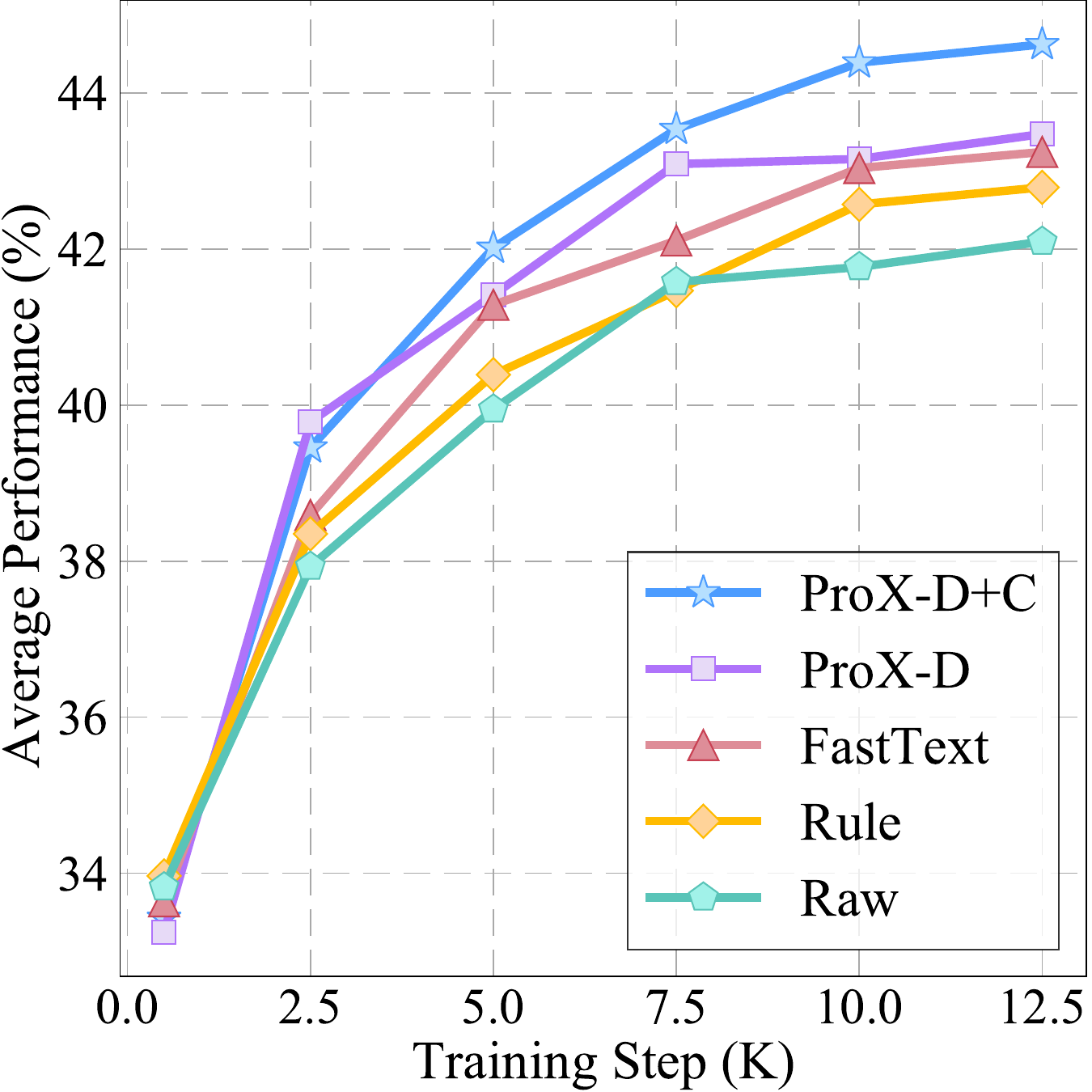}
        \caption{Downstream zero-shot performance w.r.t. different training steps: first $0.5$K, then evenly from $2.5$K to $12.5$K. Rule: the best performing \fineweb{} rule in Table~\ref{tab:exp1-zero-shot-avg-performance}.
        }
        \label{fig:exp1-bench}
    \end{minipage}
    \hfill
    \begin{minipage}[]{0.6\textwidth}
        \centering
        \captionof{table}{Comparison with different data selection methods on $8$ benchmarks using the C4 corpus and \pythia architecture.
        \textbf{\#Win} represents the count of best performance.
        }
        \input{tables/exp-1-data-selection-baselines}
        \label{tab:exp1-data-selection-baseline}
    \end{minipage}
    \label{fig:exp1-performance}
\end{figure}

%% file: tables/exp-1-data-selection-baselines.tex
\resizebox{\textwidth}{!}{%
    \setlength{\tabcolsep}{9pt}
    \begin{tabular}{l|cc|c}
    \toprule
    \textbf{Method} & \textbf{0-shot} & \textbf{2-shot} & \textbf{\#Win} \\
    \midrule
    \multicolumn{4}{c}{Model Architecture: \pythia-410M} \\ \midrule
    Random & 42.7  & 43.8 & 0~/~8 \\
    DSIR~\citep{xie2023data} & 42.5    & 43.7 & 1 / 8 \\
    DsDm~\citep{engstrom2024dsdm} & 43.4  & 44.1 & 0 / 8 \\
    QuRating~\citep{wettig2024qurating} & 43.5  & 44.6  & 0 / 8 \\
    MATES~\citep{yu2024mates} & 44.0  & 45.0  & 0 / 8 \\ 
    \midrule
    \rowcolor{cellHighlight}
    \method~(ours)  & \textbf{46.2} & \textbf{47.5} & \textbf{7} / \textbf{8} \\ \midrule
    \multicolumn{4}{c}{Model Architecture: \pythia-1B} \\ \midrule
    Random &44.7  & 45.4 & 0 / 8 \\
    MATES~\citep{yu2024mates} & 45.8  & 46.4  & 1 / 8 \\ 
    \midrule
    \rowcolor{cellHighlight}
    \method~(ours) & \textbf{46.8} & \textbf{48.4} & \textbf{7} / \textbf{8} \\
    \bottomrule
    \end{tabular}%
}

%% file: figures/exp2-beyond.tex
\begin{figure*}[!t]
  \centering
  \begin{minipage}{0.40\textwidth}
    \centering
    \vspace{-7.5mm}
    \small
    \captionof{table}{Refining model's performance on valid set and token retention ratio of original corpus.}
    \vspace{-3.5mm}
    \resizebox{0.95\textwidth}{!}{%
    \setlength{\tabcolsep}{1.5pt}
      \begin{tabular}{l|cc|c}
      \toprule
      \multicolumn{1}{l}{\textbf{Size}} & \textbf{Doc-level} & \multicolumn{1}{c}{\textbf{Chunk-level}} & \multicolumn{1}{l}{\textbf{Kept Ratio}}\\
      \midrule
      \tlmxs{}    & 82.6  & 75.2 & 23.2\%\\
      \tlms{}     & 81.3  & 75.6 & 25.6\%\\
      \tlmm{}     & 83.7  & 77.3 & 28.8\%\\
      \bottomrule
      \end{tabular}%
    }
    \label{tab:walle-adaptation-metrics}%
    \vspace{4.0mm}
    \includegraphics[width=1.\textwidth]{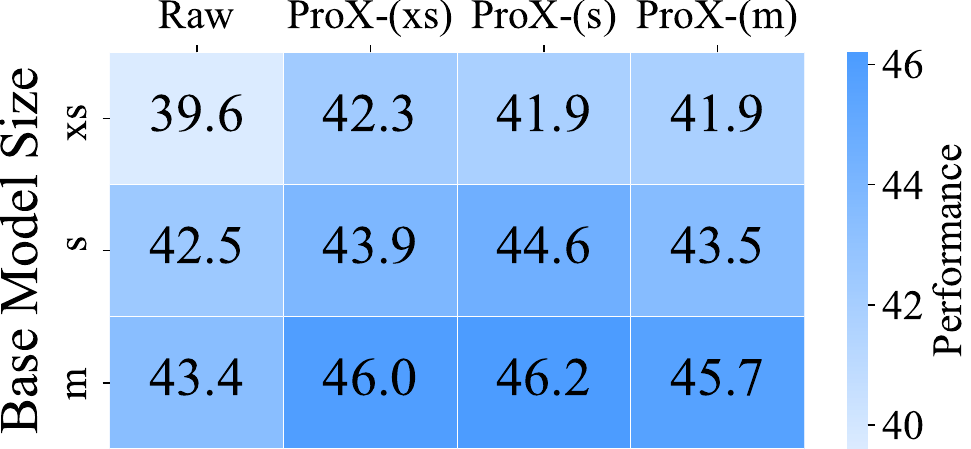}
    \vspace{-4.5mm}
    \captionof{figure}{\method's effect over different model sizes.
    }
    \label{fig:exp2-walle-on-different-model-size}%
  \end{minipage}%
  \hfill
  \begin{minipage}{0.55\textwidth}
    \centering
    \vspace{-7.5mm}
    \includegraphics[width=\textwidth]{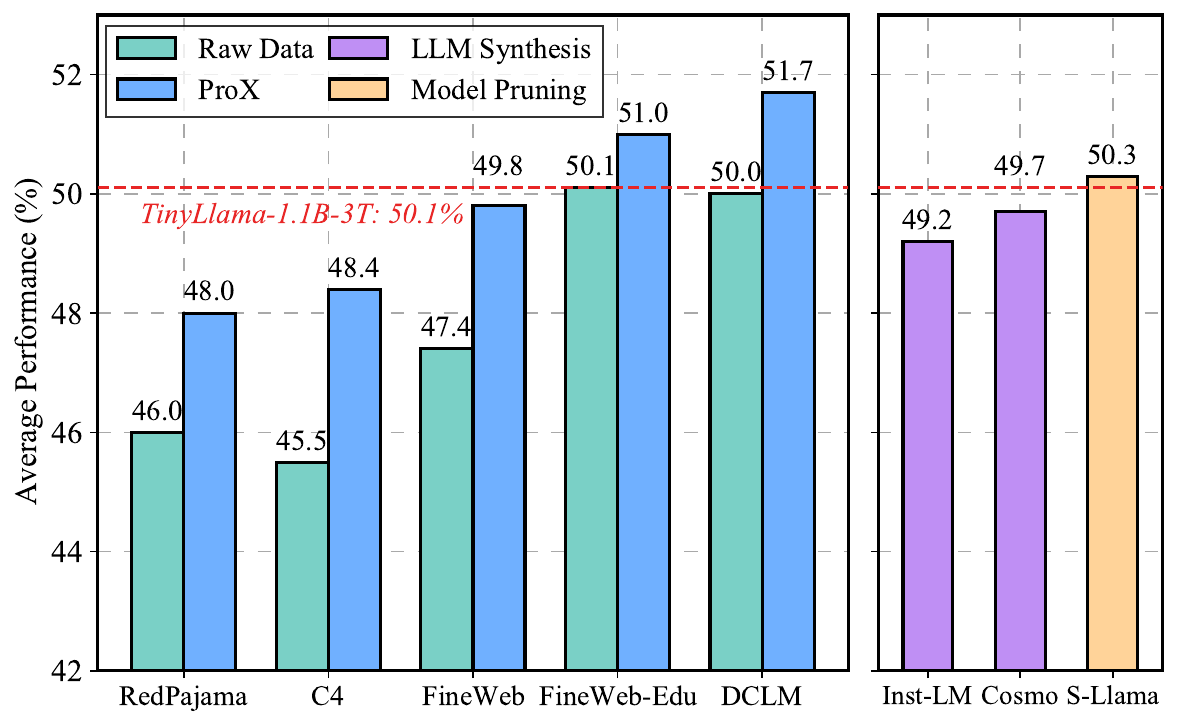}
    \vspace{-5.0mm}
    \captionof{figure}{Performance of original data and \method{} curated data trained models across different datasets using $\approx50$B tokens and comparison with existing models trained using different techniques like LLM data synthesis and direct model pruning.
    }
    \label{fig:across-corpora-bar-chart}
  \end{minipage}
\vspace{-3.0mm}
\end{figure*}

%% file: tables/exp-3.tex
\begin{table}[!t]
  \small
  \centering
  \setlength{\tabcolsep}{2.0pt}
  \caption{OpenWebMath Continual Pre-training~(CPT) Results. All models are tested using few-shot CoT prompts. \llemma{} and \internmath{} are continual pre-trained models from \codellama{} and \textsc{InternLM2}~\citep{team2023internlm} with public available data, respectively.
  \textsc{DeepSeek-LLM} denotes an internal DeepSeek model, and the model trained on \owm{} introduced by~\citet{shao2024deepseekmath}. Note that the unique tokens and training tokens in the column refer exclusively to the token numbers from math-specific corpora (calculated by corresponding tokenizers). $^\dag$: MQA evaluation of \intern{} is based on an alternative prompt due to non-prediction issues with the original prompt. The \textbf{bolded} entries represent the best results within the same base model.
  }
  \newcommand{\U}[1]{{#1}}
  \resizebox{\linewidth}{!}{%
    \begin{tabular}{l|ll|cc|ccccccccc|c}
    \toprule
    \textbf{Model} & \multicolumn{1}{c}{\textbf{Size}} & \multicolumn{1}{l}{\textbf{Method}} & \begin{tabular}[c]{@{}c@{}}\textbf{Uniq}\\ \textbf{Toks}\end{tabular} & \multicolumn{1}{c|}{\textbf{\begin{tabular}[c]{@{}c@{}}Train\\ Toks\end{tabular}}} & \textbf{GSM8K} & \textbf{MATH} & \textbf{SVAMP} & \textbf{ASDiv} & \textbf{MAWPS} & \textbf{TAB} & \textbf{MQA} & \begin{tabular}[c]{@{}c@{}}\textbf{MMLU}\\ \textbf{STEM}\end{tabular} & \textbf{\begin{tabular}[c]{@{}c@{}}{SAT}\\ {MATH}\end{tabular}} & \textbf{AVG} \\
    \midrule
    \multicolumn{15}{c}{\cellcolor[HTML]{F2F2F2}Existing Continual Pre-training for Reference} \\
    \midrule
    \multirow{2}{*}{\centering\textsc{DeepSeek-LLM}}& 1.3B 
        & - & - & - & 2.9  &  3.0 &  - & -  & - & - & - &  19.5 & 15.6 & - \\
        & 1.3B & - & 14B & 150B & 11.5  &  8.9 &  - & -  & - & - & - &  29.6 & 31.3 & - \\
    \midrule
    \multirow{2}[2]{*}{\centering\codellama{}~(Base)} & 7B    &  -     &  -  & - & 11.8 & 5.0 & 44.2 & 50.7 & 62.6 & 30.6 & 14.3 & 20.4 & 21.9 & 29.1  \\
    &  34B  & -     & -   &  -  & 31.8 & 10.8 & 61.9 & 66.0 & 83.4 & 51.6 & 23.7 & 43.0 & 53.1 & 47.3  \\
    \midrule
    \multirow{2}{*}{\centering\llemma{}} & 7B    & -     & 55B   & 200B  & 38.8  & 17.2  & 56.1  & 69.1  & 82.4  & 48.7  & 41.0  & 45.4  & 59.4  & 50.9~(+21.8) \\
         & 34B   & -     & 55B   & 50B   & 54.2  & 23.0  & 67.9  & 75.7  & 90.1  & 57.9  & 49.8  & 54.7  & 68.8  & 60.1~(+12.8) \\
    \midrule
    \multirow{2}{*}{\centering\intern{}} & 7B    &  -     &  -  & - & 27.0 & 6.6 & 49.0 & 59.3 & 74.8 & 40.1 & 20.9$^\dag$ & 19.0 & 28.1 & 36.1  \\
    &  20B  & -     &  -  & -   & 50.6 & 18.8 & 72.5 & 75.9 & 93.9 & 45.4 & 33.1 & 53.7 & 59.4 & 55.9  \\
    \midrule
    \multirow{2}{*}{\centering\internmath{}} & 7B    & -     & 31B   & 125B  & 41.8  & 14.4  & 61.6  & 66.8  & 83.7  & 50.0  & 57.3  & 24.8  & 37.5  & 48.7~(+12.6) \\
          & 20B   & -     & 120B  & 500B  & 65.4  & 30.0  & 75.7  & 79.3  & 94.0  & 50.9  & 38.5  & 53.1  & 71.9  & 62.1~(+6.2) \\
    \midrule
    \multicolumn{15}{c}{\cellcolor[HTML]{F2F2F2}Applying Data Refinement Approaches} \\ \midrule
    \tinyllama (Base) & 1.1B  & -     & -     & -     & 2.8   & 3.2   & 10.9  & 18.0  & 20.2  & 12.5  & 14.6  & 16.4  & 21.9  & 14.7 \\ \midrule
    \multirow{4}[1]{*}{\centering\tinyllama (CPT)} & 1.1B  & -     & 15B   & 15B   & 6.2   & 4.8   & 22.3  & 36.2  & 47.6  & 19.3  & 11.6  & 20.7  & \U{25.0}  & 21.5 (+6.8) \\
          
          & 1.1B  & \textsc{Rho}   & 15B   & 9B$^*$\tablefootnote{\textsc{Rho}-1 only counts the selected tokens that are used for training~(loss calculation).}   & 7.1   & 5.0   & \U{23.5}  & \U{41.2}  & 53.8  & -     & \textbf{18.0} & -     & -     & - \\
          & 1.1B  & Rule  & 6.5B  & 15B   & 4.5   & 2.8   & 17.5  & 29.4  & 39.3  & 15.1  & 12.4  & 19.4  & \U{25.0}  & 18.4 (+3.7) \\

          & \HL{1.1B}  & \HL{\method} & \HL{5B}  & \HL{15B}   & \HL\textbf{9.0} & \HL\textbf{5.6} & \HL\textbf{23.8} & \HL\textbf{41.9} & \HL\textbf{56.9} & \HL\textbf{22.2} & \HL\U{15.6} & \HL\textbf{26.8} & \HL\textbf{31.2} & \HL\textbf{25.7~(+11.0)} \\\midrule
    \llamaii (Base) & 7B    & -     & -     & -     & 14.1 & 3.8          & 39.5 & 51.6 & 63.6 & 30.9  & 12.5  & 32.9      & 34.4     & 31.5 
    \\ \midrule
    \multirow{2}[1]{*}{\llamaii (CPT)} & 7B    & -     & 15B     & 10B   & 29.6 & 13.6 & 49.2 & 61.9 & 78.4 & \U{36.3} & 31.9 & 40.5 & 43.8 & 42.8 (+11.3) \\ 
    & \HL{7B}    & \HL{\method}     & \HL{5B}     & \HL{10B}     &  \HL{\textbf{30.6}} & \HL{\textbf{16.8}} & \HL{\textbf{50.2}} & \HL{\textbf{63.7}} & \HL{\textbf{79.3}} & \HL{\textbf{37.3}} & \HL{\textbf{40.1}} & \HL{\textbf{43.8}} & \HL{\textbf{53.1}} & \HL{\textbf{46.1 (+14.6)}} \\ \midrule
    \codellama{}~(Base) & 7B    &  -     &  -  & - & 11.8 & 5.0 & 44.2 & 50.7 & 62.6 & 30.6 & 14.3 & 20.4 & 21.9 & 29.1  \\ \midrule
    \multirow{2}[1]{*}{\codellama (CPT)} & 7B    & -     & 15B   & 10B  &  31.1  & 14.8          & 51.4  & 62.1  & 81.2  & 33.6   & 30.4   & 40.5       & 43.8      & 43.2 (+14.1) \\ 
    & \HL{7B}    & \HL{\method}  & \HL{5B}   & \HL{10B}  & \HL{\textbf{35.6}}          & \HL{\textbf{17.6}} & \HL{\textbf{55.8}} & \HL{\textbf{67.9}} & \HL{\textbf{82.7}}          & \HL{\textbf{41.3}} & \HL{\textbf{38.9}}          & \HL{\textbf{42.6}}          & \HL{\textbf{62.5}} & \HL{\textbf{49.4} \textbf{(+20.3)}} \\ \midrule
    \mistral (Base) & 7B    & -     & -     & -     & 40.6  & 11.4  & \textbf{65.4}  & 68.5  & 87.0  & \textbf{52.9}  & 32.3  & 50.0  & 56.2  & 51.6 \\ \midrule
    \multirow{2}[1]{*}{\mistral (CPT)} & 7B    & -     & 15B   & 10B   & 44.4  & 19.2  & \U{65.2}  & 69.6  & 88.4  & 46.6  & 43.1  & 50.8  & \U{65.6}  & 54.8 (+3.2) \\
          & \HL{7B}    & \HL{\method} & \HL{4.7B}  & \HL{10B}   & \HL\textbf{51.0} & \HL{\textbf{22.4}} & \HL{64.9} & \HL\textbf{72.9} & \HL\textbf{89.2} & \HL\U{49.8} & \HL\textbf{53.0} & \HL\textbf{54.2} & \HL\textbf{75.0} & \HL\textbf{59.2 (+7.6)} \\ 
    \bottomrule
    \end{tabular}%
    }
  \label{tab:exp3-owm-ct}%
\end{table}%

%% file: figures/tok_analysis.tex
\begin{figure}[h]
    \centering
    \begin{minipage}{0.245\textwidth}
        \centering
        \includegraphics[width=\textwidth]{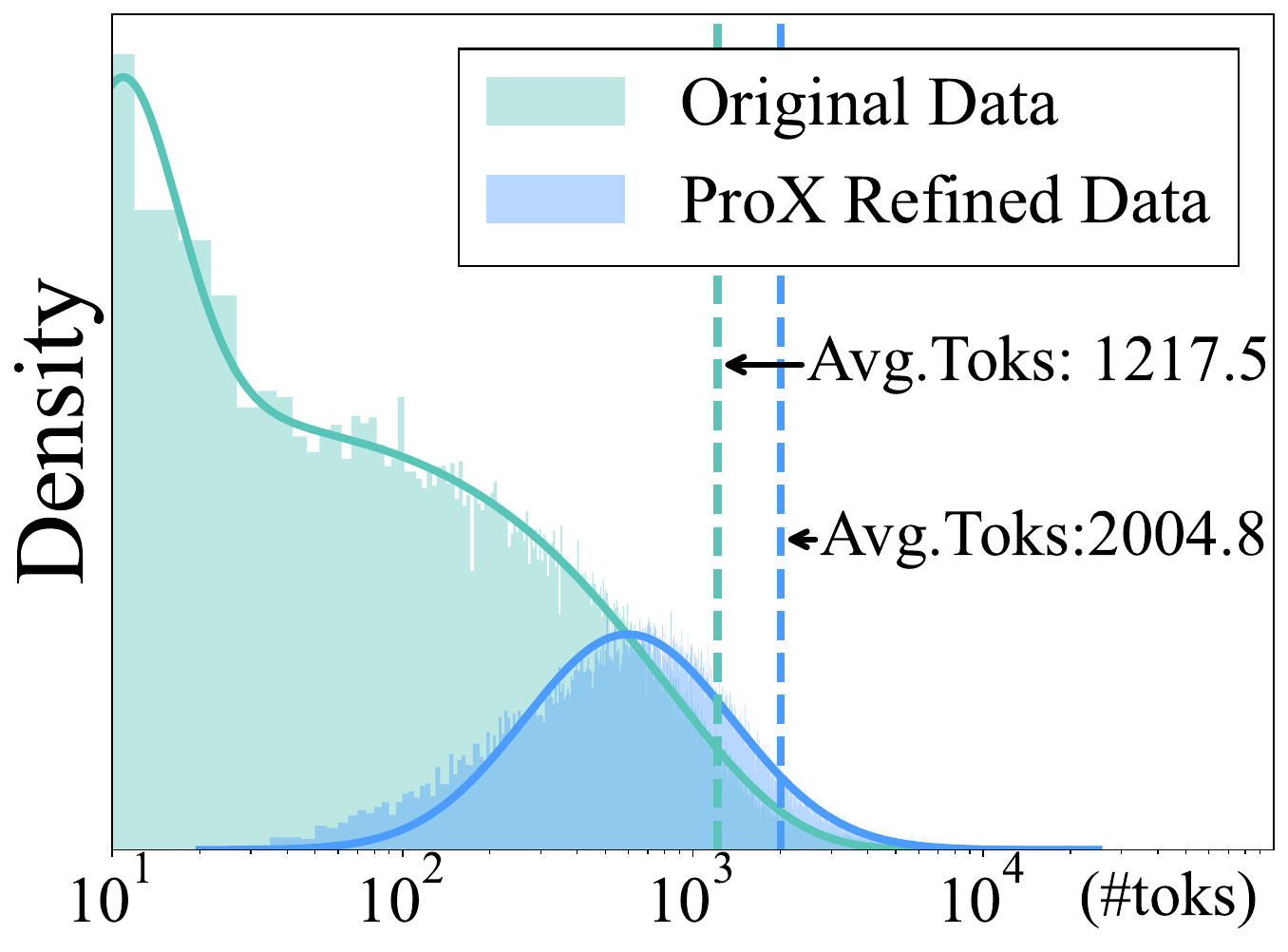}
        \vspace{-8.0mm}
        \caption*{\small{\redpj}}
    \end{minipage}
    \hfill
    \begin{minipage}{0.245\textwidth}
        \centering
        \includegraphics[width=\textwidth]{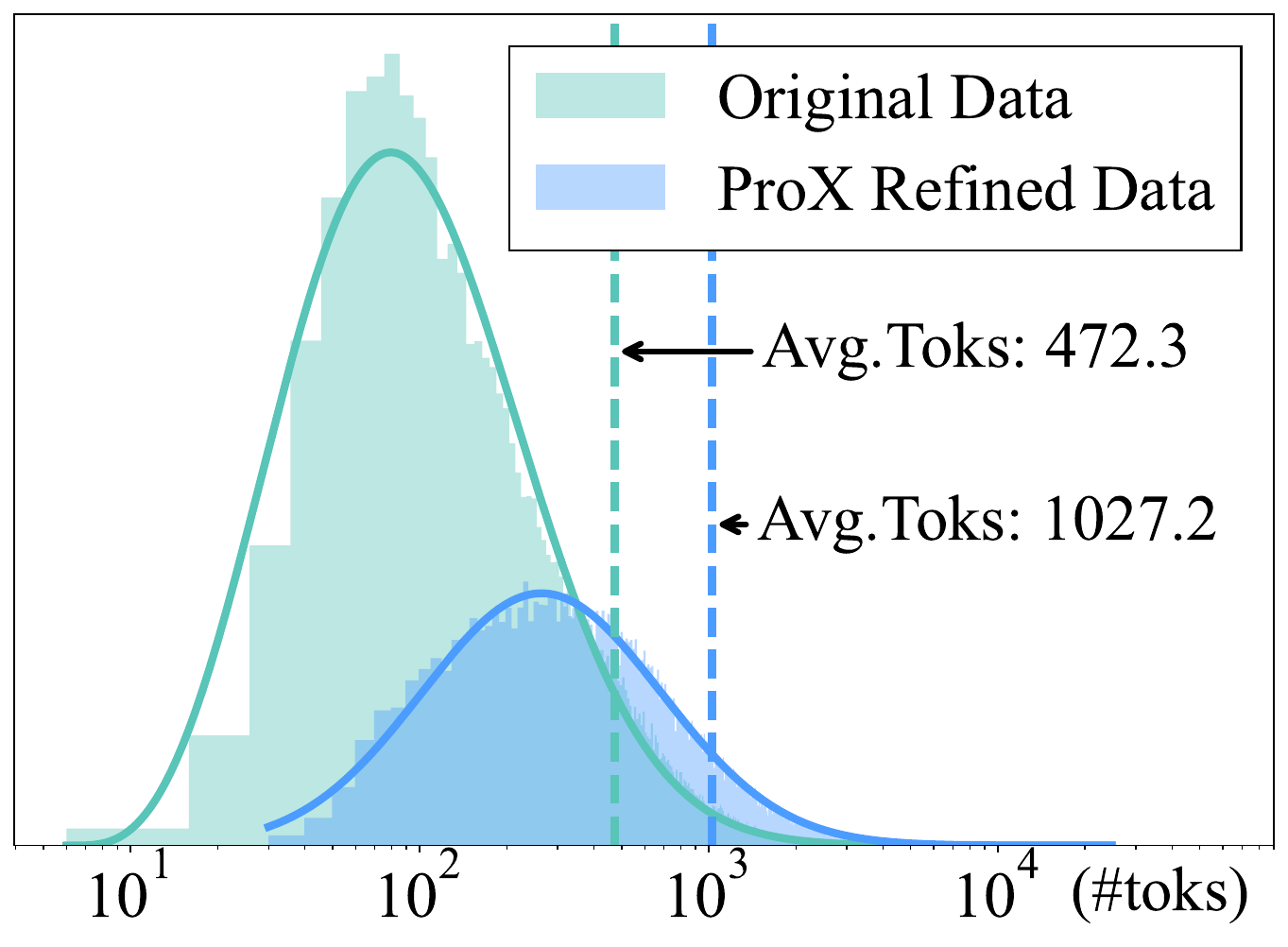}
        \vspace{-8.0mm}
        \caption*{\small{C4}}
    \end{minipage}
    \hfill
    \begin{minipage}{0.245\textwidth}
        \centering
        \includegraphics[width=\textwidth]{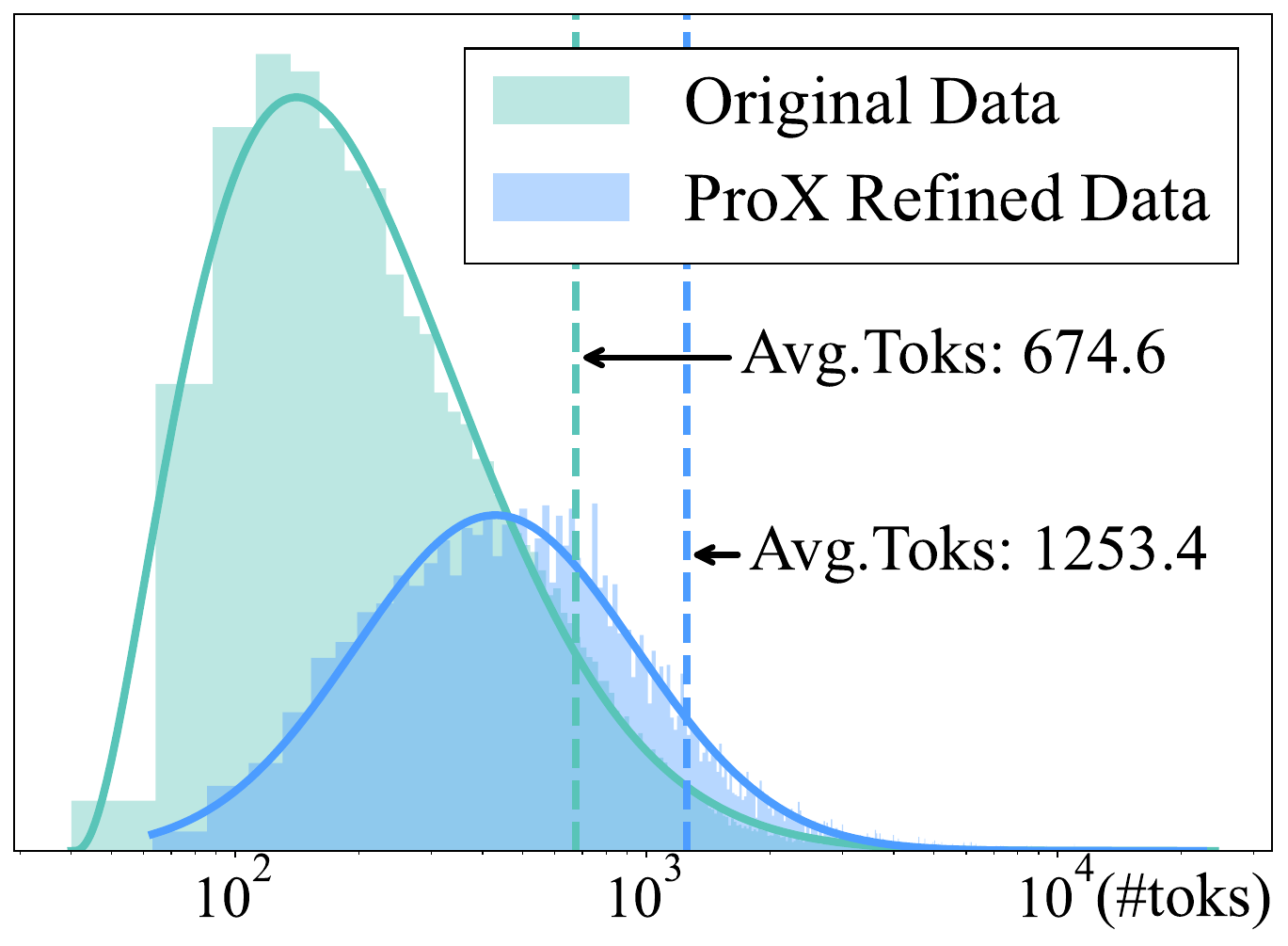}
        \vspace{-8.0mm}
        \caption*{\small{\fineweb}}
    \end{minipage}
    \hfill
    \begin{minipage}{0.245\textwidth}
        \centering
        \includegraphics[width=\textwidth]{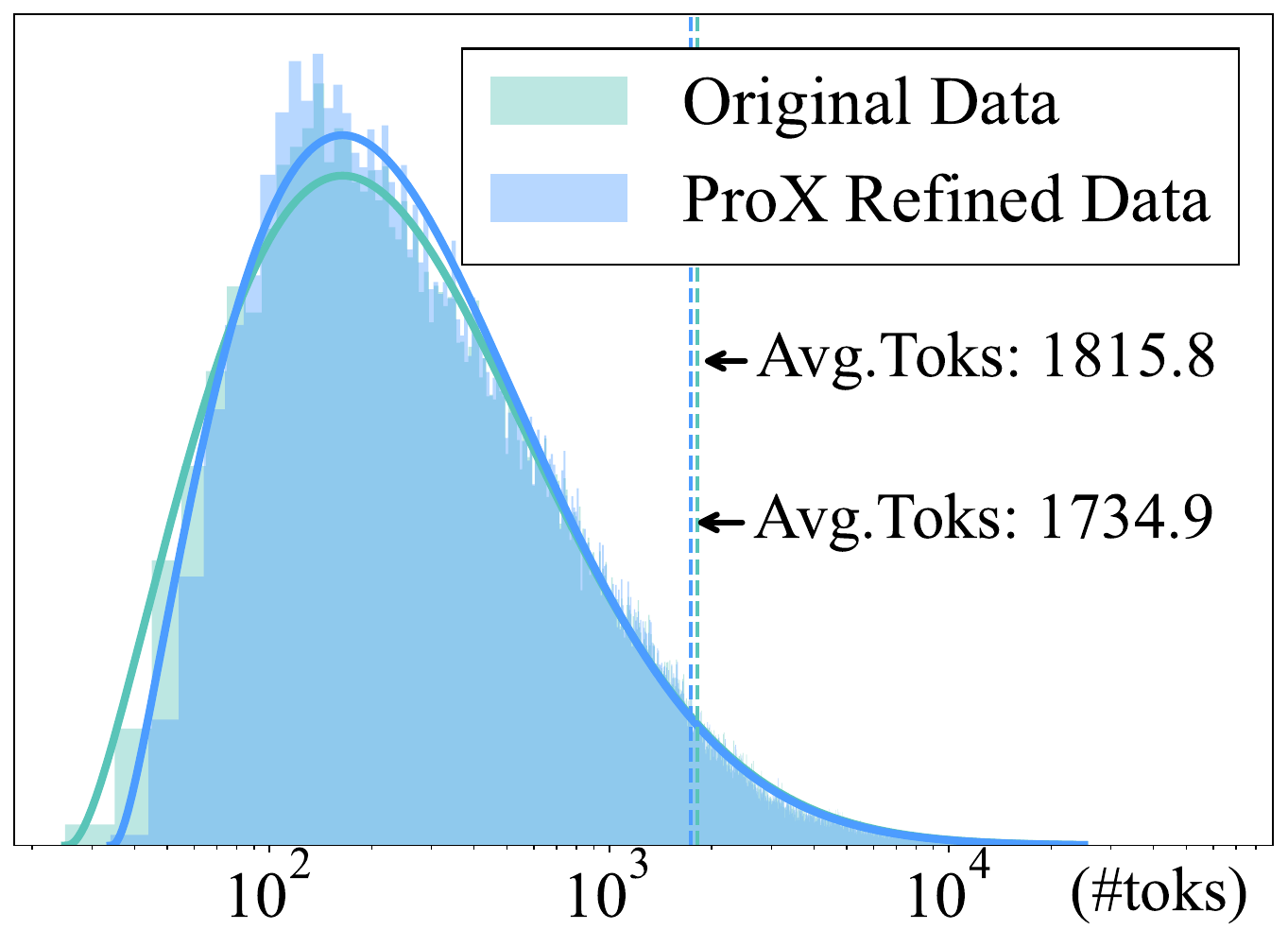}
        \vspace{-8.0mm}
        \caption*{\small{\owm}}
    \end{minipage}
    \vspace{-3.0mm}
    \caption{Comparison of doc's token length distributions between original and \method-refined data.}
    \label{fig:token-distribution}
\end{figure}

%% file: figures/flops_analysis.tex
\begin{wrapfigure}[15]{r}{0.44\textwidth}
\centering
\vspace{-0.pt}
\includegraphics[width=0.90\linewidth]{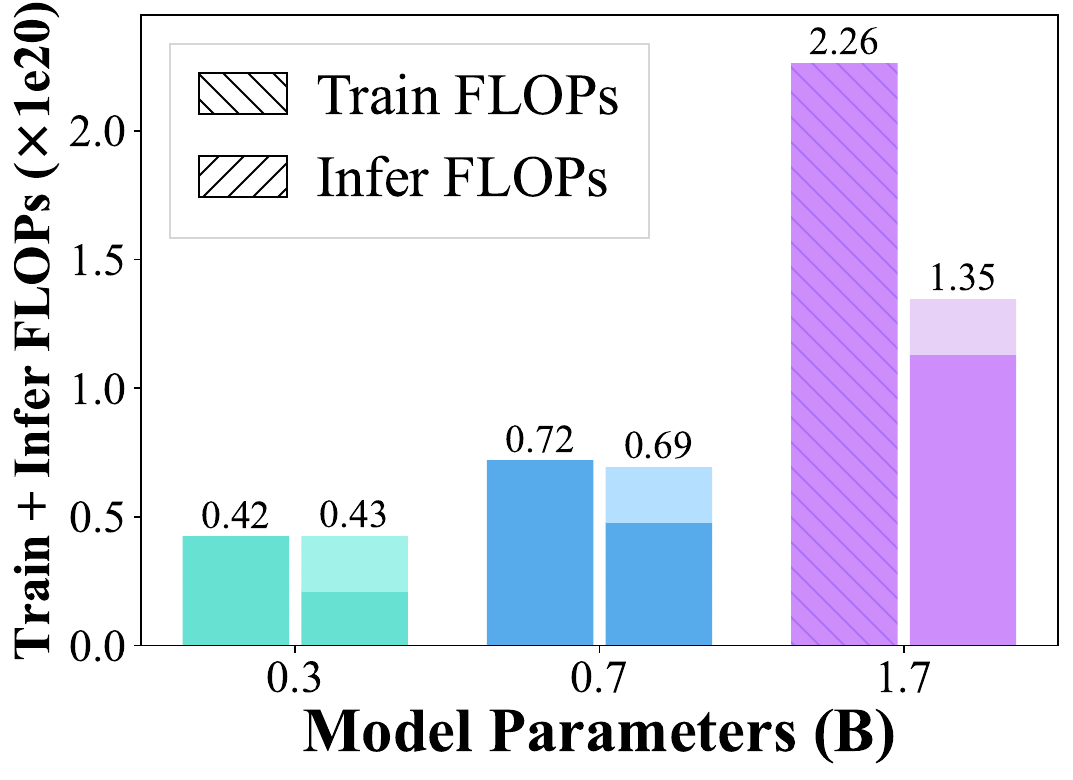}
\vspace{-2.0mm}
\caption[FLOPs comparison]{%
FLOPs comparison for comparable downstream performance with/without \method refining: 0.3B(Avg.Perf = 40.5), 0.7B (41.6), and 1.7B (42.9).\footnotemark}
\label{fig:analyze-perf-vs-flops}
\end{wrapfigure}
\footnotetext{The train FLOPs for the base model (approximately $5.3\times10^{19}$) used to create the refining model are excluded. This is because any pre-trained LLM can theoretically serve as the base for refinement. This also reflects \method's flexibility.}

%% file: ack.tex
\section*{Acknowledgement}
We extend our profound gratitude to Shanghai AI Lab and Sea AI Lab for generously providing valuable computational resources, which were instrumental in the realization of this project.
Our sincere thanks also go to Mingxuan Wang and Jiaze Chen from ByteDance for their crucial support. We are deeply thankful to Ethan Chern from Shanghai Jiao Tong University and Yuqing Yang from University of Southern California for their early discussions and insightful contributions, and equally grateful 
to Zhoujun Cheng from UC San Diego, Yiheng Xu and Tianbao Xie from University of Hong Kong, and Terry Yue Zhuo from Monash University for their valuable feedback, 
to Guilherme Penedo and Loubna Ben Allal from Hugging Face for their guidance on hyper-parameter tuning, 
to Zhibin Gou from Tsinghua University for providing advise on continual pre-training,
to Lyumanshan Ye for helping with illustrations and color scheme design.
Finally, special thanks go to Peiyuan Zhang from UC San Diego, representing the TinyLlama team, for providing a great open pre-training framework and supporting series of acceleration operators.
These collective wisdom and unwavering support have been pivotal to our project.
This project is supported by SJTU SEIEE - ByteDance Large Language Model Joint Laboratory, Shanghai Artificial Intelligence Laboratory.

%% file: appendix.tex
\appendix

\part{}
\section*{\centering \LARGE{Appendix}}
\mtcsettitle{parttoc}{}
\parttoc

\clearpage

\section{\method Implementation Details}

\subsection{Supervised Fine-tuning Data Collection}~\label{app:subsec-seed-data-collection}

In this section, we elaborate the detailed prompts used to generated the SFT data for model adaptation.
In principle, We apply the same prompts for general domain corpora~(including C4~\citep{raffel2020exploring}, \redpj~\citep{together2023redpajama}, \fineweb~\citep{penedo2024fineweb}) and mathematical corpus~(\owm~\citep{paster2023openwebmath}).
And all seed data is randomly sampled from the raw corpora.

\paragraph{Document-Level Programming}
We apply two zero-shot scoring prompts to evaluate and assign a combined score to each web document before synthesizing the \texttt{(doc, program)} pair.
One of the prompts is the same as the one used in FineWeb-Edu, which is a prompt to let the model decide the educational score.
Additionally in \method, we add a new format scoring prompt, focusing on the format and structure of the document.
Both prompts follow the additive style proposed by \citet{yuan2024self}.
Given these prompts, the language models generate short critiques and assign a score between $0$ and $5$.

In FineWeb-Edu, documents are retained only if the educational score (Edu Score) is greater than $2$. However, this approach is too aggressive when attempting to preserve a larger portion of the tokens.
For instance, FineWeb-Edu retains only 1.3 trillion tokens out of the original 15 trillion in the FineWeb corpus.
To recall more documents, we relax the filtering criteria by incorporating the format score as follows:

\begin{equation}
    \text{Filtering Criteria $=$}
    \begin{cases}
        \text{Edu Score} \geq 3, & \text{keep document;} \\
        \text{Edu Score} = 2 \text{ and } \text{Format Score} \geq 4, & \text{keep document;} \\
        \text{Edu Score} < 2, & \text{drop document.}
    \end{cases}
\end{equation}

Finally, we use \llamaiii to annotate $51$K data, splitting $5$K for validation~\footnote{In the earlier stage of experiments, we found that a dataset of thousands of data points (i.e., 5K) is also sufficient to equip the model with the ``programming'' abilities. This generally holds true for both document-level and chunk-level programming tasks. Scaling the dataset size could enhance the model's robustness across various documents.}.

The \fineweb-Edu prompt and our format scoring prompts are presented in Figure~\ref{app-fig:app-df-edu-prompt}.

\paragraph{Chunk-level Programming}
We apply chunk-level programming for more fine-grained operations.
We find three very popular patterns that keep occurring in all corpus: (1) menu, navigation bars at the top of the document; (2) button, html elements, links; (3) footers.

In general, LLMs work well given within $5$ few-shot examples. 
But to generate these program snippets more accurately, we apply few-shot prompting with \llamaiii for each type of noise.
We merge these programs aiming to clean different types of noises, perform some grammar checking, and make them the final data for training and validation during the chunk-level refining stage.
The annotated source comes from the same seed document used in the previous document filtering stage, accumulating to about $57$K data, of which $5$K is split as validation.

After the release of \textsc{Llama-3.1-405B-Instruct}, We also try to use only one prompt aiming to remove all the noises.
However, we find such practices lead to aggressive removal of the original document, often making the document less coherent.
Finally, we decide to only keep the head part and tail part of the program generated by \textsc{Llama-3.1-405B-Instruct}, which is previously mentioned in FinGPT~\citep{luukkonen2023fingpt}, and merge with the previous programs generated by \llamaiii.

The few-shot prompts used to generate program snippets are presented in Figure~\ref{app-fig:app-lf-nav-prompt}, Figure~\ref{app-fig:app-lf-url-prompt} and Figure~\ref{app-fig:app-lf-footer-prompt}.

\paragraph{Comparison with FineWeb-Edu's Approach}

Compared with the recently released FineWeb-Edu, which also uses model-based scoring by applying a BERT model to evaluate documents, we find that our relaxed design retains more tokens without compromising overall data quality. Specifically, FineWeb-Edu retains about $1.3$ trillion tokens out of a $15$ trillion token corpus (less than $9\%$), while \method curation typically keeps $23\%$ to $28\%$, providing up to $\mathbf{3\times}$ more unique tokens for training.

Moreover, we conducted a preliminary study by training $0.7$ billion parameter models on these data. We found that models trained on our curated data achieved similar downstream performance, as shown in Table~\ref{tab:compare-edu}. Therefore, we believe our current strategy is more suitable for large scale pre-training, as it is capable of retaining more tokens while maintaining very high data quality.

\begin{table}[htbp]
  \centering
  \caption{Comparing \fineweb-Edu with our strategy on \tlms.}
  \resizebox{0.98\textwidth}{!}{
    \begin{tabular}{cc|cccccccccc|cc}
    \toprule
    \begin{tabular}[c]{@{}c@{}}\textbf{Methods}\end{tabular} & 
    \textbf{Kept Ratio} &
    \textbf{ARC-C} & \textbf{ARC-E} & \textbf{CSQA} & \textbf{HellaSwag} & \textbf{MMLU} & \textbf{OBQA} & \textbf{PiQA} & \textbf{SIQA} & \textbf{WinoG} & \textbf{SciQ} & \textbf{AVG} & \textbf{\#Win} \\
    \midrule
    \fineweb-Edu  & 8.6\% & 30.3  & 58.7  & 29.0    & 42.0    & 30.4 & 31.8  & 67.7  & 38.1  & 50.4  & 73.3  & 45.2 & 5/10 \\
    \fineweb-\method & \textbf{28.0\%}   & 27.7  & 55.7  & 30.4  & 44.2  & 29.5 & 31.0    & 68.8  & 39.3  & 52.2  & 72.8  & 45.2 & 5/10 \\
    \bottomrule
    \end{tabular}%
    }
  \label{tab:compare-edu}%
\end{table}%

\begin{figure}[!t]
\input{figures/app-df-edu-prompts}
\caption{Edu scoring prompts used in \fineweb~\citep{penedo2024fineweb} and newly proposed ``format scoring'' prompts for \method.}
\label{app-fig:app-df-edu-prompt}
\end{figure}

\begin{figure}[!t]
\input{figures/app-lf-nav-prompts}
\caption{Few-shot navigation bar removal prompts.}
\label{app-fig:app-lf-nav-prompt}
\end{figure}

\begin{figure}[!t]
\input{figures/app-lf-url-prompts}
\caption{Few-shot URL removal prompts.}
\label{app-fig:app-lf-url-prompt}
\end{figure}

\begin{figure}[!t]
\input{figures/app-lf-footer-prompts}
\caption{Few-shot footer removal prompts.}
\label{app-fig:app-lf-footer-prompt}
\end{figure}

\clearpage
\newpage
\subsection{Supervised Fine-tuning Details}
\label{app-subsec:sft-training-details}
\paragraph{Training Parameters}
We use llama-factory~\citep{zheng2024llamafactory} as our main code base for Adaptation Stage. We apply full paraemter supervised fine-tuning on our base models: we train on the whole seed dataset for $3$ to $5$ epochs, with batch size as 64, and cosine learning rate schedular~(lr from 1e-5 $\rightarrow$ 1e-6).
Also, we find that base model convergent quite fast on these tasks, thus we do not apply a further tuning over hyper-parameters, and keep the same training configurations for all the adaptation tasks.

\subsection{Evaluation Metrics for \method Refining Tasks}
\label{app-subsec:sft-eval-metrics}

\paragraph{Document-level refining Task}
The document filtering task is indeed equal to a binary classification problem, where documents are classified as either to be kept ($1$) or dropped ($0$).
We evaluate the performance using the F1 score, calculated as follows:
\begin{equation}
    \text{F1} = 2 \cdot \frac{\text{Precision} \cdot \text{Recall}}{\text{Precision} + \text{Recall}}
\end{equation}

where:

\begin{equation}
    \text{Precision} = \frac{\text{TP}}{\text{TP} + \text{FP}}, \quad
    \text{Recall} = \frac{\text{TP}}{\text{TP} + \text{FN}}
\end{equation}
The F1 score ranges from 0 to 1 and we assume higher F1 score indicates better classification performance.

\paragraph{Chunk-level Refining Task}
This task actually contains two parts: line removal and string normalization.
However, we find it is rather hard to evaluate the normalization task, so we use the line removal accuracy to reflect the refining performance.
We propose a line-wise F1 score metric:

The F1 score is computed by comparing the predicted noisy lines with the labeled noisy lines.
First, we extract the noisy line indexes from both the prediction and the label.
Then, we calculate the overlap between these two sets. The true positives (TP) are the number of lines in this overlap. 
False positives (FP) are the predicted noisy lines that are not in the labeled set, and false negatives (FN) are the labeled noisy lines that are not in the predicted set.
The calculation is actually simple:
\begin{alignat}{3}
&\text{TP (True Positives)}  &\quad&=&\quad& |\text{Predicted Noisy Lines} \cap \text{Actual Noisy Lines}| \\[0.5em]
&\text{FP (False Positives)} &\quad&=&\quad& |\text{Predicted Noisy Lines} \setminus \text{Actual Noisy Lines}| \\[0.5em]
&\text{FN (False Negatives)} &\quad&=&\quad& |\text{Actual Noisy Lines} \setminus \text{Predicted Noisy Lines}|
\end{alignat}

Then we use same calculation of F1 score mentioned before, i.e., $\text{F1}=\frac{2 \cdot \text{TP}}{2 \cdot \text{TP} + \text{FP} + \text{FN}}$.

\clearpage
\newpage
\subsection{\method Inference at Scale}
\label{app:subsec-inference-at-scale}

Thanks to the Datatrove project~\citep{penedo2024datatrove}, we are able to efficiently split, and load the whole corpus to each worker~(which normally equals to the number of the GPUs since small models do not require tensor parallelism).
We use the vllm~\citep{kwon2023efficientvllm} to perform large scale inference.

For chunk-wise programming, we will split the original document into several chunks, controlling the tokens of each chunk less than the context window. 
In practice, we normally replace token count process as a word count process for saving time, and control the window size as $1,500$.
The general algorithm is implemented as below:
\begin{algorithm}
\caption{Document Chunk Splitting Algorithm}
\begin{algorithmic}[1]
\Require Document $D$, context window size $W$
\Ensure Set of chunks $C$
\State $C \gets \emptyset$, $c \gets \emptyset$
\For{each line $l$ in $D$}
    \If{$\text{TokenCount}(c + l) \leq W$}
        \State $c \gets c + l$ \Comment{Add line to current chunk}
    \Else
        \If{$c \neq \emptyset$}
            \State $C \gets C \cup \{c\}$ \Comment{Save current chunk}
        \EndIf
        \If{$\text{TokenCount}(l) \leq W$}
            \State $c \gets l$ \Comment{Start new chunk}
        \Else
            \State $C \gets C \cup \{\text{FlagAsSkipped}(l)\}$ \Comment{Flag long line}
            \State $c \gets \emptyset$
        \EndIf
    \EndIf
\EndFor
\If{$c \neq \emptyset$}
    \State $C \gets C \cup \{c\}$ \Comment{Add the final chunk}
\EndIf
\State \Return $C$
\end{algorithmic}
\end{algorithm}

\clearpage
\newpage
\section{Pre-training Details}~\label{app:sec-training-details}
\vspace{-2mm}
\subsection{Training Infrastructure}
~\label{app:subsec-training-infra}
\vspace{-8mm}
\paragraph{Code Base}
Thanks to litgpt~\citep{litgpt-2023}, and TinyLlaMA~\citep{zhang2024tinyllama}, we are able to flexibly train all our base models.
We inherit several fused kernels from the TinyLlaMA, which is installed from the FlashAttention~\citep{dao2023flashattention2} including fused rotary positional embedding~(RoPE), layer normalization, and cross entropy loss to help saving memory.
We mainly apply FSDP strategy~\citep{metafsdp} to enable training larger scale models on multiple nodes.

\subsection{Pre-training Corpora}
~\label{app:sub-sec-pt-corpora}
Due to computing constraints and fair comparison purpose, we cannot exhaustively train over the whole corpora.
Thus, we apply random sampling for some of the pre-training corpora and make them as our pre-training data pools.
\begin{itemize}
    \item For \redpj, We randomly download $70$ file shards, obtaining a total data pool consisting about $500$B tokens, we evenly separate it into $8$ dumps, with each containing about $62.5$B tokens; due to computing constraints, we use only $1$ dump for verifying effectiveness~(Section~\ref{subsec:exp1-verify-walle}) and use $2$ dumps for scaling the training to $50$B tokens~(Section~\ref{subsec:exp2-beyond-size-and-corpora});
    \item For C4, we download the whole dataset, which contains about $198$B tokens;
    \item For \fineweb, we download the official $350$B sample\footnote{\url{https://huggingface.co/datasets/HuggingFaceFW/fineweb/tree/main/sample/350BT}};
    \item For \owm, we download the whole dataset.
\end{itemize}

We report the corpora details applied in each experiment in Table~\ref{app:tab-pt-corpora}.
\input{tables/app-pt-corpora}

\subsection{Model Configuration and Training Parameters}
~\label{app:subsec-model-and-train-config}

\paragraph{Model Architecture}
The models we used in general and continual pre-training are presented at Table~\ref{app-tab:tlm-architecture} with detailed architecture configuration. 

\paragraph{Training Hyperparameter Choice}
We primarily use a cosine learning rate scheduler and follow established settings used in~\citet{zhang2024tinyllama} and \citet{lin2024rho}.
The default configurations for each experiment can be found below and we elaborate full details in Table~\ref{tab:app-training-params}.
\begin{enumerate}[leftmargin=1.25em,itemindent=0.25em,labelsep=0.4em,itemsep=0.1em]
\item For general pre-training experiments, we set the learning rate to 5e-4 for \tlmxs and \tlms, 3e-4 for \tlmm; the maximum sequence lengths are uniformly set to 2048, and the global batch size is set to 2M tokens.
\item Additionally, we align all our hyper-parameters with those used in MATES~\citep{yu2024mates} to facilitate a direct comparison with their existing data selection methods, as previously shown in Table~\ref{tab:exp1-data-selection-baseline}.
In this case, we switch to the warmup-stable-decay~(WSD) learning rate scheduler~\citep{hu2024minicpm}, as implemented in MATES.
For fair comparison with baselines implemented in MATES, we apply the exact same WSD Schedular~\citep{hu2024minicpm}:

\begin{equation}
    lr(t) = \begin{cases}
    \frac{t}{W} \cdot \eta, & \text{if } t < W \\
    \eta, & \text{if } W \leq t < S \\
    0.5^{4 \cdot (t-S)/D} \cdot \eta, & \text{if } S \leq t < S + D
\end{cases}
\end{equation}
where $W$ equals to 2000, $S$ equals to 50000, $D$ equals to 200. 
\item For continual pre-training experiments, we set different hyperparameters for different base models, as shown in Table~\ref{tab:app-training-params}. We apply an early-stop mechanism mentioned in \internmath~\citep{ying2024internlm-math} for 7B model experiments.
We mainly refer these settings to the setup reported in Rho-1~\citep{lin2024rho} and \textsc{Llemma}~\citep{azerbayev2024llemma}. We do not use warmup in continual pre-training experiments.
\end{enumerate}

\input{tables/app-model-arch}
\input{tables/app-training-params}

\clearpage
\newpage
\section{Downstream Tasks Evaluation}~\label{app:sec-eval-details}

\subsection{General Pre-training Evaluation} \label{app:general-pretrain-eval-config}

\paragraph{Lighteval Configurations}
We mainly borrow the evaluation benchmarks from the FineWeb's nine selected ``early signal'' tasks~\citep{penedo2024fineweb}, and use the implementation of lighteval~\citep{lighteval} to test all our base models.
We also introduce SciQ~\citep{welbl2017crowdsourcing-sciq} which is widely used in previous works and proved a good testbed~\citep{mehta2024openelm, wettig2024qurating}.
By default, we report the normalized zero-shot accuracy.
All the nine benchmarks at listed below:

\begin{itemize}
    \item ARC~\citep{clark2018arc}: including ARC-Easy~(\textbf{ARC-E)} and ARC-Challenge~(\textbf{ARC-C})
    \item CommonSense QA~\citep{talmor-etal-2019-commonsenseqa}~(\textbf{CSQA})
    \item HellaSwag~\citep{zellers2019hellaswag}
    \item MMLU~\citep{hendrycks2021measuring}
    \item OpenBook QA~\citep{mihaylov-etal-2018-suit-openbookqa}~(\textbf{OBQA})
    \item PIQA~\citep{bisk2020piqa}
    \item SocialIQA~\citep{sap2019socialiqa}~(\textbf{SIQA})
    \item WinoGrande~\citep{sakaguchi2021winogrande}~(\textbf{WinoG})
    \item SciQ~\citep{welbl2017crowdsourcing-sciq}
\end{itemize}

We follow the lighteval's configuration, which randomly picks $1,000$ samples for each dataset (for MMLU, it selects $1,000$ samples for each of the $57$ subsets), and report the normalized accuracy.
These average performance is calculated over the nine benchmarks, where ARC-C and ARC-E are considered as two separate benchmarks, and MMLU is treated as a single benchmark.
This approach differs slightly from the aggregation score calculation in FineWeb, as we believe MMLU's performance is relatively unstable, and we aim to give equal weight to all benchmarks, preventing MMLU from becoming a dominant factor.
For the original lighteval scores, please refer to the \S\ref{subsec:full-lighteval-performance}, where we include a dynamic result curve that clearly illustrates the fluctuations in each benchmark.

We present zero shot evaluation results in Table~\ref{tab:exp1-zero-shot-avg-performance}, Figure~\ref{fig:exp1-bench}.

\paragraph{LM-Eval Harness Configurations} 
We also include the lm-evel-harness~\citep{biderman2024lessons} for zero-shot and few-shot performance, for fair comparison with different data selection methods including DSIR~\citep{xie2023data},DsDm~\citep{engstrom2024dsdm}, Qurating~\citep{wettig2024qurating} MATES~\citep{yu2024mates}.
Similar to lighteval configuration, we include:
\begin{itemize}
    \item ARC: including ARC-E and ARC-C
    \item HellaSwag
    \item LogiQA~\citep{liu2020logiqa}
    \item OpenBook QA (OBQA)
    \item PIQA
    \item WinoGrande~({WinoG})
    \item SciQ
\end{itemize}

We exclude the BoolQ~\citep{clark2019boolq} tasks from MATES~\citep{yu2024mates}, leaving eight tasks in total. This decision was made because we observed that the BoolQ benchmark performance exhibited severe fluctuations and showed a notable declining trend in the early stages. Therefore, we decided to exclude it from our evaluation set.
Such trend is also observed earlier in the OpenELM work~\citep{mehta2024openelm}.
We report both zero-shot and two-shot performance. If the metrics include \textit{normalized accuracy}, we use that measure; otherwise, we use \textit{accuracy}.

\subsection{Continual Pre-training Evaluation}\label{app:subsec-owm-eval-details}

We evaluate all benchmarks implemented in the math-eval-harness repository,\footnote{\url{https://github.com/ZubinGou/math-evaluation-harness}} including:

\begin{itemize}
    \item Math~(\textbf{MATH})~\citep{hendrycks2021measuring}
    \item GSM8K~\citep{cobbe2021gsm8k}
    \item SVAMP~\citep{patel2021nlp}
    \item ASDiv~\citep{miao2021diverse}
    \item MAWPS~\citep{koncel2016mawps}
    \item MathQA~(\textbf{MQA})~\citep{amini2019mathqa}
    \item TableMWP~(\textbf{TAB})~\citep{lu2023tabmwp}
    \item SAT MATH~\citep{azerbayev2024llemma} 
\end{itemize}

We use few-shot CoT prompting~\citep{wei2022chain} when evaluating these tasks, and report the accuracy of each task.

\clearpage
\newpage
\section{Full Evaluation Results}~\label{app:sec-full-eval-results}

\subsection{Detailed Performance on 10 Benchmarks in Sec~\ref{subsec:exp1-verify-walle}}~\label{subsec:full-lighteval-performance}

We report full evaluation results of checkpoints saved at different training steps in Section~\ref{subsec:exp1-verify-walle}.
We present the results for 0.7B models trained on data curated by different methods in Table~\ref{tab:exp1-full-results-walle-s-redpj-raw}, including models trained on raw data, rule-based filtered data, and data curated by \method.
\input{tables/app-tinyllama-700m-full-results}

\clearpage
\newpage
\subsection{Detailed Performance on 8 Benchmarks Used in Data Selection Experiments}
~\label{app:subsec-full-lmeval-performance}

The full benchmark performance used in data-selection method comparison experiments is presented in Table~\ref{tab:exp-1-full-results-mates}.
\input{tables/app-mates-full-results}

\begin{figure}[!h]
    \centering
    \includegraphics[width=0.99\textwidth]{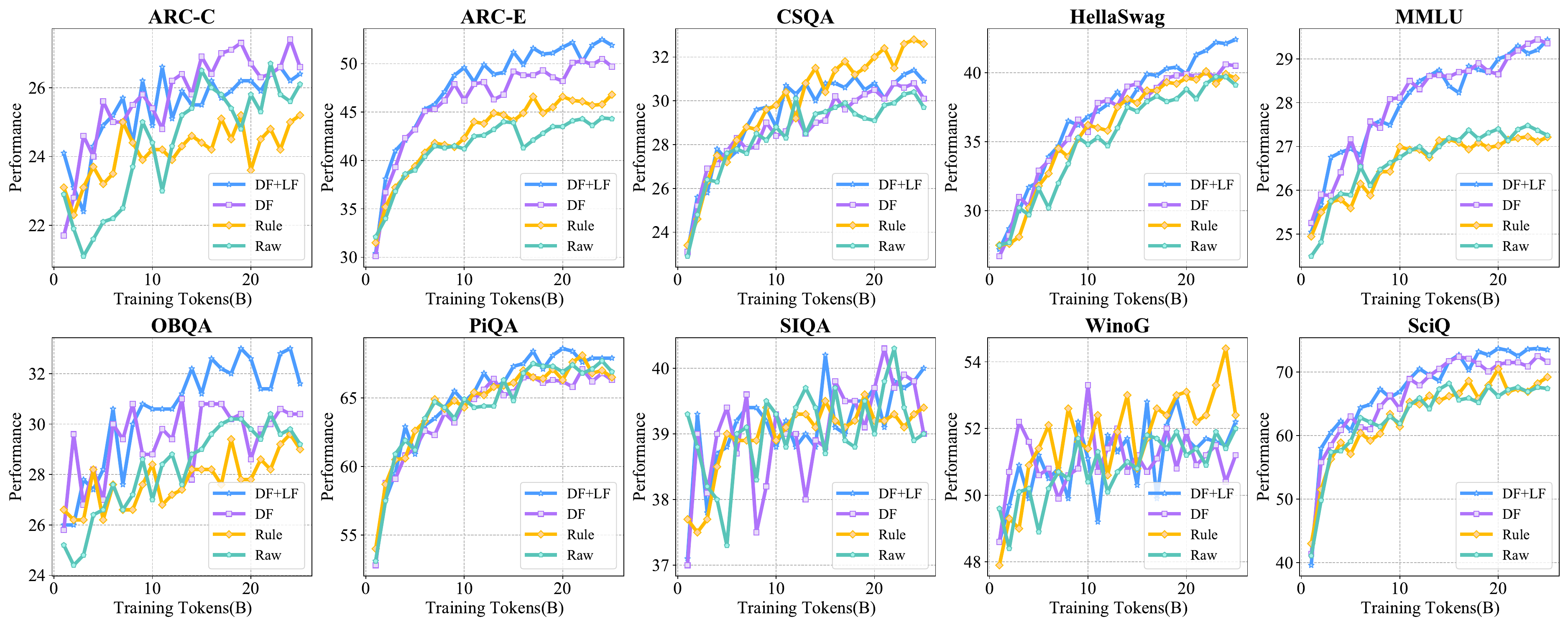}
    \caption{Visualization of dynamic performance on ten benchmarks.}
    \label{fig:exp1-full-dynamic}
\end{figure}

\clearpage
\newpage
\subsection{Detailed Performance in Sec~\ref{subsec:exp2-beyond-size-and-corpora}}

In \S~\ref{subsec:exp2-beyond-size-and-corpora}, we test \method's effectiveness using different sizes of refining models, and also train a series of models by using these curated data.
We report these detailed results in Table~\ref{tab:exp-2-full-results-model-size-xs}, Table~\ref{tab:exp-2-full-results-model-size-s} and Table~\ref{tab:exp-2-full-results-model-size-m}.

\input{tables/app-exp2-modelsizes-full}

We also further scale \method to other two pre-training corpora, C4 and \fineweb. We also scale our training to about $50$B tokens, and directly compare with existing well-trained models developed by different research groups.
We report our detailed results in Table~\ref{tab:exp-2-full-results-50B-redpj}, Table~\ref{tab:exp-2-full-results-50B-c4} and Table~\ref{tab:exp-2-full-results-50B-fw}. We also present other models' results in Table~\ref{tab:exp-2-full-results-existing-models}.
\input{tables/app-exp2-pt-corpora-full}

\clearpage
\newpage
\subsection{Evaluation Results of Continual Pre-training in Sec~\ref{subsec:exp3-math-cpt}}~\label{app:subsec:owm}

We provide full ablation results for each base model, as shown in Table~\ref{tab:exp3-owm-cpt-full-ablation-results}. We can observe that \method-D+C consistently improves average performance over \method-D across various base models. Although the performance gain from \method-D+C compared to \method-D is less pronounced than the improvement of \method-D over continual pre-training on raw \owm, this is both understandable and expected. \method-D+C does not significantly reduce the token count beyond the reductions achieved by \method-D alone. Given the scale of the \owm corpus, a more aggressive token removal strategy could potentially diminish the diversity of unique tokens below the threshold necessary for robust pre-training. This observation underscores the delicate balance between data refinement and maintaining sufficient linguistic variety for effective language model training, particularly when working with limited-scale corpora.

\input{tables/exp-3-full-ablation-results}

Besides, we report the detailed dynamic evaluation results of our continual pre-training experiments on \owm:

\begin{itemize}
    \item Tables~\ref{tab:exp3-tlm-1.1b-owm-cpt-raw}, \ref{tab:exp3-tlm-1.1b-owm-cpt-rule}, \ref{tab:exp3-tlm-1.1b-owm-cpt-df}, and \ref{tab:exp3-tlm-1.1b-owm-cpt-df+lc} present the evaluation results for \tinylm{}.
    \item Tables~\ref{tab:exp3-llama-2-7b-owm-cpt-raw}, \ref{tab:exp3-llama-2-7b-owm-cpt-df}, and \ref{tab:exp3-llama-2-7b-owm-cpt-lc} present the evaluation results for \llamaii.
    \item Tables~\ref{tab:exp3-codellama-7b-owm-cpt-raw}, \ref{tab:exp3-codellama-7b-owm-cpt-df}, \ref{tab:exp3-codellama-7b-owm-cpt-lc} present the evaluation results for \codellama.
    \item Tables~\ref{tab:exp3-mistral-7b-owm-cpt-raw}, \ref{tab:exp3-mistral-7b-owm-cpt-df}, and \ref{tab:exp3-mistral-7b-owm-cpt-lc} show the evaluation results for \mistral{}-7B.
\end{itemize}

\input{tables/exp-3-tlm-full-results}
\newpage
\input{tables/exp-3-llama2-full-results}
\newpage
\input{tables/exp-3-codellama-full-results}
\newpage
\input{tables/exp-3-mistral-full-df-results}

\newpage
\section{Analysis}~\label{app:sec-analysis}

\subsection{Case Studies}~\label{app:subsec-case-study}

We provide several cases to qualitatively illustrate the refinement effect of \method, as shown in Tables~\ref{tab:case_study_redpj}-\ref{tab:case_study_owm}. For the general domain, using \redpj as an example, we observe that \method can drop low-information documents, remove meaningless content such as navigation bars, and replace URL links (see Table~\ref{tab:case_study_redpj}). In the mathematics domain, \method demonstrates the ability to eliminate documents with minimal relevance to mathematical reasoning and remove less important elements like functional buttons (see Table~\ref{tab:case_study_owm}). These refinements enhance the quality and relevance of the processed data across different domains.

\input{tables/case_study_redpj_df_lf_all_programs}

\input{tables/case_study_owm_df_lf_all_programs}

\subsection{Error Analysis}~\label{app:subsec-error-analysis}
As shown in Table~\ref{tab:failure_ratio}, the failure ratio across both refining stages (document-level and chunk-level) and domains (General and Math) is remarkably low (\m{<0.5}\%). This demonstrates that ProX's refining tasks are well-suited for small models. Specifically, for the General domain, failure ratios are \m{0.04}\% for document-level and \m{0.36}\% for chunk-level refining, with an average of \m{3.7} function calls per program in the chunk-level stage. For the Math domain, these ratios are \m{0.06}\% and \m{0.11}\%, respectively, with an average complexity of \m{2.7} function calls at the chunk-level stage.

Despite the low failure rates, we observed two prevalent failure cases in ProX's programs:
\begin{enumerate}
    \item \textbf{Repeated output or empty output:} 
    This occurs when a program inadvertently generates duplicate outputs or fails to produce any meaningful results. Such failures are typically linked to improper loop conditions or insufficient constraints in processing logic.
    \item \textbf{Non-existent target removal:}
    In some cases, ProX's programs attempt to remove a string or line that does not exist in the input data. This leads to incomplete execution or errors in the program output, particularly in datasets with irregular formats or unexpected variations.
\end{enumerate}
\begin{table}[h!]
    \centering
    \caption{{Failure ratio and average complexity (function calls) for ProX refining stages across domains.}}
    \label{tab:failure_ratio}
    \resizebox{.9\textwidth}{!}{
    \begin{tabular}{lccc}
        \toprule
        \textbf{Domain} & \textbf{Failure Ratio (doc-level)} & \textbf{Failure Ratio (chunk-level)} & \textbf{Complexity (AVG. function calls)} \\
        \midrule
        General Domain & 0.04\% & 0.36\% & 3.7 \\
        Math Domain & 0.06\% & 0.11\% & 2.7 \\
        \bottomrule
    \end{tabular}
    }
\end{table}

As shown in Table~\ref{tab:case_study_redpj_failure_cases},
we present two failure cases to illustrate instances of repeated output and non-existent target strings.

\input{tables/case_study_redpj_failure_cases}

\clearpage
\newpage
\subsection{Computing Overhead Analysis}~\label{app:subsec-flops-analysis}

According to \citet{kaplan2020scaling}, both training and inference computational FLOPs for Transformer-based Language Models (denoted as $C_{\text{train}}$ and $C_{\text{inference}}$) can be approximated as the product of model parameters~(non-embedding parameter) $N$ and the number of tokens $D$.
This can be expressed as:
\begin{equation}
    C_{\text{train}} \approx 6\cdot{N} D_{\text{train}},
\end{equation}
\begin{equation}
    C_{\text{inference}} \approx 2 \cdot{N} \left( D_{\text{prefill}} + D_{\text{decode}} \right).
\end{equation}

In \method, we go through two data refining stages before final training, which incurs additional inference-time computational FLOPs. Suppose the refining model parameter for each stage is denoted as $N_\text{refine}$, and the raw data size in tokens is $D_\text{raw}$. 

For the first document-level stage, the computational cost can be approximated as:
\begin{equation}
    C_\text{doc} \approx 2 \cdot N_\text{refine} \left( D_\text{raw} + D_\text{output} \right) \approx 2 \cdot N_\text{refine} D_\text{raw},
{~~~~(\text{suppose } D_\text{output} \ll D_\text{raw})}
\end{equation}

resulting in a new pool of data sized $D_\text{doc}$.

Similarly, for the second chunk-level stage, the computational cost is:

\begin{equation}
    C_\text{chunk} \approx 2 \cdot N_\text{r} \left( D_\text{doc} + D_\text{output} \right) \approx 2 \cdot N_\text{r} D_\text{doc},
{~~~~(\text{suppose } D_\text{output} \ll D_\text{doc})}
\end{equation}

which produces the final refined data size of $D_\text{ProX}$.

Thus, the total computational overhead for \method can be calculated as the sum of the two stages:

\begin{equation}
    C_\text{\method} = C_\text{doc} + C_\text{chunk} \approx 2 \cdot N_\text{doc\_refine} D_\text{raw} + 2 \cdot N_\text{chunk\_refine} D_\text{doc}.
\end{equation}

In general, we use refining models with same sizes, so the final inference overhead can be estimated as
\begin{equation}
    C_\text{\method} \approx 2 \cdot N_\text{refine}(D_\text{raw}+D_\text{doc}).
\end{equation}

Additionally, we omit the FLOPs for fine-tuning since they are negligible compared to the large-scale pre-training and inference FLOPs.

%% file: figures/app-df-edu-prompts.tex
\begin{tcolorbox}[
colback=lightProxYellow!10,
colframe=lightProxYellow,
left=2mm, right=2mm,title=\textcolor{black}{\textbf{Edu Scoring Prompts~\citep{penedo2024fineweb}}}]

\begin{tiny}
Below is an extract from a web page. Evaluate whether the page has a high educational value and could be useful in an educational setting for teaching from primary school to grade school levels using the additive 5-point scoring system described below. Points are accumulated based on the satisfaction of each criterion:
\\
\\
- Add 1 point if the extract provides some basic information relevant to educational topics, even if it includes some irrelevant or non-academic content like advertisements and promotional material.
- Add another point if the extract addresses certain elements pertinent to education but does not align closely with educational standards. It might mix educational content with non-educational material, offering a superficial overview of potentially useful topics, or presenting information in a disorganized manner and incoherent writing style.
- Award a third point if the extract is appropriate for educational use and introduces key concepts relevant to school curricula. It is coherent though it may not be comprehensive or could include some extraneous information. It may resemble an introductory section of a textbook or a basic tutorial that is suitable for learning but has notable limitations like treating concepts that are too complex for grade school students. 

- Grant a fourth point if the extract highly relevant and beneficial for educational purposes for a level not higher than grade school, exhibiting a clear and consistent writing style. It could be similar to a chapter from a textbook or a tutorial, offering substantial educational content, including exercises and solutions, with minimal irrelevant information, and the concepts aren't too advanced for grade school students. The content is coherent, focused, and valuable for structured learning.

- Bestow a fifth point if the extract is outstanding in its educational value, perfectly suited for teaching either at primary school or grade school. It follows detailed reasoning, the writing style is easy to follow and offers profound and thorough insights into the subject matter, devoid of any non-educational or complex content.

The extract:

<EXAMPLE>.

After examining the extract: 

- Briefly justify your total score, up to 100 words.

- Conclude with the score using the format: ``Educational score:  <total points>''
\end{tiny}

\end{tcolorbox}

\begin{tcolorbox}[
colback=lightProxYellow!10,
colframe=lightProxYellow,
left=2mm, right=2mm,title=\textcolor{black}{\textbf{Format Scoring Prompts}}]

\begin{tiny}
Evaluate the provided web content extraction sample. Points are accumulated based on the satisfaction of each criterion:
\\
\\
0. Start with 0 points.

1. Add 1 point if the extract contains some readable content, even if it includes a significant amount of HTML tags, navigation elements, or other web page artifacts. The main content should be identifiable, albeit mixed with noise.

2. Add another point if the extract shows signs of basic cleaning. Most obvious HTML tags have been removed, though some may remain. The text structure begins to emerge, but non-content elements (e.g., footer links, button text) may still be present. The writing style may be disjointed due to remnants of page structure.

3. Award a third point if the extract is largely cleaned of HTML and most non-content elements. The main body of the content is intact and coherent. Some extraneous information (e.g., isolated URLs, timestamps, image alt text) may persist, but doesn't significantly impede readability. The extract resembles a rough draft of the original content.

4. Grant a fourth point if the extract is highly refined, with clear paragraph structure and formatting. Almost all HTML tags and non-content elements have been eliminated. Minimal noise remains. The content flows well and reads like a near-final draft, with consistent formatting and style.

5. Bestow a fifth point if the extraction is flawless. The content is entirely clean, preserving the original structure (paragraphs, headings, lists) without any HTML tags or web page elements. No extraneous information is present. The extract reads as if it were a professionally edited document, perfectly capturing the original content.

The extract:

<EXAMPLE>.

After examining the extract:

- Briefly justify your total score, up to 100 words.

- Conclude with the score using the format: "Extraction Quality Score: <total points>"
\end{tiny}

\end{tcolorbox}

%% file: figures/app-lf-nav-prompts.tex
\begin{tcolorbox}[
colback=lightProxYellow!10,
colframe=lightProxYellow,
left=2mm, right=2mm,title=\textcolor{black}{\textbf{Navigation Removal Prompts}}]

\begin{tiny}
You're tasked with generating Python programs to clean web text strings by removing navigation bars. The web text will be presented with line numbers starting from \verb|`[000]`|. Your task is to use the following pre-defined functions to clean the text:

\begin{verbatim}
```python
\end{verbatim}
\begin{lstlisting}
def untouch_doc():
    """leave the clean doc untouched, for tagging clean and high quality doc."""

def remove_lines(start: int, end: int):
    """remove noisy lines from `start` until `end`, including `end`."""
\end{lstlisting}

\begin{verbatim}
```
\end{verbatim}

Your goal is to identify navigation bars or menu items at the beginning of the text and remove them using the \verb|`remove_lines()`| function. If the text doesn't contain a navigation bar or menu items, use the \verb|`untouch_doc()`| function to indicate that no cleaning is necessary. If the line contains other text other than navigation, also call \verb|`untouch_doc`| to escape overkilling.\\
Here are some examples to guide you:\\
Example 1:\par
\begin{verbatim}
[doc]
[000] Home | Products | About Us | Contact
[001] Welcome to our website
[002] Here's our main content...
[/doc]
Program:
```python
remove_lines(start=0, end=0)
```
\end{verbatim}
Example 2:\par
\begin{verbatim}
[doc]
341 US 479 Hoffman v. United States
341 US 479 Hoffman v. United States 341 U.S. 479
95 L.Ed. 1118
HOFFMANv.UNITED STATES.
Mr. William A. Gray, Philadelphia, Pa., for petitioner.
Mr. John F. Davis, Washington, D.C., for respondent.
......
[/doc]
Program:
```python
untouch_doc()
```
\end{verbatim}
Example 3:\par
\begin{verbatim}
[doc]
[000]Police Search Tunbridge Wells House Over Human Remains Tip Off
[001]Posted: 16/04/2012 10:44 Updated: 16/04/2012 10:44	reddit stumble
[002]Crime, Body Buried In House, Buried Body, Buried Remains, Tip-Off, Uk News, Uk Police,
[003]Detectives are searching the gardens of a house following information that human remains may be
buried there.
[/doc]
Program:
```python
untouch_doc()
```
\end{verbatim}
Example 4:\par
\begin{verbatim}
[doc]
[000]Home > Bollywood News > Bollywood Stars clash on Indian TV	Bollywood Stars clash on Indian TV
[001]By Lekha Madhavan09:47 pm Betting big on the festive season, general entertainment channels (GECs) 
are launching celebrity-driven shows, but media buyers are concerned about the audience split that is set 
to happen.
[002]The fourth season of Bigg Boss on Colors is almost certain to clash with the fourth season of Kaun 
Banega Crorepati (KBC) on Sony Entertainment Television (SET) in the second week of October.
[003]Another big property, Master Chef, to be hosted by Akshay Kumar, on STAR Plus, is also expected to go 
on air in October. However, the channel is yet to disclose the launch date.
[004]Big-budget shows like these are often loss-making propositions for channels, as the operating cost is 
very high and advertisement revenues do not suffice to cover the cost.
[005]Source: IBNS
[/doc]
Program:
```python
untouch_doc()
```
\end{verbatim}
For each given web text, analyze the content and determine if there's a navigation bar or menu items at the beginning. If present, use \verb|`remove_lines()`| or \verb|`normalize()`| to remove them. If not, use \verb|`untouch_doc()`| to indicate that no cleaning is needed.

Example:
<EXAMPLE>.

After examining the web text:
- Briefly describe if the web extract contains navigation bar at the begining (10 lines). \par
- You must not mistakenly decide that title of the page is navigation bar and remove it. \par
- When the whole line is navigation bar, call \verb|`remove_lines`|; if the line contains other information, call \verb|`normalize`| to remove part of it. \par
- Give your program using the same format: \verb|```python[your code]```|
\end{tiny}
\end{tcolorbox}

%% file: figures/app-lf-url-prompts.tex
\begin{tcolorbox}[
colback=lightProxYellow!10,
colframe=lightProxYellow,
left=2mm, right=2mm,title=\textcolor{black}{\textbf{URL Removal Prompts}}]

\begin{tiny}
You're tasked with generating Python programs to clean web text strings by removing http lines. The web text will be presented with line numbers starting from \verb|`[000]`|. Your task is to use the following pre-defined functions to clean the text:

\begin{verbatim}
```python
\end{verbatim}
\begin{lstlisting}
def untouch_doc():
    """leave the clean doc untouched, for tagging clean and high quality doc."""

def remove_lines(start: int, end: int):
    """remove noisy lines from `start` until `end`, including `end`."""

def normalize(source_str: str, target_str: str=""):
    """turn noisy strings into normalized strings."""
\end{lstlisting}
\begin{verbatim}
```
\end{verbatim}

Your goal is to identify http links from the text and remove them using the \verb|`remove_lines()`| or \verb|`normalize()`| function. If the text doesn't contain http lines, use the \verb|`untouch_doc()`| function to indicate that no cleaning is necessary.\\
Here are some examples to guide you:\\
Example 1:\par
\begin{verbatim}
[doc]
[013] http://groups.google.com/group/toowoombalinuxLast
[014] Breaking News: Major Event Unfolds
[015] http://code.google.com/p/inxi/
[/doc]
Program:
```python
# the whole line-[013] is http, so remove the line-[013]
remove_lines(start=13, end=13)
# the whole line-[015] is http, so remove the line-[015]
remove_lines(start=15, end=15)
```
\end{verbatim}
Example 2:\par
\begin{verbatim}
[doc]
[000] The Impact of Climate Change on Global Ecosystems
[001] By Dr. Jane Smith
[002] Climate change continues to be a pressing issue...
[/doc]
Program:
```python
untouch_doc()
```
\end{verbatim}
Example 3:\par
\begin{verbatim}
[doc]
[021]Bow-wow
[022]http://groups.google.com/group/toowoombalinuxLast edited by Puppyt on Mon 06 Jun 2011, 00:23; edited 
1 time in total
[023]I would like to see something like Jitsi
[024]http://www.jitsi.org/. Plus some others incorporated into a puppy distro.
[/doc]
Program:
```python
# the http link in line 22 and line 24 comes with other text, so use normalize to ONLY remove the link 
without touching text.
normalize(source_str="http://groups.google.com/group/toowoombalinuxLast", target_str="")
normalize(source_str="http://www.jitsi.org/.", target_str="")
```
\end{verbatim}

For each given web text, analyze the content and determine if there's a navigation bar or menu items at the beginning. If present, use \verb|`remove_lines()`| or \verb|`normalize()`| to remove them. If not, use \verb|`untouch_doc()`| to indicate that no cleaning is needed.

Example:
<EXAMPLE>.

After examining the web text:
- do not remove text together with http. \par
- Briefly describe if the web extract contains http links; and make sure remove them will not influence the main content. \par
- Program only contain sequences of function callings and comments, no other codes. \par
- note line number starts with 0. make accurate annotations about line number. put the exact int line number of the given line. do not add 1 or minus 1. \par
- Give your program using the same format: \verb|```python[your code]```|

\end{tiny}
\end{tcolorbox}

%% file: figures/app-lf-footer-prompts.tex
\begin{tcolorbox}[
colback=lightProxYellow!10,
colframe=lightProxYellow,
left=2mm, right=2mm,title=\textcolor{black}{\textbf{Footer Removal Prompts}}]

\begin{tiny}
You're tasked with generating Python programs to clean web text strings by removing footer sections, references. The web text will be presented with line numbers starting from \verb|`[000]`|. Your task is to use the following pre-defined functions to clean the text:

\begin{verbatim}
```python
\end{verbatim}
\begin{lstlisting}
    
def untouch_doc():
    """leave the clean doc untouched, for tagging clean and high quality doc."""

def remove_lines(start: int, end: int):
    """remove noisy lines from `start` until `end`, including `end`."""

def normalize(source_str: str, target_str: str=""):
    """turn noisy strings into normalized strings."""
\end{lstlisting}

\begin{verbatim}
```
\end{verbatim}

Your goal is to identify footer sections from the text and remove them using the \verb|`remove_lines()`| function. Footers and references typically appear at the end of the text and may contain information such as copyright notices, contact details, or navigation links. If the text doesn't contain a footer section or any references, use the \verb|`untouch_doc()`| function to indicate that no cleaning is necessary. \\
Here are some examples to guide you:\\
Example 1:\par
\begin{verbatim}
[doc]
[013] In conclusion, the study demonstrates significant findings.
[014] © 2023 Research Institute. All rights reserved.
[015] Contact: info@research-institute.com
[016] Follow us on social media: @ResearchInst
[/doc]
Program:
```python
# Remove the footer section starting from line 14
remove_lines(start=14, end=16)
```
\end{verbatim}
Example 2:\par
\begin{verbatim}
[doc]
[000] The Impact of Climate Change on Global Ecosystems
[001] By Dr. Jane Smith
[002] Climate change continues to be a pressing issue...
[003] Further research is needed to fully understand its implications.
[/doc]
Program:
```python
untouch_doc()
```
\end{verbatim}
Example 3:\par
\begin{verbatim}
[doc]
[020] Thank you for reading our newsletter.
[021] Stay informed with our latest updates!
[022] ---
[023] Unsubscribe | Privacy Policy | Terms of Service
[024] NewsletterCo, 123 Main St, Anytown, USA
[/doc]
Program:
```python
# Remove the footer section starting from the divider
remove_lines(start=22, end=24)
```
\end{verbatim}

For each given web text, analyze the content and determine if there is a footer section or reference. If present, use \verb|`remove_lines()`| to remove it. If not, use \verb|`untouch_doc()`| to indicate that no cleaning is needed.

Example:
<EXAMPLE>.

After examining the web text:

- Briefly describe if the web extract contains a footer section or references; ensure that removing it will not influence the main content. If not, simply call \verb|`untouch_doc`|.

- The program should only contain sequences of function calls and comments, no other code.

- Note that line numbers start with 0. Make accurate annotations about line numbers. Put the exact int line number of the given line. Do not add 1 or subtract 1.

- Give your program using the same format: \verb|```python[your code]```|
\end{tiny}
\end{tcolorbox}

%% file: tables/app-pt-corpora.tex
\begin{table}[ht]
  \centering
  \caption{The detailed breakdown for pre-training corpora in all experiments.
  }
  \resizebox{0.98\linewidth}{!}{
    \begin{tabular}{c|c|c|l|c|c|c}
    \toprule
    \textbf{Section} & \textbf{Experiments} & \textbf{Source} & \textbf{Data Description} & \textbf{Corpora Size (B)} & \textbf{Effective Train Tokens (B)} & \textbf{Epoch} \\
    \midrule
    \multicolumn{1}{c|}{\multirow{4}[2]{*}{\begin{tabular}[c]{@{}c@{}}Section~\ref{subsec:exp1-verify-walle}\\ \end{tabular}}} &
    \multicolumn{1}{c|}{\multirow{4}[2]{*}{\begin{tabular}[c]{@{}c@{}}Table~\ref{tab:exp1-zero-shot-avg-performance}, Figure~\ref{fig:exp1-bench}\\ \end{tabular}}} &
    \multirow{4}[2]{*}{\redpj} & raw data size & 62.5  & \multirow{4}[2]{*}{26.2} & 0.42 \\
    &      &       & after rule-based filtering & 31.5  &       & 0.83 \\
    &      &       & after \method-D & 19.0  &       & 1.38 \\
    &      &       & after \method-D+C & 16.0  &       & 1.64 \\
    \midrule
    \multicolumn{1}{c|}{\multirow{3}[2]{*}{\begin{tabular}[c]{@{}c@{}}Section~\ref{subsec:exp1-verify-walle}\\ \end{tabular}}} &
    \multicolumn{1}{c|}{\multirow{3}[2]{*}{\begin{tabular}[c]{@{}c@{}}Table~\ref{tab:exp1-data-selection-baseline}\end{tabular}}} & \multirow{3}[2]{*}{C4} & random & - & \multirow{3}[2]{*}{26.2} & - \\
    &      &       & after \method-D & 41.5~(GPT-NeoX)  &       & 0.63 \\
    &      &       & other baselines & -     &       & - \\
    \midrule
    \multicolumn{1}{c|}{\multirow{4}[2]{*}{\begin{tabular}[c]{@{}c@{}}Section~\ref{subsec:exp2-beyond-size-and-corpora}\\ \end{tabular}}} &
    \multicolumn{1}{c|}{\multirow{4}[2]{*}{\begin{tabular}[c]{@{}c@{}}Figure~\ref{fig:exp2-walle-on-different-model-size}\end{tabular}}} & \multirow{4}[2]{*}{\redpj} & raw data size & 62.5  & \multirow{4}[2]{*}{26.2} & 0.42 \\
    &      &       & after \method-D+C (using \method-xs) & 14.5  &       & 1.80 \\
    &      &       & after \method-D+C (using \method-s) & 16.0  &       & 1.64 \\
    &      &       & after \method-D+C (using \method-m) & 18.0     &       & 1.46 \\
    \midrule
    \multicolumn{1}{c|}{\multirow{6}[6]{*}{\begin{tabular}[c]{@{}c@{}}Section~\ref{subsec:exp2-beyond-size-and-corpora}\\\end{tabular}}} & 
    \multicolumn{1}{c|}{\multirow{6}[6]{*}{\begin{tabular}[c]{@{}c@{}}Figure~\ref{fig:across-corpora-bar-chart}\end{tabular}}} & \multirow{2}[2]{*}{C4} & raw data size & 198.0 & \multirow{6}[6]{*}{52.4} & 0.53 \\
    &      &       & after \method-D+C (using \method-xs) & 44.5     &       & 1.18 \\
\cmidrule{3-5}\cmidrule{7-7}          
    &      & \multirow{2}[2]{*}{\redpj} & raw data size & 123.5     &       & 0.42 \\
    &      &       & after \method-D+C (using \method-xs) & 29     &       & 1.81 \\
\cmidrule{3-5}\cmidrule{7-7}          
    &      & \multirow{2}[2]{*}{\fineweb} & raw data size & 79.0     &       & 0.66 \\
    &      &       & after \method-D+C (using \method-xs) &   18.0   &       & 2.91 \\
    \midrule
    \multicolumn{1}{c|}{\multirow{4}[4]{*}{\begin{tabular}[c]{@{}c@{}}Section~\ref{subsec:exp3-math-cpt}\end{tabular}}} &
    \multicolumn{1}{c|}{\multirow{4}[4]{*}{\begin{tabular}[c]{@{}c@{}}Table~\ref{tab:exp3-owm-ct}, 1.1B model\end{tabular}}} & \multirow{4}[4]{*}{\owm} & raw data size & 15.0  & \multirow{4}[4]{*}{15.7} & 1.05 \\
    &      &       & after rule-based filtering & 6.5   &       & 2.40 \\
\cmidrule{4-5}\cmidrule{7-7}          
    &      &       & after \method-D & 5.5   &       & 2.85 \\
    &      &       & after \method-D+C & 4.7   &       & 3.49 \\
    \midrule
    \multicolumn{1}{c|}{\multirow{3}[2]{*}{\begin{tabular}[c]{@{}c@{}}Section~\ref{subsec:exp3-math-cpt}\\ \end{tabular}}} &
    \multicolumn{1}{c|}{\multirow{3}[2]{*}{\begin{tabular}[c]{@{}c@{}} Table~\ref{tab:exp3-owm-ct}, 7B model\end{tabular}}} 
    & \multirow{3}[2]{*}{\owm} & raw data size & 15.0  & \multirow{3}[2]{*}{10.5} & 0.70 \\
    &      &       & after \method-D & 5.5   &       & 1.91 \\
    &      &       & after \method-D+C & 4.7   &       & 2.23 \\
    \bottomrule
    \end{tabular}%
    }
  \label{app:tab-pt-corpora}%
\end{table}%

%% file: tables/app-model-arch.tex
\begin{table}[!t]
    \centering
    \caption{The details of the pre-training experiments' model architecture.}
    \label{app-tab:tlm-architecture}
    \resizebox{0.95\linewidth}{!}{%
    \begin{tabular}{lccccccc}
    \toprule
    \textbf{Model} & \textbf{Hidden Size} & \textbf{Intermediate Size} & \textbf{Context Len} & \textbf{Heads} & \textbf{Layers} & \textbf{Vocab Size} & \textbf{\# Params (w/o embed)} \\
    \midrule
    \multicolumn{8}{c}{Training From Scratch} \\
    \midrule
    \tlmxs  & 1,280  & 2,048  & 2,048 & 16 & 24 & 32,000   & 354,284,800 (313,324,800) \\
    \tlms   & 1,536  & 4,864  & 2,048 & 24 & 24 & 32,000  & 758,982,144 (709,830,144) \\
    \tlmm  & 2,048  & 8,192  & 2,048 & 32 & 24 & 32,000    & 1,741,785,088 (1,676,249,088) \\
    \midrule
    \pythia-410M & 1,024 & 4,096 & 1,024 & 16 & 24 & 50,304 & 405,334,016 (353,822,720)\\
    \pythia-1B & 2,048 & 8,192 & 1,024 & 8 & 16 & 50,304 & 1,011,781,632 (908,759,040) \\
    \midrule
    \multicolumn{8}{c}{Continual Pre-training} \\
    \midrule
    \tinylm & 2,048 & 5,632 & 2,048 & 32 & 22 & 32,000 & 1,100,048,384 (1,034,512,384)\\
    \textsc{Llama-2}-7B & 4,096 & 11,008 & 4,096 & 32 & 32 & 32,000 & 6,738,415,616 (6,607,343,616) \\
    \textsc{CodeLlama}-7B & 4,096 & 11,008 & 4,096 & 32 & 32 & 32,016 & 6,738,546,688 (6,607,409,152) \\
    \mistral-7B & 4,096 & 14,336 & 4,096 & 32/8~(GQA) & 32 & 32,000 &7,241,732,096 (7,110,660,096) \\
    \bottomrule  
    \end{tabular}%
    }
\end{table}

%% file: tables/app-training-params.tex
\begin{table}[!t]
  \centering
  \caption{Training hyper-parameters of all base models.}
  \resizebox{0.95\linewidth}{!}{%
    \begin{tabular}{lcccccccc}
    \toprule
    \textbf{Model} & 
    \begin{tabular}[c]{@{}c@{}}\textbf{Context}\\ \textbf{Length}\end{tabular} 
    & \textbf{Batch Size} 
    & \textbf{Max Steps} 
    & \begin{tabular}[c]{@{}c@{}}\textbf{Warmup}\\ \textbf{Steps}\end{tabular} 
    & \begin{tabular}[c]{@{}c@{}}\textbf{Weight}\\ \textbf{Decay}\end{tabular}
    & \textbf{Optimizer} 
    & \begin{tabular}[c]{@{}c@{}}\textbf{LR}\\ \textbf{Scheular}\end{tabular}
    & \textbf{LR} \\
    \midrule
    \multicolumn{9}{c}{Training from Scratch} \\
    \midrule
    \tlmxs & 1,024  & 2,048 & 12,500 & 500   & 0.1   & AdamW & cosine & 5e-4 $\rightarrow$ 5e-5 \\
    \tlms & 1,024  & 2,048 & 12,500 & 500   & 0.1   & AdamW & cosine & 5e-4 $\rightarrow$ 5e-6 \\
    \tlmm & 1,024  & 2,048 & 12,500/2,5000 & 500   & 0.1   & AdamW & cosine & 3e-4 $\rightarrow$ 3e-5 \\
    \midrule
    \pythia-410M & 512   & 1,024 & 50,200 & 2,000  & 0.1   & AdamW & WSD   & 1e-3 $\rightarrow$ 6.25e-5 \\
    \pythia-1B & 512   & 1,024 & 50,200 & 2,000  & 0.1   & AdamW & WSD   & 1e-3 $\rightarrow$ 6.25e-5 \\
    \midrule
    \midrule
    \multicolumn{9}{c}{Continual Pre-training} \\
    \midrule
    \tinylm & 2,048   & 1,024 & 7,500 & 0     & 0.1   & AdamW & cosine & 8e-5 $\rightarrow$ 8e-6 \\
    \textsc{Llama-2}-7B & 4096 & 256 & 15,000~{\tiny{(early stop at 10,000)}} & 0 & 0.1 & AdamW & cosine & 8e-5 $\rightarrow$ 8e-6 \\
    \textsc{CodeLlama}-7B & 4096 & 1024 &  3,750~(\tiny{early stop at 2,500}) & 0 & 0.1 & AdamW & cosine & 3e-4 $\rightarrow$ 3e-5 \\
    \mistral-7B & 4,096   & 256 & 15,000~{\tiny{(early stop at 10,000)}} & 0     & 0.1   & AdamW & cosine & 2e-5 $\rightarrow$ 2e-6 \\
    \bottomrule
    \end{tabular}%
    }
  \label{tab:app-training-params}%
\end{table}%

%% file: tables/app-tinyllama-700m-full-results.tex
\begin{table}[htbp]
  \centering
  \caption{Full evaluation results on \tlms.}
  \resizebox{0.95\textwidth}{!}{
    \begin{tabular}{c|cccccccccc|c}
    \toprule
    \begin{tabular}[c]{@{}c@{}}\textbf{Train}\\ \textbf{Steps}\end{tabular} & \textbf{ARC-C} & \textbf{ARC-E} & \textbf{CSQA} & \textbf{HellaSwag} & \textbf{MMLU} & \textbf{OBQA} & \textbf{PiQA} & \textbf{SIQA} & \textbf{WinoG} & \textbf{SciQ} & \textbf{AVG} \\
    \midrule
    \multicolumn{12}{c}{Raw Data} \\
    \midrule
    2500  & 22.1  & 39.0  & 27.6  & 31.6  & 25.9  & 26.6  & 61.2  & 37.3  & 48.9  & 59.1  & 37.9 \\
    5000  & 24.4  & 41.2  & 28.8  & 34.8  & 26.7  & 27.0  & 64.9  & 39.3  & 50.4  & 61.9  & 39.9 \\
    7500  & 26.5  & 43.9  & 29.5  & 37.2  & 27.2  & 29.0  & 64.8  & 38.7  & 50.8  & 68.2  & 41.6 \\
    10000 & 25.8  & 43.5  & 29.1  & 38.8  & 27.4  & 29.8  & 66.9  & 39.0  & 51.2  & 66.2  & 41.8 \\
    12500 & 26.1  & 44.3  & 29.7  & 39.1  & 27.3  & 29.2  & 66.9  & 39.0  & 52.0  & 67.4  & 42.1 \\
    \midrule
    \multicolumn{12}{c}{\gopher} \\
    \midrule
    2500  & 22.3  & 39.4  & 26.6  & 31.3  & 25.6  & 27.0  & 61.1  & 38.9  & 51.3  & 58.6  & 38.2 \\
    5000  & 25.1  & 41.4  & 29.8  & 34.3  & 26.4  & 27.2  & 64.5  & 39.6  & 52.1  & 62.9  & 40.3 \\
    7500  & 26.5  & 43.0  & 30.5  & 38.5  & 27.2  & 28.8  & 65.7  & 38.2  & 53.7  & 66.4  & 41.8 \\
    10000 & 26.2  & 44.2  & 31.8  & 39.2  & 27.5  & 29.4  & 66.6  & 38.9  & 51.3  & 68.2  & 42.3 \\
    12500 & 25.7  & 44.0  & 31.3  & 40.2  & 27.3  & 29.0  & 66.3  & 39.0  & 51.2  & 68.9  & 42.3 \\
    \midrule
    \multicolumn{12}{c}{C4} \\
    \midrule
    2500  & 22.6  & 40.6  & 28.8  & 31.3  & 26.2  & 27.4  & 61.7  & 39.3  & 51.2  & 57.1  & 38.6 \\
    5000  & 22.9  & 41.6  & 29.3  & 36.0  & 26.8  & 27.6  & 64.7  & 40.2  & 50.9  & 63.6  & 40.4 \\
    7500  & 24.2  & 44.2  & 29.5  & 39.2  & 27.2  & 28.4  & 66.2  & 40.9  & 51.6  & 63.8  & 41.5 \\
    10000 & 24.6  & 44.8  & 30.4  & 39.5  & 27.0  & 29.4  & 68.7  & 40.9  & 51.7  & 63.9  & 42.1 \\
    12500 & 25.0  & 46.0  & 31.0  & 40.5  & 27.1  & 29.2  & 68.5  & 40.5  & 51.7  & 66.6  & 42.6 \\
    \midrule
    \multicolumn{12}{c}{\fineweb{}} \\
    \midrule
    2500  & 23.2  & 39.4  & 27.2  & 31.8  & 25.6  & 26.2  & 62.6  & 39.0  & 51.4  & 57.1  & 38.3 \\
    5000  & 24.2  & 42.3  & 29.8  & 36.2  & 27.0  & 28.4  & 64.3  & 38.9  & 51.4  & 61.4  & 40.4 \\
    7500  & 24.4  & 44.1  & 30.4  & 37.8  & 27.2  & 28.2  & 66.1  & 39.5  & 50.8  & 66.2  & 41.5 \\
    10000 & 23.6  & 46.6  & 32.0  & 39.6  & 27.0  & 27.8  & 66.3  & 39.2  & 53.1  & 70.5  & 42.6 \\
    12500 & 25.2  & 46.8  & 32.6  & 39.6  & 27.2  & 29.0  & 66.5  & 39.4  & 52.4  & 69.2  & 42.8 \\
    \midrule
    \multicolumn{12}{c}{\gopher + C4 + \fineweb} \\
    \midrule
    2500  & 23.6  & 39.3  & 27.6  & 32.1  & 25.8  & 26.0  & 61.7  & 39.8  & 50.9  & 55.4  & 38.2 \\
    5000  & 23.9  & 40.9  & 29.0  & 36.2  & 26.9  & 26.8  & 65.3  & 39.3  & 52.7  & 62.4  & 40.3 \\
    7500  & 25.6  & 42.2  & 30.7  & 39.7  & 27.0  & 28.4  & 66.0  & 40.2  & 51.8  & 60.9  & 41.2 \\
    10000 & 25.8  & 43.3  & 30.8  & 41.4  & 27.5  & 29.8  & 66.9  & 39.5  & 51.8  & 63.1  & 42.0 \\
    12500 & 25.0  & 43.9  & 30.0  & 41.9  & 27.5  & 31.0  & 67.0  & 39.9  & 51.9  & 65.3  & 42.3 \\
    \midrule
    \multicolumn{12}{c}{\method-D} \\
    \midrule
    2500  & 25.6  & 43.2  & 27.7  & 32.9  & 27.2  & 27.0  & 61.3  & 39.4  & 50.6  & 63.0  & 39.8 \\
    5000  & 25.4  & 46.2  & 28.4  & 35.7  & 28.1  & 28.8  & 64.7  & 39.3  & 53.3  & 64.2  & 41.4 \\
    7500  & 26.9  & 49.2  & 29.1  & 39.2  & 28.6  & 30.8  & 65.4  & 38.8  & 51.2  & 71.7  & 43.1 \\
    10000 & 26.7  & 48.2  & 30.5  & 39.9  & 28.6  & 28.6  & 66.2  & 39.7  & 51.9  & 71.2  & 43.2 \\
    12500 & 26.6  & 49.7  & 30.1  & 40.5  & 29.4  & 30.4  & 66.3  & 39.0  & 51.2  & 71.6  & 43.5 \\
    \midrule
    \multicolumn{12}{c}{\method-D+C} \\
    \midrule
    2500  & 24.9  & 43.4  & 27.3  & 32.1  & 26.9  & 28.2  & 60.9  & 38.8  & 51.2  & 60.8  & 39.5 \\
    5000  & 24.9  & 49.6  & 28.8  & 36.8  & 27.9  & 30.6  & 64.7  & 38.8  & 51.1  & 66.9  & 42.0 \\
    7500  & 25.5  & 51.2  & 30.8  & 38.8  & 28.4  & 31.2  & 67.3  & 40.2  & 50.3  & 71.7  & 43.5 \\
    10000 & 26.2  & 51.7  & 30.8  & 39.9  & 29.0  & 32.6  & 68.6  & 39.7  & 51.7  & 73.7  & 44.4 \\
    12500 & 26.4  & 51.9  & 30.9  & 42.4  & 29.4  & 31.6  & 67.9  & 40.0  & 52.2  & 73.5  & 44.6 \\
    \bottomrule
    \end{tabular}%
    }
  \label{tab:exp1-full-results-walle-s-redpj-raw}%
\end{table}%

%% file: tables/app-mates-full-results.tex
\begin{table}[htbp]
  \centering
  \small

  \caption{Detailed evaluation results for different data selection methods.}
  \resizebox{.95\textwidth}{!}{
    \begin{tabular}{c|cccccccc|c}
    \toprule
    \textbf{Method} & \textbf{ARC-C} & \textbf{ARC-E} & \textbf{HellaSwag} & \textbf{LogiQA} & \textbf{OBQA} & \textbf{PIQA} & \textbf{WinoGrande} & \textbf{SciQ} & \textbf{AVG} \\
    \midrule
          \multicolumn{10}{c}{\texttt{\pythia-410M} $0$-shot} \\
    \midrule
    Random & 25.6  & 40.2  & 39.7  & 24.7  & 29.4  & 67.1  & 50.6  & 64.1  & 42.7 \\
    DSIR  & 23.8  & 39.9  & 39.6  & 27.0    & 28.4  & 66.8  & 51.5  & 63.1  & 42.5 \\
    DsDm  & 24.7  & 41.7  & 40.3  & \textbf{27.5} & 29    & 68.1  & 50.1  & 65.4  & 43.4 \\
    QuRating & 25.4  & 42.0    & 40.7  & 25.3  & 30.2  & 67.5  & 52.1  & 64.8  & 43.5 \\
    MATES & 25.0    & 41.8  & 41.0    & 25.7  & 30.8  & \textbf{68.7} & 52.7  & 66.0    & 44.0 \\
    \method & \textbf{27.2} & \textbf{48.9} & \textbf{43.1} & 26.9  & \textbf{31.8} & 68.4  & \textbf{54.1} & \textbf{69.5} & \textbf{46.2} \\
    \midrule
          \multicolumn{10}{c}{\texttt{\pythia-410M} $2$-shot} \\ \midrule
    Random & 25.3  & 42.6  & 39.9  & 24.1  & 28.6  & 66.9  & 52.2  & 70.6  & 43.8 \\
    DSIR  & 23.6  & 42.0    & 39.8  & \textbf{26.1} & 28.6  & 66.1  & 51.6  & 71.4  & 43.7 \\
    DsDm  & 23.6  & 44.2  & 40.1  & 23.5  & 29.2  & 66.5  & 51.5  & 74    & 44.1 \\
    QuRating & 23.6  & 43.9  & 40.4  & \textbf{26.1} & 30.2  & 67.4  & 51.4  & 74.1  & 44.6 \\
    MATES & 25.3  & 43.8  & 40.6  & 24.9  & 30.6  & 67.1  & 53.4  & 74.1  & 45.0 \\
    \method & \textbf{27.0} & \textbf{52.7} & \textbf{42.6} & 23.7  & \textbf{32.8} & \textbf{68.2} & \textbf{53.9} & \textbf{78.9} & \textbf{47.5} \\
    \midrule
          \multicolumn{10}{c}{\texttt{\pythia-1B} $0$-shot} \\
    \midrule
    Random & 25.6  & 43.7  & 43.8  & 27.5  & 31.8  & 68.9  & 50.7  & 65.8  & 44.7 \\
    MATES & 25.9  & 44.9  & 45.3  & \textbf{28.7} & \textbf{32.2} & 69.5  & 52.4  & 67.3  & 45.8 \\
    \method & \textbf{26.2} & \textbf{49.1} & \textbf{46.6} & 24.8  & \textbf{32.2} & \textbf{70.3} & \textbf{54.2} & \textbf{70.9} & \textbf{46.8} \\
    \midrule
          \multicolumn{10}{c}{\texttt{\pythia-1B} $2$-shot} \\
    \midrule
    Random & 25.5  & 45.1  & 42.9  & 24.6  & 30.0    & 68.3  & 52.1  & 74.6  & 45.4 \\
    MATES & 26.8  & 46.1  & 44.8  & 25.2  & 30.6  & 68.7  & 51.6  & 75.7  & 46.2 \\
    \method & \textbf{27.3} & \textbf{54.5} & \textbf{46.2} & \textbf{26.6} & \textbf{32.2} & \textbf{69.0} & \textbf{53.9} & \textbf{77.4} & \textbf{48.4} \\
    \bottomrule
    \end{tabular}%
    }
\label{tab:exp-1-full-results-mates}%
\end{table}%

%% file: tables/app-exp2-modelsizes-full.tex
\begin{table}[htbp]
  \centering
  \caption{Full evaluation results of \tlmxs trained on different \method model curated data.}
  \resizebox{.95\textwidth}{!}{
    \begin{tabular}{c|ccccccccccc}
    \toprule
    \begin{tabular}[c]{@{}c@{}}\textbf{Train}\\ \textbf{Steps}\end{tabular} & \textbf{ARC-C} & \textbf{ARC-E} & \textbf{CSQA} & \textbf{HellaSwag} & \textbf{MMLU} & \textbf{OBQA} & \textbf{PiQA} & \textbf{SIQA} & \textbf{WinoG} & \textbf{SciQ} & \textbf{AVG} \\
    \midrule
    \multicolumn{12}{c}{\tlmxs trained on Raw data} \\
    \midrule
    2500  & 22.5  & 38.5  & 27.0  & 29.1  & 25.8  & 25.0  & 60.2  & 38.8  & 50.4  & 58.6  & 37.6 \\
    5000  & 23.6  & 39.2  & 28.7  & 33.1  & 26.1  & 26.6  & 62.2  & 39.5  & 49.9  & 66.2  & 39.5 \\
    7500  & 23.8  & 42.7  & 28.0  & 33.4  & 26.0  & 26.2  & 64.0  & 39.3  & 51.5  & 67.0  & 40.2 \\
    10000 & 23.8  & 41.2  & 27.8  & 35.0  & 26.6  & 28.0  & 65.3  & 40.9  & 50.1  & 65.9  & 40.5 \\
    12500 & 22.6  & 41.9  & 29.7  & 32.8  & 26.2  & 26.4  & 62.2  & 39.3  & 51.3  & 63.3  & 39.6 \\
    \midrule
    \multicolumn{12}{c}{\tlmxs trained on \method-xs data} \\
    \midrule
    2500  & 24.8  & 43.5  & 26.5  & 30.3  & 26.8  & 26.6  & 59.3  & 38.6  & 50.8  & 60.7  & 38.8 \\
    5000  & 23.7  & 44.3  & 28.1  & 33.8  & 27.3  & 28.8  & 61.3  & 38.9  & 50.9  & 70.2  & 40.7 \\
    7500  & 24.1  & 46.0  & 29.2  & 35.0  & 27.7  & 30.6  & 63.4  & 38.7  & 52.0  & 70.4  & 41.7 \\
    10000 & 25.3  & 46.1  & 28.3  & 35.7  & 28.1  & 29.2  & 64.4  & 38.5  & 51.2  & 70.6  & 41.7 \\
    12500 & 25.9  & 47.5  & 29.2  & 36.7  & 28.1  & 30.2  & 64.6  & 38.0  & 51.7  & 71.4  & 42.3 \\
    \midrule
    \multicolumn{12}{c}{\tlmxs trained on \method-s data} \\
    \midrule
    2500  & 23.5  & 41.9  & 24.9  & 30.4  & 26.6  & 27.6  & 62.0  & 37.8  & 49.3  & 61.4  & 38.5 \\
    5000  & 24.7  & 44.5  & 27.0  & 33.8  & 27.5  & 28.0  & 62.4  & 38.0  & 50.6  & 67.0  & 40.3 \\
    7500  & 25.3  & 45.3  & 27.3  & 34.0  & 27.9  & 29.2  & 63.4  & 37.7  & 52.9  & 68.7  & 41.2 \\
    10000 & 25.6  & 45.7  & 27.6  & 35.6  & 28.6  & 30.2  & 63.6  & 37.4  & 52.0  & 71.1  & 41.7 \\
    12500 & 26.4  & 46.7  & 27.5  & 37.2  & 28.1  & 29.8  & 62.8  & 37.8  & 52.2  & 70.1  & 41.9 \\
    \midrule
    \multicolumn{12}{c}{\tlmxs trained on \method-m curated data} \\
    \midrule
    2500  & 22.9  & 41.3  & 26.5  & 31.1  & 26.9  & 27.0  & 62.2  & 37.6  & 50.6  & 62.4  & 38.9 \\
    5000  & 25.8  & 44.0  & 27.3  & 34.0  & 27.1  & 29.6  & 63.1  & 38.5  & 51.8  & 64.9  & 40.6 \\
    7500  & 26.0  & 45.3  & 28.5  & 36.6  & 27.7  & 29.8  & 63.6  & 39.4  & 51.3  & 68.5  & 41.7 \\
    10000 & 26.0  & 46.6  & 28.8  & 37.3  & 27.6  & 30.6  & 63.3  & 38.7  & 51.6  & 70.3  & 42.1 \\
    12500 & 26.5  & 46.4  & 29.1  & 37.6  & 28.1  & 29.4  & 64.1  & 38.7  & 51.5  & 68.0  & 41.9 \\
    \bottomrule
    \end{tabular}%
    }
  \label{tab:exp-2-full-results-model-size-xs}%
\end{table}%

\begin{table}[htbp]
  \centering
  \caption{Full evaluation results of \tlms trained on different \method model curated data.}
  \resizebox{.95\textwidth}{!}{
    \begin{tabular}{c|ccccccccccc}
    \toprule
    \begin{tabular}[c]{@{}c@{}}\textbf{Train}\\ \textbf{Steps}\end{tabular} & \textbf{ARC-C} & \textbf{ARC-E} & \textbf{CSQA} & \textbf{HellaSwag} & \textbf{MMLU} & \textbf{OBQA} & \textbf{PiQA} & \textbf{SIQA} & \textbf{WinoG} & \textbf{SciQ} & \textbf{AVG} \\
    \midrule
    \multicolumn{12}{c}{\tlms trained on Raw data} \\
    \midrule
    2500  & 22.1  & 39.0  & 27.6  & 31.6  & 25.9  & 26.6  & 61.2  & 37.3  & 48.9  & 59.1  & 37.9 \\
    5000  & 24.4  & 41.2  & 28.8  & 34.8  & 26.7  & 27.0  & 64.9  & 39.3  & 50.4  & 61.9  & 39.9 \\
    7500  & 26.5  & 43.9  & 29.5  & 37.2  & 27.2  & 29.0  & 64.8  & 38.7  & 50.8  & 68.2  & 41.6 \\
    10000 & 25.8  & 43.5  & 29.1  & 38.8  & 27.4  & 29.8  & 66.9  & 39.0  & 51.2  & 66.2  & 41.8 \\
    12500 & 26.1  & 44.3  & 29.7  & 39.1  & 27.3  & 29.2  & 66.9  & 39.0  & 52.0  & 67.4  & 42.1 \\
    \midrule
    \multicolumn{12}{c}{\tlms trained on \method-xs curated data} \\
    \midrule
    2500  & 23.8  & 44.1  & 26.5  & 33.5  & 26.9  & 29.4  & 60.7  & 38.9  & 50.6  & 62.1  & 39.6 \\
    5000  & 26.8  & 48.1  & 28.4  & 36.7  & 28.0  & 30.6  & 64.0  & 38.6  & 50.3  & 65.6  & 41.7 \\
    7500  & 26.9  & 49.0  & 30.6  & 39.5  & 28.2  & 29.6  & 65.3  & 39.6  & 52.2  & 69.6  & 43.0 \\
    10000 & 26.7  & 51.3  & 29.4  & 40.1  & 28.3  & 31.8  & 64.1  & 39.3  & 51.4  & 69.9  & 43.2 \\
    12500 & 26.8  & 52.1  & 30.2  & 41.8  & 28.5  & 31.6  & 65.5  & 39.5  & 51.9  & 70.8  & 43.9 \\
    \midrule
    \multicolumn{12}{c}{\tlms trained on \method-s curated data} \\
    \midrule
    2500  & 24.9  & 43.4  & 27.3  & 32.1  & 26.9  & 28.2  & 60.9  & 38.8  & 51.2  & 60.8  & 39.5 \\
    5000  & 24.9  & 49.6  & 28.8  & 36.8  & 27.9  & 30.6  & 64.7  & 38.8  & 51.1  & 66.9  & 42.0 \\
    7500  & 25.5  & 51.2  & 30.8  & 38.8  & 28.4  & 31.2  & 67.3  & 40.2  & 50.3  & 71.7  & 43.5 \\
    10000 & 26.2  & 51.7  & 30.8  & 39.9  & 29.0  & 32.6  & 68.6  & 39.7  & 51.7  & 73.7  & 44.4 \\
    12500 & 26.4  & 51.9  & 30.9  & 42.4  & 29.4  & 31.6  & 67.9  & 40.0  & 52.2  & 73.5  & 44.6 \\
    \midrule
    \multicolumn{12}{c}{\tlms trained on \method-m curated data} \\
    \midrule
    2500  & 25.3  & 45.3  & 27.5  & 32.2  & 26.7  & 27.0  & 62.4  & 38.7  & 50.6  & 60.8  & 39.6 \\
    5000  & 26.1  & 45.4  & 28.6  & 37.2  & 27.4  & 27.8  & 65.7  & 38.9  & 50.9  & 65.6  & 41.4 \\
    7500  & 27.1  & 47.5  & 30.6  & 41.0  & 28.6  & 29.2  & 66.8  & 39.3  & 51.1  & 69.9  & 43.1 \\
    10000 & 26.7  & 50.5  & 30.7  & 41.5  & 28.4  & 30.2  & 67.0  & 40.1  & 49.9  & 70.9  & 43.6 \\
    12500 & 27.4  & 50.7  & 30.6  & 42.0  & 28.8  & 30.2  & 67.4  & 39.4  & 48.8  & 70.1  & 43.5 \\
    \bottomrule
    \end{tabular}%
    }
  \label{tab:exp-2-full-results-model-size-s}%
\end{table}%

\begin{table}[htbp]
  \centering
  \caption{Full evaluation results of \tlmm trained on different \method model curated data.}
  \resizebox{.95\textwidth}{!}{
    \begin{tabular}{c|ccccccccccc}
    \toprule
    \begin{tabular}[c]{@{}c@{}}\textbf{Train}\\ \textbf{Steps}\end{tabular} & \textbf{ARC-C} & \textbf{ARC-E} & \textbf{CSQA} & \textbf{HellaSwag} & \textbf{MMLU} & \textbf{OBQA} & \textbf{PiQA} & \textbf{SIQA} & \textbf{WinoG} & \textbf{SciQ} & \textbf{AVG} \\
    \midrule
    \multicolumn{12}{c}{\tlms trained on Raw data} \\
    \midrule
    2500  & 23.5  & 41.5  & 27.5  & 32.9  & 26.4  & 25.2  & 62.1  & 39.4  & 51.5  & 65.1  & 39.5 \\
    5000  & 24.0  & 42.1  & 29.6  & 37.6  & 27.6  & 27.2  & 65.0  & 39.7  & 53.2  & 68.5  & 41.4 \\
    7500  & 24.3  & 44.9  & 28.9  & 39.3  & 27.8  & 27.6  & 66.4  & 40.4  & 51.3  & 69.2  & 42.0 \\
    10000 & 24.8  & 46.1  & 29.6  & 41.4  & 27.9  & 28.4  & 67.5  & 39.8  & 51.9  & 70.9  & 42.8 \\
    12500 & 26.3  & 46.8  & 29.0  & 43.2  & 28.3  & 27.8  & 68.2  & 40.5  & 50.7  & 72.5  & 43.3 \\
    \midrule
    \multicolumn{12}{c}{\tlmm trained on \method-xs curated data} \\
    \midrule
    2500  & 24.9  & 49.6  & 26.5  & 34.0  & 27.3  & 30.4  & 61.8  & 37.9  & 51.3  & 65.1  & 40.9 \\
    5000  & 26.7  & 47.6  & 28.6  & 39.7  & 28.5  & 31.8  & 65.4  & 39.5  & 50.2  & 70.7  & 42.9 \\
    7500  & 27.5  & 52.1  & 30.4  & 41.8  & 29.6  & 31.8  & 67.6  & 39.6  & 51.7  & 75.2  & 44.7 \\
    10000 & 28.4  & 54.7  & 29.8  & 45.2  & 30.8  & 31.8  & 67.9  & 39.7  & 52.0  & 77.7  & 45.8 \\
    12500 & 28.8  & 54.2  & 29.7  & 46.5  & 30.9  & 31.8  & 68.2  & 39.9  & 51.3  & 78.3  & 46.0 \\
    \midrule
    \multicolumn{12}{c}{\tlmm trained on \method-s curated data} \\
    \midrule
    2500  & 25.3  & 45.7  & 27.8  & 34.2  & 27.8  & 29.0  & 64.4  & 37.5  & 49.3  & 66.3  & 40.7 \\
    5000  & 26.1  & 49.0  & 28.8  & 40.2  & 29.2  & 30.8  & 65.6  & 39.0  & 50.5  & 71.2  & 43.0 \\
    7500  & 27.7  & 53.6  & 31.1  & 44.1  & 29.6  & 34.8  & 67.6  & 39.4  & 52.5  & 72.2  & 45.3 \\
    10000 & 27.2  & 54.0  & 31.5  & 45.1  & 30.3  & 33.8  & 67.7  & 39.7  & 52.9  & 74.2  & 45.6 \\
    12500 & 28.6  & 56.1  & 31.8  & 45.5  & 30.5  & 34.4  & 68.5  & 39.4  & 51.3  & 76.1  & 46.2 \\
    \midrule
    \multicolumn{12}{c}{\tlmm trained on \method-m curated data} \\
    \midrule
    2500  & 24.7  & 44.1  & 25.9  & 34.8  & 27.4  & 27.8  & 62.9  & 38.9  & 49.2  & 67.0  & 40.3 \\
    5000  & 27.7  & 48.0  & 26.8  & 40.5  & 28.5  & 30.6  & 67.4  & 39.4  & 50.3  & 69.1  & 42.8 \\
    7500  & 26.7  & 51.9  & 26.7  & 42.9  & 29.3  & 31.4  & 69.1  & 40.3  & 50.4  & 73.3  & 44.2 \\
    10000 & 28.4  & 52.4  & 27.9  & 45.0  & 29.7  & 32.0  & 70.2  & 40.0  & 51.9  & 75.4  & 45.3 \\
    12500 & 28.3  & 53.7  & 28.4  & 45.9  & 30.1  & 33.8  & 70.6  & 41.1  & 52.3  & 72.5  & 45.7 \\
    \bottomrule
    \end{tabular}%
    }
  \label{tab:exp-2-full-results-model-size-m}%
\end{table}%

%% file: tables/app-exp2-pt-corpora-full.tex
\begin{table}[htbp]
  \centering
  \caption{Full evaluation results on scaling pre-training to about $50$B tokens on \redpj.}
  \resizebox{.95\textwidth}{!}{
    \begin{tabular}{cccccccccccc}
    \toprule
    \begin{tabular}[c]{@{}c@{}}\textbf{Train}\\ \textbf{Steps}\end{tabular} & \textbf{ARC-C} & \textbf{ARC-E} & \textbf{CSQA} & \textbf{HellaSwag} & \textbf{MMLU} & \textbf{OBQA} & \textbf{PiQA} & \textbf{SIQA} & \textbf{WinoG} & \textbf{SciQ} & \textbf{AVG} \\
    \midrule
    \multicolumn{12}{c}{\tlmm trained on \redpj raw data.} \\
    \midrule
    2500  & 24.0  & 42.9  & 26.6  & 33.7  & 25.9  & 26.0  & 62.4  & 39.4  & 52.3  & 64.0  & 39.7 \\
    5000  & 24.3  & 45.9  & 26.4  & 37.4  & 27.0  & 27.6  & 64.1  & 39.7  & 49.5  & 66.2  & 40.8 \\
    7500  & 25.1  & 45.3  & 28.8  & 40.3  & 27.1  & 29.2  & 66.3  & 39.1  & 51.7  & 66.9  & 42.0 \\
    10000 & 25.8  & 49.3  & 31.5  & 42.5  & 28.0  & 28.8  & 66.7  & 39.6  & 51.5  & 74.0  & 43.8 \\
    12500 & 25.3  & 50.1  & 30.2  & 43.0  & 28.2  & 30.0  & 66.6  & 39.2  & 51.1  & 74.2  & 43.8 \\
    15000 & 26.2  & 50.3  & 31.2  & 44.3  & 28.8  & 28.4  & 68.2  & 39.8  & 51.7  & 76.2  & 44.5 \\
    17500 & 25.8  & 51.1  & 30.8  & 44.7  & 29.0  & 29.6  & 67.7  & 39.2  & 52.6  & 75.2  & 44.6 \\
    20000 & 26.7  & 52.5  & 31.7  & 47.2  & 28.6  & 30.4  & 69.0  & 39.6  & 53.0  & 78.2  & 45.7 \\
    22500 & 27.4  & 51.7  & 32.1  & 47.2  & 29.3  & 30.4  & 69.5  & 39.5  & 51.9  & 78.5  & 45.7 \\
    25000 & 26.9  & 51.4  & 32.4  & 47.3  & 29.3  & 32.2  & 69.7  & 39.6  & 52.1  & 79.1  & 46.0 \\
    \midrule
    \multicolumn{12}{c}{\tlmm trained on \method refined \redpj data.} \\
    \midrule
    2500  & 24.8  & 46.8  & 27.2  & 33.8  & 27.3  & 28.2  & 61.3  & 38.6  & 50.3  & 65.1  & 40.3 \\
    5000  & 26.9  & 49.3  & 28.5  & 40.1  & 28.0  & 30.6  & 66.2  & 39.7  & 50.2  & 70.1  & 43.0 \\
    7500  & 28.5  & 53.1  & 29.2  & 41.7  & 29.4  & 33.2  & 66.9  & 39.3  & 53.0  & 73.0  & 44.7 \\
    10000 & 28.2  & 53.5  & 30.1  & 43.6  & 29.8  & 31.6  & 68.4  & 39.6  & 52.0  & 75.3  & 45.2 \\
    12500 & 29.5  & 55.3  & 30.2  & 46.4  & 30.5  & 32.2  & 68.6  & 40.2  & 52.6  & 76.9  & 46.2 \\
    15000 & 30.0  & 57.1  & 30.2  & 47.6  & 30.9  & 33.0  & 69.5  & 39.8  & 52.2  & 77.8  & 46.8 \\
    17500 & 31.5  & 59.6  & 29.4  & 49.5  & 31.6  & 33.6  & 69.4  & 39.8  & 53.0  & 78.9  & 47.6 \\
    20000 & 31.2  & 61.2  & 29.4  & 50.4  & 31.4  & 35.2  & 70.6  & 40.1  & 53.7  & 79.6  & 48.3 \\
    22500 & 32.0  & 61.7  & 30.2  & 51.4  & 31.4  & 34.0  & 70.0  & 39.9  & 53.2  & 79.5  & 48.3 \\
    25000 & 31.1  & 60.7  & 29.8  & 51.0  & 31.7  & 33.2  & 70.9  & 39.2  & 53.3  & 79.1  & 48.0 \\
    \bottomrule
    \end{tabular}%
    }
  \label{tab:exp-2-full-results-50B-redpj}%
\end{table}%

\begin{table}[htbp]
  \centering
  \caption{Full evaluation results on scaling pre-training to about $50$B tokens on C4.}
  \resizebox{.95\textwidth}{!}{
    \begin{tabular}{cccccccccccc}
    \toprule
    \begin{tabular}[c]{@{}c@{}}\textbf{Train}\\ \textbf{Steps}\end{tabular} & \textbf{ARC-C} & \textbf{ARC-E} & \textbf{CSQA} & \textbf{HellaSwag} & \textbf{MMLU} & \textbf{OBQA} & \textbf{PiQA} & \textbf{SIQA} & \textbf{WinoG} & \textbf{SciQ} & \textbf{AVG} \\
    \midrule
    \multicolumn{12}{c}{\tlmm trained on C4 raw data.} \\
    \midrule
    2500  & 22.4  & 39.7  & 26.8  & 36.5  & 26.5  & 27.6  & 64.8  & 40.2  & 50.1  & 60.0  & 39.5 \\
    5000  & 23.9  & 42.9  & 27.5  & 42.3  & 27.1  & 29.6  & 68.2  & 39.6  & 50.3  & 66.6  & 41.8 \\
    7500  & 25.1  & 44.8  & 28.2  & 45.4  & 27.1  & 29.2  & 70.7  & 40.7  & 51.6  & 66.3  & 42.9 \\
    10000 & 25.5  & 46.0  & 32.3  & 48.2  & 27.9  & 31.6  & 71.1  & 39.7  & 52.3  & 67.6  & 44.2 \\
    12500 & 25.8  & 48.8  & 30.3  & 49.7  & 27.9  & 31.6  & 71.2  & 40.9  & 52.0  & 69.4  & 44.8 \\
    15000 & 26.9  & 48.0  & 28.2  & 50.5  & 28.5  & 31.4  & 71.9  & 41.1  & 51.4  & 69.7  & 44.8 \\
    17500 & 26.6  & 48.8  & 30.3  & 52.1  & 28.6  & 31.2  & 73.2  & 41.6  & 52.0  & 70.0  & 45.4 \\
    20000 & 26.3  & 50.1  & 29.7  & 52.5  & 28.5  & 32.6  & 72.3  & 41.7  & 52.3  & 71.0  & 45.7 \\
    22500 & 25.8  & 50.7  & 31.0  & 52.9  & 28.8  & 33.8  & 73.0  & 41.6  & 53.0  & 71.5  & 46.2 \\
    25000 & 25.3  & 48.8  & 30.1  & 52.4  & 28.8  & 32.2  & 72.0  & 40.6  & 53.6  & 71.7  & 45.5 \\
    \midrule
    \multicolumn{12}{c}{\tlmm trained on \method refined C4 data.} \\
    \midrule
    2500  & 24.1  & 45.9  & 26.0  & 37.3  & 27.2  & 29.0  & 66.3  & 39.8  & 50.8  & 65.9  & 41.2 \\
    5000  & 27.3  & 50.0  & 26.6  & 42.4  & 28.6  & 33.8  & 68.1  & 40.5  & 53.0  & 71.9  & 44.2 \\
    7500  & 28.3  & 53.7  & 27.7  & 47.7  & 29.3  & 35.4  & 71.1  & 39.3  & 54.0  & 73.1  & 46.0 \\
    10000 & 30.0  & 54.3  & 28.1  & 50.9  & 30.0  & 33.6  & 71.2  & 40.6  & 52.0  & 74.2  & 46.5 \\
    12500 & 29.3  & 56.7  & 27.5  & 52.3  & 30.9  & 33.8  & 72.8  & 39.9  & 52.5  & 77.5  & 47.3 \\
    15000 & 29.6  & 55.9  & 28.3  & 53.9  & 30.6  & 35.0  & 72.9  & 41.0  & 53.8  & 75.8  & 47.7 \\
    17500 & 30.6  & 55.5  & 28.7  & 53.3  & 31.2  & 34.2  & 73.6  & 40.4  & 53.4  & 76.7  & 47.8 \\
    20000 & 30.0  & 57.6  & 28.3  & 54.9  & 31.1  & 37.2  & 74.6  & 40.7  & 53.6  & 79.4  & 48.7 \\
    22500 & 30.1  & 56.7  & 28.6  & 55.2  & 31.4  & 37.2  & 73.8  & 41.6  & 53.3  & 77.7  & 48.6 \\
    25000 & 31.1  & 56.0  & 28.4  & 55.2  & 31.1  & 36.2  & 74.0  & 41.0  & 54.1  & 76.8  & 48.4 \\
    \bottomrule
    \end{tabular}%
    }
  \label{tab:exp-2-full-results-50B-c4}%
\end{table}%

\begin{table}[htbp]
  \centering
  \caption{Full evaluation results on scaling pre-training to about $50$B tokens on \fineweb.}
  \resizebox{.95\textwidth}{!}{
    \begin{tabular}{cccccccccccc}
    \toprule
    \begin{tabular}[c]{@{}c@{}}\textbf{Train}\\ \textbf{Steps}\end{tabular} & \textbf{ARC-C} & \textbf{ARC-E} & \textbf{CSQA} & \textbf{HellaSwag} & \textbf{MMLU} & \textbf{OBQA} & \textbf{PiQA} & \textbf{SIQA} & \textbf{WinoG} & \textbf{SciQ} & \textbf{AVG} \\
    \midrule
    \multicolumn{12}{c}{\tlmm trained on \fineweb raw data.} \\
    \midrule
    2500  & 22.9  & 41.2  & 28.9  & 34.3  & 26.1  & 27.6  & 64.8  & 39.3  & 52.1  & 62.8  & 40.0 \\
    5000  & 25.5  & 44.5  & 30.4  & 39.8  & 26.9  & 32.0  & 68.4  & 39.2  & 52.1  & 67.2  & 42.6 \\
    7500  & 26.8  & 45.6  & 31.4  & 44.1  & 27.6  & 30.2  & 70.9  & 38.8  & 52.2  & 70.3  & 43.8 \\
    10000 & 27.2  & 46.2  & 31.3  & 47.2  & 28.3  & 31.6  & 72.1  & 38.8  & 53.4  & 69.0  & 44.5 \\
    12500 & 26.4  & 49.2  & 32.1  & 48.7  & 28.7  & 31.6  & 71.5  & 40.1  & 52.6  & 74.7  & 45.6 \\
    15000 & 27.1  & 49.6  & 32.8  & 49.5  & 28.9  & 31.0  & 72.7  & 39.0  & 52.3  & 77.1  & 46.0 \\
    17500 & 26.4  & 50.9  & 33.8  & 51.3  & 29.3  & 31.0  & 71.9  & 39.3  & 53.0  & 78.0  & 46.5 \\
    20000 & 27.1  & 53.1  & 33.2  & 51.2  & 29.6  & 32.2  & 73.4  & 39.7  & 52.3  & 76.3  & 46.8 \\
    22500 & 27.1  & 51.2  & 34.9  & 51.7  & 29.5  & 33.4  & 73.7  & 40.1  & 52.4  & 78.0  & 47.2 \\
    25000 & 28.5  & 52.6  & 33.9  & 53.2  & 29.8  & 32.6  & 72.9  & 40.2  & 53.0  & 77.1  & 47.4 \\
    \midrule
    \multicolumn{12}{c}{\tlmm trained on \method refined \fineweb data.} \\
    \midrule
    2500  & 25.8  & 46.8  & 27.4  & 36.1  & 27.7  & 28.8  & 63.9  & 39.3  & 51.9  & 69.1  & 41.7 \\
    5000  & 28.5  & 52.1  & 28.8  & 43.5  & 29.3  & 32.6  & 66.4  & 38.7  & 51.2  & 71.3  & 44.2 \\
    7500  & 28.2  & 52.0  & 30.6  & 45.9  & 29.9  & 33.0  & 69.3  & 39.5  & 51.7  & 71.8  & 45.2 \\
    10000 & 29.3  & 54.3  & 30.6  & 48.5  & 30.8  & 33.2  & 69.7  & 40.7  & 50.6  & 74.4  & 46.2 \\
    12500 & 28.7  & 57.8  & 30.7  & 48.1  & 31.1  & 32.6  & 72.0  & 40.4  & 52.7  & 77.4  & 47.2 \\
    15000 & 31.1  & 59.6  & 31.9  & 50.4  & 31.8  & 34.4  & 71.9  & 40.5  & 50.8  & 78.0  & 48.0 \\
    17500 & 32.6  & 60.9  & 31.9  & 51.5  & 32.2  & 33.8  & 72.3  & 39.7  & 52.5  & 78.9  & 48.6 \\
    20000 & 33.2  & 62.5  & 32.5  & 51.6  & 32.4  & 34.6  & 72.4  & 39.7  & 51.7  & 80.7  & 49.1 \\
    22500 & 34.7  & 63.6  & 32.9  & 53.3  & 32.9  & 34.8  & 73.1  & 40.3  & 54.2  & 80.5  & 50.0 \\
    25000 & 34.4  & 63.9  & 32.6  & 53.0  & 33.1  & 34.4  & 73.1  & 39.3  & 52.7  & 81.5  & 49.8 \\
    \bottomrule
    \end{tabular}%
    }
  \label{tab:exp-2-full-results-50B-fw}%
\end{table}%

\begin{table}[htbp]
  \centering
  \caption{Detailed evaluation results of existing base models trained on different corpora and trained using different techniques.}
  \resizebox{.95\textwidth}{!}{
    \begin{tabular}{ccccccccccc}
    \toprule
    \textbf{ARC-C} & \textbf{ARC-E} & \textbf{CSQA} & \textbf{HellaSwag} & \textbf{MMLU} & \textbf{OBQA} & \textbf{PiQA} & \textbf{SIQA} & \textbf{WinoG} & \textbf{SciQ} & \textbf{AVG} \\
    \midrule
    \multicolumn{11}{c}{\tinylm{} (trained on 3T tokens)} \\
    \midrule
    31.5  & 59.0  & 35.5  & 57.8  & 32.8  & 33.4  & 72.8  & 40.0  & 56.0  & 82.4  & 50.1 \\
    \midrule
    \multicolumn{11}{c}{\textsc{OLMo}-1B (trained on 2T tokens)}\\
    \midrule
    31.4  & 59.7  & 38.9  & 61.9  & 32.2  & 38.4  & 76.1  & 41.5  & 53.9  & 78.8  & 51.3 \\
    \midrule
    \multicolumn{11}{c}{\textsc{Pythia}-1.4B} \\
    \midrule
    28.7  & 56.9  & 34.7  & 51.7  & 31.5  & 36.0  & 71.8  & 40.8  & 55.1  & 79.3  & 48.7 \\
    \midrule
    \multicolumn{11}{c}{\textsc{Pythia}-2.8B} \\
    \midrule
    32.9  & 61.0  & 36.5  & 60.4  & 33.3  & 35.0  & 73.5  & 41.1  & 57.0  & 83.1  & 51.4 \\
    \midrule
    \multicolumn{11}{c}{\textsc{ShearedLlama}-1.3B~(pruned from \llamaii{}-7B)} \\
    \midrule
    22.4  & 39.7  & 29.3  & 36.0  & 26.4  & 28.4  & 62.6  & 39.9  & 52.0  & 71.4  & 40.8 \\
    \midrule
    \multicolumn{11}{c}{\textsc{ShearedLLama-1.3B}~(pruned from \llamaii{}-7B, and further trained on $50$B tokens)} \\
    \midrule
    29.0  & 58.3  & 34.8  & 59.6  & 32.0  & 35.0  & 74.6  & 41.0  & 56.3  & 82.3  & 50.3 \\
    \midrule
    \multicolumn{11}{c}{\textsc{InstructLM}-1.3B~(LLM data synthesis)} \\
    \midrule
    28.1  & 57.9  & 32.5  & 52.3  & 30.0  & 34.0  & 74.5  & 39.9  & 56.1  & 86.9  & 49.2 \\
    \midrule
    \multicolumn{11}{c}{\textsc{Cosmo}-1.8B~(LLM data synthesis)} \\
    \midrule
    33.4  & 57.0  & 31.2  & 55.1  & 32.4  & 35.2  & 71.4  & 42.0  & 54.7  & 84.4  & 49.7 \\
    \bottomrule
    \end{tabular}%
    }
  \label{tab:exp-2-full-results-existing-models}%
\end{table}%

%% file: tables/exp-3-full-ablation-results.tex
\begin{table}[h]
  \small
  \centering
  \setlength{\tabcolsep}{2.0pt}
  \caption{Full ablation results on OpenWebMath Continual Pre-training~(CPT). All models are tested using few-shot CoT prompts. \llemma{} and \internmath{} are continual pre-trained models from \codellama{}~\citep{DBLP:journals/corr/abs-2308-12950-codellama} and \textsc{InternLM2}~\citep{team2023internlm} with public available data, respectively.
  \textsc{DeepSeek-LLM} denotes an internal DeepSeek model, and the model trained on \owm{} introduced by~\citet{shao2024deepseekmath}. Note that the unique tokens and training tokens in the column refer exclusively to the token numbers from math-specific corpora  (calculated by corresponding tokenizers). $^\dag$: MQA evaluation of \intern{} is based on an alternative prompt due to non-prediction issues with the original prompt. The \textbf{bolded} entries represent the best results within the same base model and CPT experiments.
  }
  \newcommand{\U}[1]{{#1}}
  \resizebox{\linewidth}{!}{%
    \begin{tabular}{l|ll|cc|ccccccccc|c}
    \toprule
    \textbf{Model} & \multicolumn{1}{c}{\textbf{Size}} & \multicolumn{1}{l}{\textbf{Method}} & \begin{tabular}[c]{@{}c@{}}\textbf{Uniq}\\ \textbf{Toks}\end{tabular} & \multicolumn{1}{c|}{\textbf{\begin{tabular}[c]{@{}c@{}}Train\\ Toks\end{tabular}}} & \textbf{GSM8K} & \textbf{MATH} & \textbf{SVAMP} & \textbf{ASDiv} & \textbf{MAWPS} & \textbf{TAB} & \textbf{MQA} & \begin{tabular}[c]{@{}c@{}}\textbf{MMLU}\\ \textbf{STEM}\end{tabular} & \textbf{\begin{tabular}[c]{@{}c@{}}{SAT}\\ {MATH}\end{tabular}} & \textbf{AVG} \\
    \midrule
    \multicolumn{15}{c}{\cellcolor[HTML]{F2F2F2}Existing Continual Pre-training for Reference} \\
    \midrule
    \multirow{2}{*}{\centering\textsc{DeepSeek-LLM}}& 1.3B 
        & - & - & - & 2.9  &  3.0 &  - & -  & - & - & - &  19.5 & 15.6 & - \\
        & 1.3B & - & 14B & 150B & 11.5  &  8.9 &  - & -  & - & - & - &  29.6 & 31.3 & - \\
    \midrule
    \multirow{2}[2]{*}{\centering\codellama{}~(Base)} & 7B    &  -     &  -  & - & 11.8 & 5.0 & 44.2 & 50.7 & 62.6 & 30.6 & 14.3 & 20.4 & 21.9 & 29.1  \\
    &  34B  & -     & -   &  -  & 31.8 & 10.8 & 61.9 & 66.0 & 83.4 & 51.6 & 23.7 & 43.0 & 53.1 & 47.3  \\ \midrule
    \multirow{2}{*}{\centering\llemma{}} & 7B    & -     & 55B   & 200B  & 38.8  & 17.2  & 56.1  & 69.1  & 82.4  & 48.7  & 41.0  & 45.4  & 59.4  & 50.9~(+21.8) \\
         & 34B   & -     & 55B   & 50B   & 54.2  & 23.0  & 67.9  & 75.7  & 90.1  & 57.9  & 49.8  & 54.7  & 68.8  & 60.1~(+12.8) \\
    \midrule
    \multirow{2}{*}{\centering\intern{}} & 7B    &  -     &  -  & - & 27.0 & 6.6 & 49.0 & 59.3 & 74.8 & 40.1 & 20.9$^\dag$ & 19.0 & 28.1 & 36.1  \\
    &  20B  & -     &  -  & -   & 50.6 & 18.8 & 72.5 & 75.9 & 93.9 & 45.4 & 33.1 & 53.7 & 59.4 & 55.9  \\
    \midrule
    \multirow{2}{*}{\centering\internmath{}} & 7B    & -     & 31B   & 125B  & 41.8  & 14.4  & 61.6  & 66.8  & 83.7  & 50.0  & 57.3  & 24.8  & 37.5  & 48.7~(+12.6) \\
          & 20B   & -     & 120B  & 500B  & 65.4  & 30.0  & 75.7  & 79.3  & 94.0  & 50.9  & 38.5  & 53.1  & 71.9  & 62.1~(+6.2) \\
    \midrule
    \multicolumn{15}{c}{\cellcolor[HTML]{F2F2F2}Applying Data Refinement Approaches} \\ \midrule
    \tinyllama (Base) & 1.1B  & -     & -     & -     & 2.8   & 3.2   & 10.9  & 18.0  & 20.2  & 12.5  & 14.6  & 16.4  & 21.9  & 14.7 \\ \midrule
    \multirow{4}[1]{*}{\centering\tinyllama (CPT)} & 1.1B  & -     & 15B   & 15B   & 6.2   & 4.8   & 22.3  & 36.2  & 47.6  & 19.3  & 11.6  & 20.7  & \U{25.0}  & 21.5 (+8.1) \\
          
          & 1.1B  & \textsc{Rho}   & 15B   & 9B$^*$\tablefootnote{\textsc{Rho}-1 only counts the selected tokens that are used for training~(loss calculation).}   & 7.1   & 5.0   & \U{23.5}  & \U{41.2}  & 53.8  & -     & \textbf{18.0} & -     & -     & - \\
          & 1.1B  & Rule  & 6.5B  & 15B   & 4.5   & 2.8   & 17.5  & 29.4  & 39.3  & 15.1  & 12.4  & 19.4  & \U{25.0}  & 18.4 (+3.7) \\
          
          & \HL{1.1B}  & \HL{\method-D}    & \HL{5.4B}  & \HL{15B}   & \HL{\textbf{9.3}}  & \HL{\textbf{7.4}} & \HL{23.4} & \HL{\textbf{41.9}} & \HL{\U{55.6}} & \HL{\U{22.1}}  & \HL{14.6}  & \HL{\U{24.1}} & \HL{\U{25.0}}  & \HL{\U{24.8 (+10.1)}} \\
          
          & \HL{1.1B}  & \HL{\method-D+C} & \HL{5B}  & \HL{15B}   & \HL{9.0} & \HL{5.6} & \HL\textbf{23.8} & \HL\textbf{41.9} & \HL\textbf{56.9} & \HL\textbf{22.2} & \HL\U{15.6} & \HL\textbf{26.8} & \HL\textbf{31.2} & \HL\textbf{25.7~(+11.0)} \\\midrule
    \llamaii (Base) & 7B    & -     & -     & -     & 14.1 & 3.8          & 39.5 & 51.6 & 63.6 & 30.9  & 12.5  & 32.9      & 34.4     & 31.5 
    \\ \midrule
    \multirow{2}[1]{*}{\llamaii (CPT)} & 7B    & -     & 15B     & 10B   & 29.6 & 13.6 & 49.2 & 61.9 & 78.4 & \U{36.3} & 31.9 & 40.5 & 43.8 & 42.8 (+11.3)  \\ 
     & \HL{7B}    & \HL{\method-D}     & \HL{5.4B}     & \HL{10B}     & \HL{\U{30.3}} & \HL{\U{16.0}} & \HL{\textbf{54.2}} & \HL{\textbf{63.8}} & \HL{\textbf{79.5}} & \HL{\textbf{37.3}} & \HL{\U{37.2}} & \HL{\textbf{44.2}} & \HL{\U{46.9}} & \HL{\U{45.5 (+14.0)}} \\ 
    & \HL{7B}    & \HL{\method-D+C}     & \HL{5B}     & \HL{10B}     &  \HL{\textbf{30.6}} & \HL{\textbf{16.8}} & \HL{50.2} & \HL{63.7} & \HL{79.3} & \HL{\textbf{37.3}} & \HL{\textbf{40.1}} & \HL{43.8} & \HL{\textbf{53.1}} & \HL{\textbf{46.1 (+14.6)}} \\ \midrule
    \codellama (Base) & 7B    & -     & -     & -    &  11.8 & 5.0         & 44.2 & 50.7 & 62.6 & 30.6  & 14.3  & 20.4      & 21.9     & 29.1  \\ 
    \midrule
    \multirow{2}[1]{*}{\codellama (CPT)} & 7B    & -     & 15B   & 10B  &  31.1  & 14.8          & 51.4  & 62.1  & 81.2  & 33.6   & 30.4   & 40.5       & 43.8      & 43.2 (+14.1) \\ 
    & \HL{7B}    & \HL{\method-D}   & \HL{5.4B}   & \HL{10B}  & \HL{\textbf{38.1}} & \HL{\U{17.0}}          & \HL{\U{54.2}}          & \HL{\U{67.0}}          & \HL{\textbf{83.1}} & \HL{\U{40.9}}          & \HL{\textbf{39.8}} & \HL{\textbf{43.7}} & \HL{\U{50.0}}          & \HL{\U{48.2 (+19.1)}} \\ 
    & \HL{7B}    & \HL{\method-D+C}  & \HL{5B}   & \HL{10B}  & \HL{35.6}          & \HL{\textbf{17.6}} & \HL{\textbf{55.8}} & \HL{\textbf{67.9}} & \HL{82.7}        & \HL{\textbf{41.3}} & \HL{38.9}         & \HL{42.6}          & \HL{\textbf{62.5}} & \HL{\textbf{49.4} \textbf{(+20.3)}} \\ \midrule
    \mistral (Base) & 7B    & -     & -     & -     & 40.6  & 11.4  & \textbf{65.4}  & 68.5  & 87.0  & \textbf{52.9}  & 32.3  & 50.0  & 56.2  & 51.6 \\ \midrule
    \multirow{2}[1]{*}{\mistral (CPT)} & 7B    & -     & 15B   & 10B   & 44.4  & 19.2  & \U{65.2}  & 69.6  & 88.4  & 46.6  & 43.1  & 50.8  & \U{65.6}  & 54.8 (+3.2) \\
          & \HL{7B}    & \HL{\method-D}    & \HL{5.5B}  & \HL{10B}   & \HL{\U{47.8}} & \HL\textbf{24.8} & \HL63.5 & \HL\U{72.4} & \HL\U{88.9} & \HL48.3 & \HL\U{48.2} & \HL\U{54.1} & \HL62.5 & \HL\U{56.4 (+4.8)} \\
          & \HL{7B}    & \HL{\method-D+C} & \HL{4.7B}  & \HL{10B}   & \HL\textbf{51.0} & \HL{22.4} & \HL{64.9} & \HL\textbf{72.9} & \HL\textbf{89.2} & \HL\U{49.8} & \HL\textbf{53.0} & \HL\textbf{54.2} & \HL\textbf{75.0} & \HL\textbf{59.2 (+7.6)} \\ 
    \bottomrule
    \end{tabular}%
    }
  \label{tab:exp3-owm-cpt-full-ablation-results}%
\end{table}%

%% file: tables/exp-3-tlm-full-results.tex
\begin{table}[htbp]
  \centering
  \caption{Full evaluation results of \tinylm continual pre-training on \owm with raw data. Note that about 1B tokens are trained per 500 steps.}
  \resizebox{0.95\textwidth}{!}{
    \begin{tabular}{c|ccccccccc|c}
    \toprule
    \begin{tabular}[c]{@{}l@{}}\textbf{Train}\\ \textbf{Steps}\end{tabular} &
    \textbf{GSM8K} &
    \textbf{MATH} &
    \textbf{SVAMP} &
    \textbf{ASDiv} &
    \textbf{MAWPS} &
    \textbf{TAB} &
    \textbf{MQA} &
    \begin{tabular}[c]{@{}c@{}}\textbf{MMLU}\\ \textbf{STEM}\end{tabular} &
    \begin{tabular}[c]{@{}c@{}}\textbf{SAT}\\ \textbf{MATH}\end{tabular} &
    \textbf{AVG} \\
    \midrule
    0     & 2.8   & 3.2   & 10.9  & 18    & 20.2  & 12.5  & 14.6  & 16.4  & 21.9  & 14.7 \\
    \midrule
    500    & 1.9   & 3.4   & 16.3  & 23.9  & 30.3  & 13.9  & 10.3  & 14.8  & 18.8  & 14.8 \\
    1000    & 3.1   & 2.2   & 16.6  & 25.6  & 32.4  & 12.5  & 12.0  & 16.6  & 25.0  & 16.2 \\
    1500    & 2.7   & 3.0   & 17.6  & 28.5  & 34.5  & 13.9  & 8.7   & 14.1  & 15.6  & 15.4 \\
    2000    & 4.5   & 3.2   & 16.4  & 28.5  & 39.0  & 15.1  & 10.2  & 16.6  & 34.4  & 18.7 \\
    2500    & 4.9   & 3.4   & 19.3  & 31.0  & 39.2  & 16.0  & 12.1  & 18.6  & 9.4   & 17.1 \\
    3000    & 4.1   & 5.2   & 19.1  & 32.0  & 43.0  & 15.3  & 9.6   & 16.1  & 18.8  & 18.1 \\
    3500    & 4.9   & 3.6   & 19.7  & 31.4  & 40.4  & 18.1  & 11.3  & 19.6  & 15.6  & 18.3 \\
    4000    & 4.8   & 4.8   & 19.5  & 33.8  & 44.5  & 16.4  & 10.7  & 19.9  & 12.5  & 18.5 \\
    4500    & 5.4   & 4.8   & 20.2  & 35.0  & 45.2  & 17.9  & 12.7  & 21.0  & 18.8  & 20.1 \\
    5000    & 5.5   & 4.6   & 22.3  & 34.6  & 42.9  & 16.0  & 10.6  & 21.7  & 28.1  & 20.7 \\
    5500    & 4.9   & 5.8   & 23.6  & 35.2  & 44.0  & 20.4  & 11.0  & 21.1  & 21.9  & 20.9 \\
    6000    & 6.1   & 4.4   & 22.8  & 36.2  & 45.4  & 17.8  & 12.7  & 21.4  & 15.6  & 20.3 \\
    6500    & 6.3   & 3.6   & 23.2  & 37.3  & 48.0  & 19.7  & 10.3  & 21.0  & 18.8  & 20.9 \\
    7000    & 6.1   & 4.6   & 22.2  & 36.6  & 46.9  & 19.4  & 12.0  & 21.5  & 21.9  & 21.2 \\
    7500    & 6.2   & 4.8   & 22.3  & 36.2  & 47.6  & 19.3  & 11.6  & 20.7  & 25.0  & 21.5 \\
    \bottomrule
    \end{tabular}%
    }
  \label{tab:exp3-tlm-1.1b-owm-cpt-raw}%
\end{table}

\begin{table}[htbp]
  \centering
  \caption{Full evaluation results of \tinylm continual pre-training on \owm with data after rule-based filtering. Note that about 1B tokens are trained per 500 steps.}
  \resizebox{0.95\textwidth}{!}{
    \begin{tabular}{c|ccccccccc|c}
    \toprule
    \begin{tabular}[c]{@{}l@{}}\textbf{Train}\\ \textbf{Steps}\end{tabular} &
    \textbf{GSM8K} &
    \textbf{MATH} &
    \textbf{SVAMP} &
    \textbf{ASDiv} &
    \textbf{MAWPS} &
    \textbf{TAB} &
    \textbf{MQA} &
    \begin{tabular}[c]{@{}c@{}}\textbf{MMLU}\\ \textbf{STEM}\end{tabular} &
    \begin{tabular}[c]{@{}c@{}}\textbf{SAT}\\ \textbf{MATH}\end{tabular} &
    \textbf{AVG} \\
    \midrule
    0     & 2.8   & 3.2   & 10.9  & 18    & 20.2  & 12.5  & 14.6  & 16.4  & 21.9  & 14.7 \\
    \midrule
    500    & 3.4   & 3.6   & 13.6  & 22.5  & 25.9  & 13.1  & 14.2  & 13.5  & 28.1  & 15.3 \\
    1000    & 3.0   & 2.8   & 14.1  & 22.5  & 27.8  & 11.4  & 11.0  & 16.4  & 12.5  & 13.5 \\
    1500    & 3.6   & 3.2   & 13.6  & 24.0  & 31.2  & 13.9  & 9.2   & 18.0  & 18.8  & 15.1 \\
    2000    & 3.5   & 2.4   & 15.0  & 25.1  & 33.0  & 12.5  & 10.6  & 13.9  & 15.6  & 14.6 \\
    2500    & 3.3   & 1.6   & 15.0  & 25.3  & 33.5  & 13.7  & 11.1  & 18.1  & 25.0  & 16.3 \\
    3000    & 3.5   & 3.0   & 16.4  & 25.5  & 33.4  & 14.1  & 10.2  & 18.4  & 18.8  & 15.9 \\
    3500    & 3.2   & 3.4   & 17.2  & 27.0  & 37.7  & 14.6  & 11.2  & 13.3  & 25.0  & 17.0 \\
    4000    & 3.5   & 3.6   & 15.6  & 26.2  & 36.5  & 13.4  & 12.1  & 15.9  & 18.8  & 16.2 \\
    4500    & 4.1   & 3.8   & 15.6  & 27.9  & 38.2  & 14.9  & 11.6  & 17.1  & 18.8  & 16.9 \\
    5000    & 4.2   & 3.6   & 18.6  & 28.7  & 37.7  & 14.3  & 12.7  & 17.5  & 21.9  & 17.7 \\
    5500    & 4.1   & 3.8   & 16.3  & 29.3  & 38.4  & 14.7  & 10.8  & 17.5  & 18.8  & 17.1 \\
    6000    & 4.3   & 3.6   & 16.0  & 28.7  & 39.1  & 13.5  & 12.8  & 19.5  & 21.9  & 17.7 \\
    6500    & 4.2   & 3.2   & 16.4  & 29.5  & 39.0  & 15.1  & 11.7  & 17.9  & 21.9  & 17.7 \\
    7000    & 4.0   & 4.0   & 16.2  & 29.6  & 37.9  & 16.0  & 13.8  & 17.8  & 21.9  & 17.9 \\
    7500    & 4.5   & 2.8   & 17.5  & 29.4  & 39.3  & 15.1  & 12.4  & 19.4  & 25.0  & 18.4 \\
    \bottomrule
    \end{tabular}%
    }
  \label{tab:exp3-tlm-1.1b-owm-cpt-rule}%
\end{table}

\begin{table}[htbp]
  \centering
  \caption{Full evaluation results of \tinylm continual pre-training on \owm with data after \method-D. Note that about 1B tokens are trained per 500 steps.}
  \resizebox{0.95\textwidth}{!}{
    \begin{tabular}{c|ccccccccc|c}
    \toprule
    \begin{tabular}[c]{@{}l@{}}\textbf{Train}\\ \textbf{Steps}\end{tabular} &
    \textbf{GSM8K} &
    \textbf{MATH} &
    \textbf{SVAMP} &
    \textbf{ASDiv} &
    \textbf{MAWPS} &
    \textbf{TAB} &
    \textbf{MQA} &
    \begin{tabular}[c]{@{}c@{}}\textbf{MMLU}\\ \textbf{STEM}\end{tabular} &
    \begin{tabular}[c]{@{}c@{}}\textbf{SAT}\\ \textbf{MATH}\end{tabular} &
    \textbf{AVG} \\
    \midrule
    0     & 2.8   & 3.2   & 10.9  & 18    & 20.2  & 12.5  & 14.6  & 16.4  & 21.9  & 14.7 \\
    \midrule
    500    & 3.3   & 2.8   & 17.7  & 29.0  & 38.7  & 12.4  & 9.5   & 15.7  & 15.6  & 16.1 \\
    1000    & 4.6   & 4.0   & 18.1  & 31.6  & 41.9  & 15.9  & 11.9  & 18.2  & 25.0  & 19.0 \\
    1500    & 5.2   & 5.4   & 21.1  & 32.9  & 43.1  & 15.3  & 11.1  & 20.4  & 12.5  & 18.6 \\
    2000    & 6.8   & 5.8   & 20.2  & 33.5  & 46.6  & 18.2  & 10.7  & 20.3  & 12.5  & 19.4 \\
    2500    & 7.1   & 3.8   & 20.7  & 37.0  & 48.6  & 18.3  & 12.0  & 21.4  & 18.8  & 20.9 \\
    3000    & 7.4   & 4.4   & 22.9  & 37.1  & 50.5  & 18.3  & 12.3  & 21.2  & 25.0  & 22.1 \\
    3500    & 8.8   & 4.8   & 22.8  & 39.4  & 53.3  & 19.2  & 12.0  & 22.8  & 34.4  & 24.2 \\
    4000    & 8.6   & 4.6   & 24.0  & 38.7  & 51.4  & 18.8  & 14.8  & 24.4  & 18.8  & 22.7 \\
    4500    & 8.6   & 4.2   & 24.2  & 39.2  & 53.6  & 20.4  & 13.5  & 23.9  & 18.8  & 22.9 \\
    5000    & 8.9   & 5.2   & 24.0  & 40.0  & 52.6  & 20.0  & 13.6  & 23.9  & 18.8  & 23.0 \\
    5500    & 8.0   & 6.2   & 23.2  & 41.4  & 55.0  & 22.3  & 14.3  & 24.9  & 25.0  & 24.5 \\
    6000    & 8.3   & 5.2   & 22.2  & 39.8  & 54.0  & 24.3  & 12.6  & 25.1  & 31.2  & 24.7 \\
    6500    & 9.4   & 5.6   & 24.4  & 40.2  & 54.5  & 20.3  & 13.0  & 24.9  & 31.2  & 24.8 \\
    7000    & 9.2   & 5.8   & 25.8  & 40.6  & 55.3  & 22.5  & 12.5  & 24.5  & 21.9  & 24.2 \\
    7500    & 9.3   & 7.4   & 23.4  & 41.9  & 55.6  & 22.1  & 14.6  & 24.1  & 25.0  & 24.8 \\
    \bottomrule
    \end{tabular}%
    }
  \label{tab:exp3-tlm-1.1b-owm-cpt-df}%
\end{table}

\begin{table}[htbp]
  \centering
  \caption{Full evaluation results of \tinylm continual pre-training on \owm with data after \method-D+C. Note that about 1B tokens are trained per 500 steps.}
  \resizebox{0.95\textwidth}{!}{
    \begin{tabular}{c|ccccccccc|c}
    \toprule
    \begin{tabular}[c]{@{}l@{}}\textbf{Train}\\ \textbf{Steps}\end{tabular} &
    \textbf{GSM8K} &
    \textbf{MATH} &
    \textbf{SVAMP} &
    \textbf{ASDiv} &
    \textbf{MAWPS} &
    \textbf{TAB} &
    \textbf{MQA} &
    \begin{tabular}[c]{@{}c@{}}\textbf{MMLU}\\ \textbf{STEM}\end{tabular} &
    \begin{tabular}[c]{@{}c@{}}\textbf{SAT}\\ \textbf{MATH}\end{tabular} &
    \textbf{AVG} \\
    \midrule
    0     & 2.8   & 3.2   & 10.9  & 18    & 20.2  & 12.5  & 14.6  & 16.4  & 21.9  & 14.7 \\ \midrule
    500    & 4.3   & 5.0   & 16.4  & 28.8  & 36.4  & 15.3  & 11.4  & 18.5  & 15.6  & 16.9 \\
    1000    & 5.5   & 3.8   & 20.5  & 34.6  & 44.6  & 15.3  & 12.1  & 19.6  & 28.1  & 20.5 \\
    1500    & 5.2   & 4.4   & 21.4  & 34.5  & 44.7  & 16.1  & 11.2  & 21.4  & 34.4  & 21.5 \\
    2000    & 6.3   & 5.4   & 20.1  & 33.7  & 46.2  & 19.4  & 10.5  & 21.2  & 12.5  & 19.5 \\
    2500    & 7.8   & 5.4   & 22.1  & 37.0  & 49.5  & 17.9  & 13.3  & 22.9  & 21.9  & 22.0 \\
    3000    & 6.4   & 3.4   & 23.0  & 38.6  & 51.1  & 18.5  & 12.6  & 24.3  & 18.8  & 21.9 \\
    3500    & 8.5   & 4.6   & 24.1  & 40.2  & 53.8  & 22.1  & 12.5  & 23.1  & 25.0  & 23.8 \\
    4000    & 8.2   & 6.0   & 24.1  & 41.0  & 52.4  & 19.8  & 10.2  & 26.1  & 31.2  & 24.3 \\
    4500    & 8.3   & 5.4   & 24.1  & 41.3  & 54.4  & 20.6  & 15.2  & 24.2  & 28.1  & 24.6 \\
    5000    & 8.5   & 7.0   & 26.0  & 40.5  & 54.9  & 21.7  & 13.9  & 25.5  & 34.4  & 25.8 \\
    5500    & 8.7   & 4.0   & 23.2  & 41.1  & 54.8  & 20.5  & 14.4  & 26.5  & 21.9  & 23.9 \\
    6000    & 8.3   & 5.0   & 24.8  & 41.3  & 54.3  & 23.2  & 14.0  & 25.3  & 25.0  & 24.6 \\
    6500    & 8.6   & 6.4   & 24.5  & 41.6  & 55.1  & 22.2  & 14.4  & 26.5  & 25.0  & 24.9 \\
    7000    & 8.9   & 6.0   & 23.4  & 40.5  & 53.4  & 22.0  & 15.8  & 27.3  & 28.1  & 25.0 \\
    7500    & 9.0   & 4.4   & 23.8  & 41.9  & 56.4  & 22.2  & 15.6  & 26.8  & 31.2  & 25.7 \\
    \bottomrule
    \end{tabular}%
    }
  \label{tab:exp3-tlm-1.1b-owm-cpt-df+lc}%
\end{table}

%% file: tables/exp-3-llama2-full-results.tex
\begin{table}[htbp]
  \centering
  \caption{Full evaluation results of \llamaii continual pre-training on \owm with raw data. Note that about 1B tokens are trained per 1000 steps.}
  \resizebox{0.95\textwidth}{!}{
    \begin{tabular}{c|ccccccccc|c}
    \toprule
    \begin{tabular}[c]{@{}l@{}}\textbf{Train}\\ \textbf{Steps}\end{tabular} &
    \textbf{GSM8K} &
    \textbf{MATH} &
    \textbf{SVAMP} &
    \textbf{ASDiv} &
    \textbf{MAWPS} &
    \textbf{TAB} &
    \textbf{MQA} &
    \begin{tabular}[c]{@{}c@{}}\textbf{MMLU}\\ \textbf{STEM}\end{tabular} &
    \begin{tabular}[c]{@{}c@{}}\textbf{SAT}\\ \textbf{MATH}\end{tabular} &
    \textbf{AVG} \\
    \midrule
0    & 14.1  & 3.8           & 39.5  & 51.6  & 63.6  & 30.9   & 12.5   & 32.9       & 34.4      & 31.5    \\ \midrule
1k   & 17.2  & 3.6           & 39.1  & 50.4  & 63.0  & 30.2   & 18.9   & 31.8       & 31.2      & 31.7    \\
2k  & 19.7  & 6.0           & 43.9  & 55.5  & 68.3  & 32.9   & 19.0   & 33.0       & 37.5      & 35.1    \\
3k  & 19.6  & 8.6           & 42.9  & 56.3  & 68.4  & 32.2   & 17.4   & 34.6       & 40.6      & 35.6    \\
4k  & 21.8  & 8.8           & 44.6  & 57.3  & 72.0  & 28.9   & 23.6   & 35.8       & 40.6      & 37.0    \\
5k  & 22.6  & 10.4          & 45.9  & 57.0  & 73.5  & 31.5   & 23.9   & 39.0       & 43.8      & 38.6    \\
6k  & 24.5  & 10.0          & 44.9  & 57.6  & 73.7  & 35.5   & 25.8   & 36.1       & 43.8      & 39.1    \\
7k  & 23.3  & 10.4          & 46.5  & 59.0  & 75.3  & 32.9   & 27.7   & 39.0       & 50.0      & 40.5    \\
8k  & 29.0  & 12.4          & 46.4  & 59.7  & 77.0  & 33.1   & 30.2   & 38.8       & 50.0      & 41.8    \\
9k  & 26.1  & 12.8          & 48.8  & 59.9  & 74.3  & 35.0   & 28.3   & 39.2       & 50.0      & 41.6    \\
10k & 29.6  & 13.6          & 49.2  & 61.9  & 78.4  & 36.3   & 31.9   & 40.5       & 43.8      & 42.8    \\ \bottomrule
\end{tabular}}
\label{tab:exp3-llama-2-7b-owm-cpt-raw}
\end{table}

\begin{table}[htbp]
  \centering
  \caption{Full evaluation results of \llamaii continual pre-training on \owm with \textbf{\method-D}. Note that about 1B tokens are trained per 1000 steps.}
  \resizebox{0.95\textwidth}{!}{
    \begin{tabular}{c|ccccccccc|c}
    \toprule
    \begin{tabular}[c]{@{}l@{}}\textbf{Train}\\ \textbf{Steps}\end{tabular} &
    \textbf{GSM8K} &
    \textbf{MATH} &
    \textbf{SVAMP} &
    \textbf{ASDiv} &
    \textbf{MAWPS} &
    \textbf{TAB} &
    \textbf{MQA} &
    \begin{tabular}[c]{@{}c@{}}\textbf{MMLU}\\ \textbf{STEM}\end{tabular} &
    \begin{tabular}[c]{@{}c@{}}\textbf{SAT}\\ \textbf{MATH}\end{tabular} &
    \textbf{AVG} \\
    \midrule
0    & 14.1  & 3.8           & 39.5  & 51.6  & 63.6  & 30.9   & 12.5   & 32.9       & 34.4      & 31.5    \\ \midrule
1k   & 17.1  & 7.2           & 39.8  & 51.6  & 68.4  & 31.4   & 21.4   & 35.2       & 40.6      & 34.7    \\
2k   & 21.9  & 9.2           & 43.2  & 57.0  & 72.8  & 33.1   & 24.0   & 37.6       & 56.2      & 39.4    \\
3k   & 20.5  & 10.8          & 45.7  & 58.6  & 76.2  & 35.3   & 25.8   & 38.3       & 53.1      & 40.5    \\
4k   & 27.2  & 11.8          & 45.7  & 58.7  & 76.6  & 35.9   & 29.2   & 41.0       & 31.2      & 39.7    \\
5k   & 28.9  & 14.2          & 49.3  & 60.2  & 77.9  & 38.8   & 32.8   & 41.7       & 53.1      & 44.1    \\
6k   & 31.9  & 15.0          & 51.5  & 62.0  & 79.0  & 39.2   & 33.3   & 41.4       & 68.8      & 46.9    \\
7k   & 31.5  & 16.8          & 51.9  & 63.2  & 77.9  & 36.5   & 35.9   & 43.8       & 43.8      & 44.6    \\
8k   & 30.3  & 13.8          & 51.9  & 63.7  & 80.6  & 38.3   & 36.1   & 41.3       & 59.4      & 46.2    \\
9k   & 30.6  & 14.0          & 52.7  & 62.6  & 78.7  & 37.5   & 36.1   & 43.2       & 43.8      & 44.4    \\
10k  & 30.3  & 16.0          & 54.2  & 63.8  & 79.5  & 37.3   & 37.2   & 44.2       & 46.9      & 45.5    \\
\bottomrule
\end{tabular}}
\label{tab:exp3-llama-2-7b-owm-cpt-df}
\end{table}

\begin{table}[htbp]
  \centering
  \caption{Full evaluation results of \llamaii continual pre-training on \owm with \textbf{\method-D+C}. Note that about 1B tokens are trained per 1000 steps.}
  \resizebox{0.95\textwidth}{!}{
    \begin{tabular}{c|ccccccccc|c}
    \toprule
    \begin{tabular}[c]{@{}l@{}}\textbf{Train}\\ \textbf{Steps}\end{tabular} &
    \textbf{GSM8K} &
    \textbf{MATH} &
    \textbf{SVAMP} &
    \textbf{ASDiv} &
    \textbf{MAWPS} &
    \textbf{TAB} &
    \textbf{MQA} &
    \begin{tabular}[c]{@{}c@{}}\textbf{MMLU}\\ \textbf{STEM}\end{tabular} &
    \begin{tabular}[c]{@{}c@{}}\textbf{SAT}\\ \textbf{MATH}\end{tabular} &
    \textbf{AVG} \\
    \midrule
0    & 14.1  & 3.8           & 39.5  & 51.6  & 63.6  & 30.9   & 12.5   & 32.9       & 34.4      & 31.5    \\ \midrule
1k   & 18.8  & 6.8           & 40.1  & 54.4  & 66.1  & 29.7   & 22.9   & 35.6       & 53.1      & 36.4    \\
2k   & 23.1  & 8.6           & 45.7  & 56.5  & 72.7  & 30.7   & 25.1   & 35.6       & 46.9      & 38.3    \\
3k   & 23.4  & 11.8          & 47.9  & 59.1  & 74.6  & 30.4   & 28.2   & 38.3       & 59.4      & 41.5    \\
4k   & 25.2  & 14.2          & 49.0  & 57.8  & 72.7  & 32.8   & 33.1   & 40.7       & 40.6      & 40.7    \\
5k   & 24.4  & 13.6          & 48.0  & 58.7  & 72.1  & 28.9   & 33.0   & 40.6       & 50.0      & 41.0    \\
6k   & 29.6  & 12.8          & 46.1  & 63.4  & 75.6  & 33.7   & 31.6   & 42.8       & 53.1      & 43.2    \\
7k   & 29.9  & 13.6          & 50.5  & 61.5  & 75.2  & 36.4   & 34.5   & 41.7       & 53.1      & 44.0    \\
8k   & 30.2  & 15.8          & 50.8  & 63.7  & 77.1  & 37.7   & 36.3   & 43.4       & 43.8      & 44.3    \\
9k   & 34.0  & 15.4          & 52.1  & 62.4  & 79.3  & 35.9   & 40.2   & 44.0       & 56.2      & 46.6    \\
10k  & 30.6  & 16.8          & 50.2  & 63.7  & 79.3  & 37.3   & 40.1   & 43.8       & 53.1      & 46.1    \\
\bottomrule
\end{tabular}}
\label{tab:exp3-llama-2-7b-owm-cpt-lc}
\end{table}

%% file: tables/exp-3-codellama-full-results.tex
\begin{table}[htbp]
  \centering
  \caption{Full evaluation results of \codellama-7B continual pre-training on \owm with raw data. Note that about 1B tokens are trained per 250 steps.}
  \resizebox{0.95\textwidth}{!}{
    \begin{tabular}{c|ccccccccc|c}
    \toprule
    \begin{tabular}[c]{@{}l@{}}\textbf{Train}\\ \textbf{Steps}\end{tabular} &
    \textbf{GSM8K} &
    \textbf{MATH} &
    \textbf{SVAMP} &
    \textbf{ASDiv} &
    \textbf{MAWPS} &
    \textbf{TAB} &
    \textbf{MQA} &
    \begin{tabular}[c]{@{}c@{}}\textbf{MMLU}\\ \textbf{STEM}\end{tabular} &
    \begin{tabular}[c]{@{}c@{}}\textbf{SAT}\\ \textbf{MATH}\end{tabular} &
    \textbf{AVG} \\ \midrule
 0    & 11.8  & 5.0           & 44.2  & 50.7  & 62.6  & 30.6   & 14.3   & 20.4       & 21.9      & 29.1    \\ \midrule
250   & 16.7  & 8.2           & 45.2  & 52.2  & 65.3  & 33.9   & 16.0   & 28.8       & 43.8      & 34.5    \\
500   & 18.3  & 7.8           & 43.1  & 53.9  & 69.0  & 29.3   & 15.3   & 22.5       & 37.5      & 33.0    \\
750   & 20.2  & 8.0           & 45.2  & 54.2  & 71.9  & 29.9   & 17.1   & 31.2       & 37.5      & 35.0    \\
1000   & 24.7  & 9.8           & 40.6  & 58.6  & 72.7  & 29.3   & 20.7   & 31.9       & 34.4      & 35.9    \\
1250   & 24.3  & 10.4          & 44.0  & 57.5  & 74.8  & 29.2   & 21.4   & 36.1       & 50.0      & 38.6    \\
1500   & 26.2  & 13.2          & 48.4  & 58.8  & 75.4  & 29.4   & 28.1   & 34.9       & 50.0      & 40.5    \\
1750   & 25.5  & 11.8          & 49.1  & 58.7  & 76.6  & 32.4   & 26.7   & 37.3       & 43.8      & 40.2    \\
2000   & 28.0  & 13.6          & 46.3  & 61.7  & 80.0  & 33.8   & 29.4   & 37.2       & 50.0      & 42.2    \\
2250   & 27.7  & 13.6          & 48.9  & 62.2  & 80.3  & 32.5   & 28.9   & 39.1       & 59.4      & 43.6    \\
2500  & 31.1  & 14.8          & 51.4  & 62.1  & 81.2  & 33.6   & 30.4   & 40.5       & 43.8      & 43.2  \\ 
 \bottomrule
\end{tabular}}
\label{tab:exp3-codellama-7b-owm-cpt-raw}%
\end{table}

\begin{table}[htbp]
  \centering
  \caption{Full evaluation results of \codellama continual pre-training on \owm with \textbf{\method-D}. Note that about 1B tokens are trained per 250 steps.}
  \resizebox{0.95\textwidth}{!}{
    \begin{tabular}{c|ccccccccc|c}
    \toprule
    \begin{tabular}[c]{@{}l@{}}\textbf{Train}\\ \textbf{Steps}\end{tabular} &
    \textbf{GSM8K} &
    \textbf{MATH} &
    \textbf{SVAMP} &
    \textbf{ASDiv} &
    \textbf{MAWPS} &
    \textbf{TAB} &
    \textbf{MQA} &
    \begin{tabular}[c]{@{}c@{}}\textbf{MMLU}\\ \textbf{STEM}\end{tabular} &
    \begin{tabular}[c]{@{}c@{}}\textbf{SAT}\\ \textbf{MATH}\end{tabular} &
    \textbf{AVG} \\ \midrule
    0    & 11.8  & 5.0           & 44.2  & 50.7  & 62.6  & 30.6   & 14.3   & 20.4       & 21.9      & 29.1    \\ \midrule
250   & 21.1  & 9.2           & 48.7  & 56.1  & 71.3  & 33.4   & 22.2   & 34.1       & 50.0      & 38.5    \\
500   & 23.7  & 11.6          & 49.8  & 57.4  & 74.7  & 32.9   & 28.5   & 35.8       & 59.4      & 41.5    \\
750   & 25.1  & 15.4          & 48.1  & 58.9  & 78.8  & 36.8   & 29.4   & 37.6       & 53.1      & 42.6    \\
1000   & 28.4  & 14.2          & 50.9  & 61.2  & 79.8  & 36.7   & 27.7   & 37.6       & 50.0      & 42.9    \\
1250   & 33.0  & 15.2          & 49.3  & 62.9  & 81.1  & 33.4   & 32.8   & 41.0       & 46.9      & 44.0    \\
1500   & 36.0  & 15.0          & 54.2  & 65.0  & 81.0  & 39.3   & 34.1   & 42.0       & 62.5      & 47.7    \\
1750   & 34.7  & 14.6          & 53.1  & 63.6  & 83.3  & 40.6   & 35.9   & 43.4       & 62.5      & 48.0    \\
2000   & 35.7  & 17.6          & 53.3  & 65.4  & 83.5  & 42.4   & 37.1   & 42.4       & 56.2      & 48.2    \\
2250   & 37.2  & 18.8          & 54.5  & 65.4  & 83.2  & 41.9   & 41.0   & 44.9       & 71.9      & 51.0    \\
2500  & 38.1  & 17.0          & 54.2  & 67.0  & 83.1  & 40.9   & 39.8   & 43.7       & 50.0      & 48.2  \\ \bottomrule
\end{tabular}}
\label{tab:exp3-codellama-7b-owm-cpt-df}%
\end{table}

\begin{table}[htbp]
  \centering
  \caption{Full evaluation results of \codellama continual pre-training on \owm with \textbf{\method-D+C}. Note that about 1B tokens are trained per 250 steps.}
  \resizebox{0.95\textwidth}{!}{
    \begin{tabular}{c|ccccccccc|c}
    \toprule
    \begin{tabular}[c]{@{}l@{}}\textbf{Train}\\ \textbf{Steps}\end{tabular} &
    \textbf{GSM8K} &
    \textbf{MATH} &
    \textbf{SVAMP} &
    \textbf{ASDiv} &
    \textbf{MAWPS} &
    \textbf{TAB} &
    \textbf{MQA} &
    \begin{tabular}[c]{@{}c@{}}\textbf{MMLU}\\ \textbf{STEM}\end{tabular} &
    \begin{tabular}[c]{@{}c@{}}\textbf{SAT}\\ \textbf{MATH}\end{tabular} &
    \textbf{AVG} \\ \midrule
    0    & 11.8  & 5.0           & 44.2  & 50.7  & 62.6  & 30.6   & 14.3   & 20.4       & 21.9      & 29.1    \\ \midrule
250   & 18.1  & 10.2          & 46.0  & 54.5  & 71.9  & 33.0   & 21.3   & 34.4       & 50.0      & 37.7    \\
500   & 22.4  & 10.0          & 50.3  & 59.7  & 76.4  & 31.3   & 26.1   & 36.0       & 59.4      & 41.3    \\
750   & 26.8  & 11.4          & 51.2  & 61.0  & 78.5  & 34.9   & 26.4   & 38.0       & 53.1      & 42.4    \\
1000   & 29.0  & 14.4          & 54.1  & 62.8  & 80.1  & 36.9   & 34.2   & 40.4       & 62.5      & 46.0    \\
1250   & 31.4  & 15.0          & 51.7  & 63.8  & 81.1  & 37.2   & 32.5   & 41.4       & 75.0      & 47.7    \\
1500   & 31.5  & 17.4          & 53.4  & 64.4  & 80.7  & 39.6   & 35.4   & 41.6       & 71.9      & 48.4    \\
1750   & 33.7  & 15.2          & 50.6  & 64.3  & 81.5  & 39.2   & 36.1   & 40.5       & 53.1      & 46.0    \\
2000   & 36.2  & 16.0          & 54.7  & 65.1  & 83.1  & 39.9   & 39.1   & 43.4       & 71.9      & 49.9    \\
2250   & 37.1  & 16.6          & 55.3  & 65.6  & 82.4  & 41.3   & 36.5   & 42.7       & 75.0      & 50.3    \\
2500  & 35.6  & 17.6          & 55.8  & 67.9  & 82.7  & 41.3   & 38.9   & 42.6       & 62.5      & 49.4   \\
\bottomrule
\end{tabular}}
\label{tab:exp3-codellama-7b-owm-cpt-lc}%
\end{table}

%% file: tables/exp-3-mistral-full-df-results.tex
\begin{table}[htbp]
  \centering
  \caption{Full evaluation results of \mistral-7B continual pre-training on \owm with raw data. Note that about 1B tokens are trained per 1000 steps.}
  \resizebox{0.95\textwidth}{!}{
    \begin{tabular}{c|ccccccccc|c}
    \toprule
    \begin{tabular}[c]{@{}l@{}}\textbf{Train}\\ \textbf{Steps}\end{tabular} &
    \textbf{GSM8K} &
    \textbf{MATH} &
    \textbf{SVAMP} &
    \textbf{ASDiv} &
    \textbf{MAWPS} &
    \textbf{TAB} &
    \textbf{MQA} &
    \begin{tabular}[c]{@{}c@{}}\textbf{MMLU}\\ \textbf{STEM}\end{tabular} &
    \begin{tabular}[c]{@{}c@{}}\textbf{SAT}\\ \textbf{MATH}\end{tabular} &
    \textbf{AVG} \\
    \midrule
    0     & 40.6  & 11.4  & 65.4  & 68.5  & 87.0  & 52.9  & 32.3  & 50.0  & 56.2  & 51.6 \\ \midrule
    1k    & 31.6  & 12.0  & 56.5  & 66.0  & 80.1  & 43.9  & 27.1  & 45.1  & 56.2  & 46.5 \\
    2k    & 32.4  & 10.8  & 54.7  & 63.5  & 82.6  & 40.8  & 31.6  & 45.7  & 59.4  & 46.8 \\
    3k    & 33.6  & 14.8  & 60.4  & 64.7  & 84.5  & 43.5  & 33.1  & 47.2  & 68.8  & 50.1 \\
    4k    & 35.1  & 14.8  & 58.7  & 65.2  & 84.4  & 41.2  & 38.5  & 47.3  & 62.5  & 49.7 \\
    5k    & 33.4  & 16.0  & 59.3  & 65.0  & 83.8  & 46.7  & 34.6  & 49.1  & 62.5  & 50.0 \\
    6k    & 38.7  & 16.6  & 61.5  & 68.1  & 86.1  & 47.4  & 35.3  & 48.5  & 37.5  & 48.9 \\
    7k    & 39.6  & 17.2  & 60.5  & 68.2  & 86.2  & 44.4  & 38.5  & 49.3  & 53.1  & 50.8 \\
    8k    & 44.0  & 16.4  & 64.5  & 69.8  & 88.7  & 45.5  & 41.3  & 50.6  & 59.4  & 53.4 \\
    9k    & 43.9  & 19.4  & 63.7  & 69.7  & 87.6  & 44.9  & 42.9  & 51.0  & 62.5  & 54.0 \\
    10k   & 44.4  & 19.2  & 65.2  & 69.6  & 88.4  & 46.6  & 43.1  & 50.8  & 65.6  & 54.8 \\
    \bottomrule
    \end{tabular}%
    }
  \label{tab:exp3-mistral-7b-owm-cpt-raw}%
\end{table}

\begin{table}[htbp]
  \centering
  \caption{Full evaluation results of \mistral-7B continual pre-training on \owm with \textbf{\method-D}. Note that about 1B tokens are trained per 1000 steps.}
  \resizebox{0.95\textwidth}{!}{
    \begin{tabular}{c|ccccccccc|c}
    \toprule
    \begin{tabular}[c]{@{}l@{}}\textbf{Train}\\ \textbf{Steps}\end{tabular} &
    \textbf{GSM8K} &
    \textbf{MATH} &
    \textbf{SVAMP} &
    \textbf{ASDiv} &
    \textbf{MAWPS} &
    \textbf{TAB} &
    \textbf{MQA} &
    \begin{tabular}[c]{@{}c@{}}\textbf{MMLU}\\ \textbf{STEM}\end{tabular} &
    \begin{tabular}[c]{@{}c@{}}\textbf{SAT}\\ \textbf{MATH}\end{tabular} &
    \textbf{AVG} \\
    \midrule
    0     & 40.6  & 11.4  & 65.4  & 68.5  & 87.0  & 52.9  & 32.3  & 50.0  & 56.2  & 51.6 \\ \midrule
    1k    & 36.8  & 14.6  & 57.2  & 66.1  & 83.1  & 45.7  & 32.6  & 47.7  & 59.4  & 49.2 \\
    2k    & 38.5  & 17.0  & 57.9  & 69.0  & 86.3  & 44.7  & 33.6  & 49.2  & 56.2  & 50.3 \\
    3k    & 40.0  & 19.0  & 59.3  & 68.7  & 87.0  & 46.8  & 41.0  & 48.0  & 68.8  & 53.2 \\
    4k    & 38.5  & 20.4  & 59.3  & 66.2  & 85.1  & 42.6  & 42.8  & 49.5  & 68.8  & 52.6 \\
    5k    & 42.5  & 20.2  & 63.0  & 70.5  & 86.6  & 47.2  & 43.4  & 49.8  & 62.5  & 54.0 \\
    6k    & 46.8  & 17.8  & 62.5  & 72.7  & 88.2  & 51.2  & 47.7  & 51.3  & 56.2  & 54.9 \\
    7k    & 47.5  & 22.4  & 64.1  & 71.8  & 89.1  & 51.4  & 47.9  & 52.4  & 65.6  & 56.9 \\
    8k    & 44.6  & 23.8  & 63.2  & 70.8  & 87.7  & 47.6  & 49.1  & 54.1  & 65.6  & 56.3 \\
    9k    & 46.6  & 24.6  & 61.6  & 72.3  & 86.4  & 46.9  & 49.8  & 53.2  & 65.6  & 56.3 \\
    10k   & 46.7  & 22.6  & 63.5  & 72.4  & 88.9  & 48.3  & 48.2  & 54.1  & 62.5  & 56.4 \\
    \bottomrule
    \end{tabular}%
    }
  \label{tab:exp3-mistral-7b-owm-cpt-df}%
\end{table}

\begin{table}[htbp]
  \centering
  \caption{Full evaluation results of Mistral-7B continual pre-training on \owm with \textbf{\method-D+C}. Note that about 1B tokens are trained per 1000 steps.}
  \resizebox{0.95\textwidth}{!}{
    \begin{tabular}{c|ccccccccc|c}
    \toprule
    \begin{tabular}[c]{@{}l@{}}\textbf{Train}\\ \textbf{Steps}\end{tabular} &
    \textbf{GSM8K} &
    \textbf{MATH} &
    \textbf{SVAMP} &
    \textbf{ASDiv} &
    \textbf{MAWPS} &
    \textbf{TAB} &
    \textbf{MQA} &
    \begin{tabular}[c]{@{}c@{}}\textbf{MMLU}\\ \textbf{STEM}\end{tabular} &
    \begin{tabular}[c]{@{}c@{}}\textbf{SAT}\\ \textbf{MATH}\end{tabular} &
    \textbf{AVG} \\
    \midrule
    0     & 40.6  & 11.4  & 65.4  & 68.5  & 87.0  & 52.9  & 32.3  & 50.0  & 56.2  & 51.6 \\ \midrule
    1k    & 30.9  & 16.0  & 60.1  & 64.5  & 85.3  & 40.8  & 33.9  & 48.0  & 59.4  & 48.8 \\
    2k    & 40.3  & 17.6  & 63.0  & 66.3  & 86.2  & 48.0  & 33.9  & 48.7  & 53.1  & 50.8 \\
    3k    & 42.4  & 17.8  & 59.6  & 69.1  & 85.7  & 50.1  & 38.5  & 49.9  & 59.4  & 52.5 \\
    4k    & 43.8  & 20.4  & 63.7  & 69.3  & 88.2  & 46.2  & 46.3  & 50.9  & 65.6  & 54.9 \\
    5k    & 42.5  & 18.4  & 59.3  & 69.6  & 87.9  & 44.3  & 46.1  & 51.9  & 65.6  & 54.0 \\
    6k    & 47.7  & 21.8  & 62.7  & 71.7  & 89.2  & 47.9  & 48.4  & 54.0  & 68.8  & 56.9 \\
    7k    & 46.8  & 21.6  & 62.9  & 72.1  & 88.4  & 50.1  & 46.4  & 52.5  & 68.8  & 56.6 \\
    8k    & 48.4  & 21.6  & 65.0  & 72.7  & 89.2  & 51.1  & 49.4  & 52.9  & 65.6  & 57.3 \\
    9k    & 48.5  & 24.8  & 64.4  & 72.6  & 88.3  & 50.7  & 48.1  & 53.4  & 62.5  & 57.0 \\
    10k   & 51.0  & 22.4  & 64.9  & 72.9  & 89.2  & 49.8  & 53.0  & 54.2  & 75.0  & 59.2 \\
    \bottomrule
    \end{tabular}%
    }
  \label{tab:exp3-mistral-7b-owm-cpt-lc}%
\end{table}

%% file: tables/case_study_redpj_df_lf_all_programs.tex
\begin{table*}[ht]
\centering
\caption{Cases from \redpj after applying \method. Text in \textcolor{red}{red} indicates content to be removed or replaced. ``\texttt{...}'' denotes omitted content due to limited space.}
\label{tab:case_study_redpj}
\begin{small}
\begin{tabular}{p{5.8in}}
\toprule

\multicolumn{1}{c}{\cellcolor[HTML]{F2F2F2}Case 1}\\ \midrule

TagCollegeEducationJournalismWar
\vspace{2mm}

: Michael Lewis
\vspace{2mm}

ContributorMichael Lewis
\vspace{2mm}

Michael Lewis is possibly the most entertaining nonfiction writer alive. If that's not true it's at least close to true. Liar's Poker, Moneyball, The Blind Side, his NYT article about Jonathan Lebed (Google it): what's not to love?
\vspace{2mm}

504: How I Got Into College
\vspace{2mm}

Act Two: My Ames is True
\vspace{2mm}

Writer Michael Lewis tells the story of a man named Emir Kamenica, whose path to college started with fleeing the war in Bosnia and becoming a refugee in the United States. Then he had a stroke of luck: a student teacher read an essay he’d plagiarized from a book he’d stolen from a library back in Bosnia, and was so impressed that she got him out of a bad high school and into a much better one.
\vspace{2mm}

Act Three
\vspace{2mm}

Michael Lewis’ story continues, and he figures out why Emir Kamenica insists on remembering, and telling, the story of his life the way he does — even when he finds out that some of the facts may be wrong.

\\
\HL{\textbf{Output by \method :}}

\hspace{1mm} \texttt{drop\_doc()} \\ \midrule

\multicolumn{1}{c}{\cellcolor[HTML]{F2F2F2}Case 2}\\ \midrule

\begin{tabular}{p{15cm}}

\textcolor{red}{Home > Staff > Staff search > Dr Tim Overton} \\
\textcolor{red}{Dr Tim Overton BSc PhD} \\
\textcolor{red}{School of Chemical EngineeringSenior Lecturer} \\
\textcolor{red}{Telephone (+44) (0) 121 414 5306Emailt.w.overton@bham.ac.uk} \\
\textcolor{red}{AddressSchool of Chemical EngineeringUniversity of Birmingham} \\
B15 2TT \\
Dr Tim Overton is a biochemist and molecular microbiologist who is interested in applying molecular biology and single-cell techniques to understand and develop bioprocesses. He is active in microbial flow cytometry research and collaborates widely with bioprocess engineers, molecular microbiologists, cell biologists and environmental microbiologists to develop new methods of answering fundamental questions on a single-cell level. \\
His research also focuses on using bacteria to make useful products such as protein drugs and small molecules, and the bacterial responses to stress encountered in such processes. Current and recent research funding has come from the BBSRC, TSB and EU FP7. He is the director of the MSc in Biochemical Engineering. \textcolor{red}{Pages:  1    3  4}

\vspace{2mm}
\texttt{...}
\vspace{2mm}

Google scholar: \textcolor{red}{http://scholar.google.co.uk/citations?user=tF\_eBKEAAAAJ}

\texttt{...}

\end{tabular} \\

\HL{\textbf{Output by \method :}}

\hspace{1mm}\texttt{keep\_doc()}

\hspace{1mm}\texttt{remove\_lines(line\_start=0, line\_end=5)}

\hspace{1mm}\texttt{normalize(source\_str="http://scholar.google.co.uk/citations?user", target\_str="")}

\hspace{1mm}\texttt{normalize(source\_str="Pages:  1    3  4", target\_str="")}

\hspace{1mm} \texttt{...} \\

\bottomrule
\end{tabular}%
\end{small}

\end{table*}

%% file: tables/case_study_owm_df_lf_all_programs.tex
\begin{table*}[ht]
\centering
\caption{Cases from \owm after applying \method. Text in \textcolor{red}{red} indicates content to be removed or replaced. ``\texttt{...}'' denotes omitted content due to limited space.}
\label{tab:case_study_owm}
\begin{small}
\begin{tabular}{p{5.8in}}
\toprule
\multicolumn{1}{c}{\cellcolor[HTML]{F2F2F2}Case 1}\\ \midrule

\#\# unhybridized pi bonds
\vspace{2mm}

$sp, sp^{2}, sp^{3}, dsp^{3}, d^{2}sp^{3}$
\vspace{2mm}

Tatiana 4B
\vspace{2mm}

Posts: 30
\vspace{2mm}

Joined: Fri Sep 28, 2018 12:28 am
\vspace{2mm}

\#\#\# unhybridized pi bonds
\vspace{2mm}

\texttt{...}

\vspace{2mm}

\#\#\# Re: unhybridized pi bonds
\vspace{2mm}

I am not too sure in my knowledge about this, but I think that both have hybridized orbitals. Since hybridization is defined as the phenomenon of intermixing of the orbitals such as sp, sigma and pi bonds are just different types of covalent bonds formed depending on the way the atomic orbitals hybridize with each other. Sigma bonds are a result of when the overlap of orbitals of two atoms takes place along the line joining the two orbitals, while pi bonds are when two atoms overlap due to the sideways overlap of their 'p' orbitals.
\vspace{2mm}

Hannah Yates 1K
\vspace{2mm}

Posts: 59
\vspace{2mm}

Joined: Fri Sep 28, 2018 12:27 am
\vspace{2mm}

\#\#\# Re: unhybridized pi bonds
\vspace{2mm}

I am also not too sure on my answer, but I am pretty sure that a sigma bond has just hybridized orbitals, but the reason a pi bond can form is because of an extra (not hybridized) p orbital. This allows for a double and triple bond to form.

\\

\HL{\textbf{Output by \method :}}

\hspace{1mm} \texttt{drop\_doc()} \\ \midrule

\multicolumn{1}{c}{\cellcolor[HTML]{F2F2F2}Case 2}\\ \midrule

\begin{tabular}{p{10cm}}
\textcolor{red}{Solution - Trigonometric Identities} \\
\\
\textcolor{red}{Account} \\
\textcolor{red}{Register} \\
 \\
\textcolor{red}{Share} \\
\\
Books Shortlist \\
ConceptTrigonometric Identities \\
\\
Question \\
\\
Prove the following trigonometric identities: \\
\\
$(\text{i})\text{ }\frac{\sin \theta }{1-\cos \theta }=\text{cosec}\theta+\cot \theta$
\\
Solution \\
\\
\textcolor{red}{You need to to view the solution} \\
\textcolor{red}{Is there an error in this question or solution?} \\
\\
\textcolor{red}{Reference Material} \\
\\
\textcolor{red}{Solution for concept: Trigonometric Identities. For the course CBSE} \\
\textcolor{red}{S}  
\end{tabular} 

\\

\HL{\textbf{Output by \method :}}

\hspace{1mm} \texttt{keep\_doc()} 

\hspace{1mm} \texttt{remove\_lines(line\_start=0, line\_end=7)}

\hspace{1mm} \texttt{remove\_lines(line\_start=18, line\_end=24)}  \\ 

\bottomrule
\end{tabular}%
\end{small}

\end{table*}

%% file: tables/case_study_redpj_failure_cases.tex
\begin{table*}[ht]
\centering
\caption{{Failure cases from \redpj during applying \method. ``\texttt{...}'' denotes omitted content due to limited space. The notation like \texttt{[004]} is used to indicate the line number.}}
\label{tab:case_study_redpj_failure_cases}
\begin{small}
{
\begin{tabular}{p{5.8in}}
\toprule

\multicolumn{1}{c}{\cellcolor[HTML]{F2F2F2}Case 1: Repeated output (or Empty output)}\\ \midrule

\vspace{2mm}
\texttt{...}
\vspace{2mm}

\texttt{[004]} P: 114 1. The problem statement, all variables and given/known data Mercury is poured into a U-tube as in Figure P15.18a....Basically I don't understand why you would know to set the two volumes equal to each other? How do you know the volumes are the same?

\vspace{2mm}
\texttt{...}
\vspace{2mm}

\texttt{[007]} Related Discussions Mechanical Engineering 6 Introductory Physics Homework 0 General Engineering 1 Introductory Physics Homework 2 Introductory Physics Homework 2

\\
\HL{\textbf{Output by \method :}}

\hspace{1mm} \texttt{remove\_lines(start=1, end=1)} 

\hspace{1mm} \texttt{remove\_lines(start=6, end=6)}

\hspace{1mm} \texttt{remove\_lines(start=7, end=7)} 

\hspace{1mm} \texttt{remove\_lines(start=7, end=7)} 

\hspace{1mm} \texttt{remove\_lines(start=7, end=7)} 

\hspace{1mm} \texttt{remove\_lines(start=7, end} \\ \midrule

\multicolumn{1}{c}{\cellcolor[HTML]{F2F2F2}Case 2: Non-existent target string}\\ \midrule

\begin{tabular}{p{15cm}}

\vspace{2mm}
\texttt{...}
\vspace{2mm}

\texttt{[195]} 18. Sathyamoorthi, C. R., Mbekomize, C., Mapharing, M., \& Selinkie, P. (2018). The Impact of Corporate Governance on Working Capital Management Efficiency: Evidence from the Listed Companies in the Consumer Services Sector in Botswana. International Journal of Economics and Finance, 10, 135. https://doi.org/10.5539/ijef.v10n12p135

\vspace{2mm}

\texttt{[196]} 19. Vu, T. M. T., Tran, C. Q., Doan, D. T., \& Le, T. N. (2020). Determinants of Capital Structure: The Case in Vietnam. Journal of Asian Finance, Economics, And Business, 7(9), 159-168. https://doi.org/10.13106/jafeb.2020.vol7.no9.159

\vspace{2mm}
\texttt{...}
\vspace{2mm}

\end{tabular} \\

\HL{\textbf{Output by \method :}}

\# Analysis: this `source\_str` can not be found in the original text
\hspace{1mm}\texttt{normalize(source\_str="https://doi.org/10.13106/jafeb.2020.vol6.no2.53", target\_str="")} \\

\bottomrule
\end{tabular}%
}
\end{small}

\end{table*}

%% file: iclr2024_conference.bbl
\begin{thebibliography}{88}
\providecommand{\natexlab}[1]{#1}
\providecommand{\url}[1]{\texttt{#1}}
\expandafter\ifx\csname urlstyle\endcsname\relax
  \providecommand{\doi}[1]{doi: #1}\else
  \providecommand{\doi}{doi: \begingroup \urlstyle{rm}\Url}\fi

\bibitem[Meta(2024)]{metallama3}
Meta.
\newblock Introducing meta llama 3: The most capable openly available llm to date, 2024.
\newblock URL \url{https://ai.meta.com/blog/meta-llama-3}.

\bibitem[Achiam et~al.(2023)Achiam, Adler, Agarwal, Ahmad, Akkaya, Aleman, Almeida, Altenschmidt, Altman, Anadkat, et~al.]{achiam2023gpt}
Josh Achiam, Steven Adler, Sandhini Agarwal, Lama Ahmad, Ilge Akkaya, Florencia~Leoni Aleman, Diogo Almeida, Janko Altenschmidt, Sam Altman, Shyamal Anadkat, et~al.
\newblock Gpt-4 technical report.
\newblock \emph{arXiv preprint arXiv:2303.08774}, 2023.

\bibitem[Anthropic(2024)]{anthropic2024claude}
AI~Anthropic.
\newblock The claude 3 model family: Opus, sonnet, haiku.
\newblock \emph{Claude-3 Model Card}, 2024.
\newblock URL \url{https://www-cdn.anthropic.com/de8ba9b01c9ab7cbabf5c33b80b7bbc618857627/Model_Card_Claude_3.pdf}.

\bibitem[Reid et~al.(2024)Reid, Savinov, Teplyashin, Lepikhin, Lillicrap, Alayrac, Soricut, Lazaridou, Firat, Schrittwieser, et~al.]{reid2024gemini}
Machel Reid, Nikolay Savinov, Denis Teplyashin, Dmitry Lepikhin, Timothy Lillicrap, Jean-baptiste Alayrac, Radu Soricut, Angeliki Lazaridou, Orhan Firat, Julian Schrittwieser, et~al.
\newblock Gemini 1.5: Unlocking multimodal understanding across millions of tokens of context.
\newblock \emph{arXiv preprint arXiv:2403.05530}, 2024.

\bibitem[Yuan et~al.(2022)Yuan, Coenen, Reif, and Ippolito]{yuan2022wordcraft}
Ann Yuan, Andy Coenen, Emily Reif, and Daphne Ippolito.
\newblock Wordcraft: story writing with large language models.
\newblock In \emph{27th International Conference on Intelligent User Interfaces}, pages 841--852, 2022.

\bibitem[Wei et~al.(2022)Wei, Wang, Schuurmans, Bosma, Xia, Chi, Le, Zhou, et~al.]{wei2022chain}
Jason Wei, Xuezhi Wang, Dale Schuurmans, Maarten Bosma, Fei Xia, Ed~Chi, Quoc~V Le, Denny Zhou, et~al.
\newblock Chain-of-thought prompting elicits reasoning in large language models.
\newblock \emph{Advances in neural information processing systems}, 35:\penalty0 24824--24837, 2022.

\bibitem[Kojima et~al.(2022)Kojima, Gu, Reid, Matsuo, and Iwasawa]{kojima2022large}
Takeshi Kojima, Shixiang~Shane Gu, Machel Reid, Yutaka Matsuo, and Yusuke Iwasawa.
\newblock Large language models are zero-shot reasoners.
\newblock \emph{Advances in neural information processing systems}, 35:\penalty0 22199--22213, 2022.

\bibitem[Fan et~al.(2022)Fan, Wang, Jiang, Mandlekar, Yang, Zhu, Tang, Huang, Zhu, and Anandkumar]{fan2022minedojo}
Linxi Fan, Guanzhi Wang, Yunfan Jiang, Ajay Mandlekar, Yuncong Yang, Haoyi Zhu, Andrew Tang, De-An Huang, Yuke Zhu, and Anima Anandkumar.
\newblock Minedojo: Building open-ended embodied agents with internet-scale knowledge.
\newblock \emph{Advances in Neural Information Processing Systems}, 35:\penalty0 18343--18362, 2022.

\bibitem[Park et~al.(2023)Park, O'Brien, Cai, Morris, Liang, and Bernstein]{park2023generative}
Joon~Sung Park, Joseph O'Brien, Carrie~Jun Cai, Meredith~Ringel Morris, Percy Liang, and Michael~S Bernstein.
\newblock Generative agents: Interactive simulacra of human behavior.
\newblock In \emph{Proceedings of the 36th Annual ACM Symposium on User Interface Software and Technology}, pages 1--22, 2023.

\bibitem[Together(2023)]{together2023redpajama}
Together.
\newblock Redpajama: an open dataset for training large language models, October 2023.
\newblock URL \url{https://github.com/togethercomputer/RedPajama-Data}.

\bibitem[Penedo et~al.(2024{\natexlab{a}})Penedo, Kydl{\'\i}{\v{c}}ek, Lozhkov, Mitchell, Raffel, Von~Werra, Wolf, et~al.]{penedo2024fineweb}
Guilherme Penedo, Hynek Kydl{\'\i}{\v{c}}ek, Anton Lozhkov, Margaret Mitchell, Colin Raffel, Leandro Von~Werra, Thomas Wolf, et~al.
\newblock The fineweb datasets: Decanting the web for the finest text data at scale.
\newblock \emph{arXiv preprint arXiv:2406.17557}, 2024{\natexlab{a}}.

\bibitem[Rae et~al.(2021)Rae, Borgeaud, Cai, Millican, Hoffmann, Song, Aslanides, Henderson, Ring, Young, et~al.]{rae2021gopher}
Jack~W Rae, Sebastian Borgeaud, Trevor Cai, Katie Millican, Jordan Hoffmann, Francis Song, John Aslanides, Sarah Henderson, Roman Ring, Susannah Young, et~al.
\newblock Scaling language models: Methods, analysis \& insights from training gopher.
\newblock \emph{arXiv preprint arXiv:2112.11446}, 2021.

\bibitem[Soldaini et~al.(2024)Soldaini, Kinney, Bhagia, Schwenk, Atkinson, Authur, Bogin, Chandu, Dumas, Elazar, Hofmann, Jha, Kumar, Lucy, Lyu, Lambert, Magnusson, Morrison, Muennighoff, Naik, Nam, Peters, Ravichander, Richardson, Shen, Strubell, Subramani, Tafjord, Walsh, Zettlemoyer, Smith, Hajishirzi, Beltagy, Groeneveld, Dodge, and Lo]{soldaini-etal-2024-dolma}
Luca Soldaini, Rodney Kinney, Akshita Bhagia, Dustin Schwenk, David Atkinson, Russell Authur, Ben Bogin, Khyathi Chandu, Jennifer Dumas, Yanai Elazar, Valentin Hofmann, Ananya Jha, Sachin Kumar, Li~Lucy, Xinxi Lyu, Nathan Lambert, Ian Magnusson, Jacob Morrison, Niklas Muennighoff, Aakanksha Naik, Crystal Nam, Matthew Peters, Abhilasha Ravichander, Kyle Richardson, Zejiang Shen, Emma Strubell, Nishant Subramani, Oyvind Tafjord, Evan Walsh, Luke Zettlemoyer, Noah Smith, Hannaneh Hajishirzi, Iz~Beltagy, Dirk Groeneveld, Jesse Dodge, and Kyle Lo.
\newblock Dolma: an open corpus of three trillion tokens for language model pretraining research.
\newblock In Lun-Wei Ku, Andre Martins, and Vivek Srikumar, editors, \emph{Proceedings of the 62nd Annual Meeting of the Association for Computational Linguistics (Volume 1: Long Papers)}, pages 15725--15788, Bangkok, Thailand, August 2024. Association for Computational Linguistics.
\newblock URL \url{https://aclanthology.org/2024.acl-long.840}.

\bibitem[Zhang et~al.(2024{\natexlab{a}})Zhang, Qu, Liu, Zhang, Lin, Yu, Pan, Cheng, Liu, Lin, et~al.]{zhang2024map}
Ge~Zhang, Scott Qu, Jiaheng Liu, Chenchen Zhang, Chenghua Lin, Chou~Leuang Yu, Danny Pan, Esther Cheng, Jie Liu, Qunshu Lin, et~al.
\newblock Map-neo: Highly capable and transparent bilingual large language model series.
\newblock \emph{arXiv preprint arXiv:2405.19327}, 2024{\natexlab{a}}.

\bibitem[Xie et~al.(2023)Xie, Santurkar, Ma, and Liang]{xie2023data}
Sang~Michael Xie, Shibani Santurkar, Tengyu Ma, and Percy~S Liang.
\newblock Data selection for language models via importance resampling.
\newblock \emph{Advances in Neural Information Processing Systems}, 36:\penalty0 34201--34227, 2023.

\bibitem[Wettig et~al.(2024)Wettig, Gupta, Malik, and Chen]{wettig2024qurating}
Alexander Wettig, Aatmik Gupta, Saumya Malik, and Danqi Chen.
\newblock {QuRating}: Selecting high-quality data for training language models.
\newblock In \emph{International Conference on Machine Learning (ICML)}, 2024.

\bibitem[Yu et~al.(2024)Yu, Das, and Xiong]{yu2024mates}
Zichun Yu, Spandan Das, and Chenyan Xiong.
\newblock Mates: Model-aware data selection for efficient pretraining with data influence models.
\newblock \emph{arXiv preprint arXiv:2406.06046}, 2024.

\bibitem[Dubey et~al.(2024)Dubey, Jauhri, Pandey, Kadian, Al-Dahle, Letman, Mathur, Schelten, Yang, Fan, et~al.]{dubey2024llama3}
Abhimanyu Dubey, Abhinav Jauhri, Abhinav Pandey, Abhishek Kadian, Ahmad Al-Dahle, Aiesha Letman, Akhil Mathur, Alan Schelten, Amy Yang, Angela Fan, et~al.
\newblock The llama 3 herd of models.
\newblock \emph{arXiv preprint arXiv:2407.21783}, 2024.

\bibitem[Gunasekar et~al.(2023)Gunasekar, Zhang, Aneja, Mendes, Del~Giorno, Gopi, Javaheripi, Kauffmann, de~Rosa, Saarikivi, et~al.]{gunasekar2023textbooks-phi1}
Suriya Gunasekar, Yi~Zhang, Jyoti Aneja, Caio C{\'e}sar~Teodoro Mendes, Allie Del~Giorno, Sivakanth Gopi, Mojan Javaheripi, Piero Kauffmann, Gustavo de~Rosa, Olli Saarikivi, et~al.
\newblock Textbooks are all you need.
\newblock \emph{arXiv preprint arXiv:2306.11644}, 2023.

\bibitem[Li et~al.(2023)Li, Bubeck, Eldan, Del~Giorno, Gunasekar, and Lee]{li2023textbooks-phi1.5}
Yuanzhi Li, S{\'e}bastien Bubeck, Ronen Eldan, Allie Del~Giorno, Suriya Gunasekar, and Yin~Tat Lee.
\newblock Textbooks are all you need ii: phi-1.5 technical report.
\newblock \emph{arXiv preprint arXiv:2309.05463}, 2023.

\bibitem[Ben~Allal et~al.(2024)Ben~Allal, Lozhkov, Penedo, Wolf, and von Werra]{benallal2024cosmopedia}
Loubna Ben~Allal, Anton Lozhkov, Guilherme Penedo, Thomas Wolf, and Leandro von Werra.
\newblock Cosmopedia, February 2024.
\newblock URL \url{https://huggingface.co/datasets/HuggingFaceTB/cosmopedia}.

\bibitem[Maini et~al.(2024)Maini, Seto, Bai, Grangier, Zhang, and Jaitly]{maini2024rephrasing}
Pratyush Maini, Skyler Seto, He~Bai, David Grangier, Yizhe Zhang, and Navdeep Jaitly.
\newblock Rephrasing the web: A recipe for compute and data-efficient language modeling.
\newblock \emph{arXiv preprint arXiv:2401.16380}, 2024.

\bibitem[Liu et~al.(2024{\natexlab{a}})Liu, Wei, Liu, Si, Zhang, Rao, Zheng, Peng, Yang, Zhou, et~al.]{liu2024best}
Ruibo Liu, Jerry Wei, Fangyu Liu, Chenglei Si, Yanzhe Zhang, Jinmeng Rao, Steven Zheng, Daiyi Peng, Diyi Yang, Denny Zhou, et~al.
\newblock Best practices and lessons learned on synthetic data for language models.
\newblock \emph{arXiv preprint arXiv:2404.07503}, 2024{\natexlab{a}}.

\bibitem[Raffel et~al.(2020)Raffel, Shazeer, Roberts, Lee, Narang, Matena, Zhou, Li, and Liu]{raffel2020exploring}
Colin Raffel, Noam Shazeer, Adam Roberts, Katherine Lee, Sharan Narang, Michael Matena, Yanqi Zhou, Wei Li, and Peter~J Liu.
\newblock Exploring the limits of transfer learning with a unified text-to-text transformer.
\newblock \emph{Journal of machine learning research}, 21\penalty0 (140):\penalty0 1--67, 2020.

\bibitem[Li et~al.(2024)Li, Fang, Smyrnis, Ivgi, Jordan, Gadre, Bansal, Guha, Keh, Arora, et~al.]{li2024datacomp}
Jeffrey Li, Alex Fang, Georgios Smyrnis, Maor Ivgi, Matt Jordan, Samir Gadre, Hritik Bansal, Etash Guha, Sedrick Keh, Kushal Arora, et~al.
\newblock Datacomp-lm: In search of the next generation of training sets for language models.
\newblock \emph{arXiv preprint arXiv:2406.11794}, 2024.

\bibitem[Paster et~al.(2024)Paster, Santos, Azerbayev, and Ba]{paster2023openwebmath}
Keiran Paster, Marco~Dos Santos, Zhangir Azerbayev, and Jimmy Ba.
\newblock Openwebmath: An open dataset of high-quality mathematical web text.
\newblock In \emph{The Twelfth International Conference on Learning Representations}, 2024.
\newblock URL \url{https://openreview.net/forum?id=jKHmjlpViu}.

\bibitem[Zhuo et~al.(2024)Zhuo, Vu, Chim, Hu, Yu, Widyasari, Yusuf, Zhan, He, Paul, et~al.]{zhuo2024bigcodebench}
Terry~Yue Zhuo, Minh~Chien Vu, Jenny Chim, Han Hu, Wenhao Yu, Ratnadira Widyasari, Imam Nur~Bani Yusuf, Haolan Zhan, Junda He, Indraneil Paul, et~al.
\newblock Bigcodebench: Benchmarking code generation with diverse function calls and complex instructions.
\newblock \emph{arXiv preprint arXiv:2406.15877}, 2024.

\bibitem[Yuan et~al.(2024)Yuan, Pang, Cho, Sukhbaatar, Xu, and Weston]{yuan2024self}
Weizhe Yuan, Richard~Yuanzhe Pang, Kyunghyun Cho, Sainbayar Sukhbaatar, Jing Xu, and Jason Weston.
\newblock Self-rewarding language models.
\newblock \emph{arXiv preprint arXiv:2401.10020}, 2024.

\bibitem[Touvron et~al.(2023{\natexlab{a}})Touvron, Martin, Stone, Albert, Almahairi, Babaei, Bashlykov, Batra, Bhargava, Bhosale, et~al.]{touvron2023llama}
Hugo Touvron, Louis Martin, Kevin Stone, Peter Albert, Amjad Almahairi, Yasmine Babaei, Nikolay Bashlykov, Soumya Batra, Prajjwal Bhargava, Shruti Bhosale, et~al.
\newblock Llama 2: Open foundation and fine-tuned chat models.
\newblock \emph{arXiv preprint arXiv:2307.09288}, 2023{\natexlab{a}}.

\bibitem[Biderman et~al.(2023)Biderman, Schoelkopf, Anthony, Bradley, O’Brien, Hallahan, Khan, Purohit, Prashanth, Raff, et~al.]{biderman2023pythia}
Stella Biderman, Hailey Schoelkopf, Quentin~Gregory Anthony, Herbie Bradley, Kyle O’Brien, Eric Hallahan, Mohammad~Aflah Khan, Shivanshu Purohit, USVSN~Sai Prashanth, Edward Raff, et~al.
\newblock Pythia: A suite for analyzing large language models across training and scaling.
\newblock In \emph{International Conference on Machine Learning}, pages 2397--2430. PMLR, 2023.

\bibitem[Zhang et~al.(2024{\natexlab{b}})Zhang, Zeng, Wang, and Lu]{zhang2024tinyllama}
Peiyuan Zhang, Guangtao Zeng, Tianduo Wang, and Wei Lu.
\newblock Tinyllama: An open-source small language model.
\newblock \emph{arXiv preprint arXiv:2401.02385}, 2024{\natexlab{b}}.

\bibitem[Rozi{\`{e}}re et~al.(2023)Rozi{\`{e}}re, Gehring, Gloeckle, Sootla, Gat, Tan, Adi, Liu, Remez, Rapin, Kozhevnikov, Evtimov, Bitton, Bhatt, Canton{-}Ferrer, Grattafiori, Xiong, D{\'{e}}fossez, Copet, Azhar, Touvron, Martin, Usunier, Scialom, and Synnaeve]{DBLP:journals/corr/abs-2308-12950-codellama}
Baptiste Rozi{\`{e}}re, Jonas Gehring, Fabian Gloeckle, Sten Sootla, Itai Gat, Xiaoqing~Ellen Tan, Yossi Adi, Jingyu Liu, Tal Remez, J{\'{e}}r{\'{e}}my Rapin, Artyom Kozhevnikov, Ivan Evtimov, Joanna Bitton, Manish Bhatt, Cristian Canton{-}Ferrer, Aaron Grattafiori, Wenhan Xiong, Alexandre D{\'{e}}fossez, Jade Copet, Faisal Azhar, Hugo Touvron, Louis Martin, Nicolas Usunier, Thomas Scialom, and Gabriel Synnaeve.
\newblock Code llama: Open foundation models for code.
\newblock \emph{CoRR}, abs/2308.12950, 2023.
\newblock \doi{10.48550/ARXIV.2308.12950}.
\newblock URL \url{https://doi.org/10.48550/arXiv.2308.12950}.

\bibitem[Jiang et~al.(2023)Jiang, Sablayrolles, Mensch, Bamford, Chaplot, Casas, Bressand, Lengyel, Lample, Saulnier, et~al.]{jiang2023mistral}
Albert~Q Jiang, Alexandre Sablayrolles, Arthur Mensch, Chris Bamford, Devendra~Singh Chaplot, Diego de~las Casas, Florian Bressand, Gianna Lengyel, Guillaume Lample, Lucile Saulnier, et~al.
\newblock Mistral 7b.
\newblock \emph{arXiv preprint arXiv:2310.06825}, 2023.

\bibitem[Engstrom et~al.(2024)Engstrom, Feldmann, and Madry]{engstrom2024dsdm}
Logan Engstrom, Axel Feldmann, and Aleksander Madry.
\newblock Dsdm: Model-aware dataset selection with datamodels.
\newblock \emph{arXiv preprint arXiv:2401.12926}, 2024.

\bibitem[Xia et~al.(2024)Xia, Gao, Zeng, and Chen]{xiasheared}
Mengzhou Xia, Tianyu Gao, Zhiyuan Zeng, and Danqi Chen.
\newblock Sheared llama: Accelerating language model pre-training via structured pruning.
\newblock In \emph{The Twelfth International Conference on Learning Representations}, 2024.

\bibitem[Cheng et~al.(2024)Cheng, Gu, Huang, Bi, Huang, and Wei]{cheng2024instruction}
Daixuan Cheng, Yuxian Gu, Shaohan Huang, Junyu Bi, Minlie Huang, and Furu Wei.
\newblock Instruction pre-training: Language models are supervised multitask learners.
\newblock \emph{arXiv preprint arXiv:2406.14491}, 2024.

\bibitem[Lin et~al.(2024)Lin, Gou, Gong, Liu, Shen, Xu, Lin, Yang, Jiao, Duan, et~al.]{lin2024rho}
Zhenghao Lin, Zhibin Gou, Yeyun Gong, Xiao Liu, Yelong Shen, Ruochen Xu, Chen Lin, Yujiu Yang, Jian Jiao, Nan Duan, et~al.
\newblock Rho-1: Not all tokens are what you need.
\newblock \emph{arXiv preprint arXiv:2404.07965}, 2024.

\bibitem[Ying et~al.(2024)Ying, Zhang, Li, Zhou, Shao, Fei, Ma, Hong, Liu, Wang, et~al.]{ying2024internlm-math}
Huaiyuan Ying, Shuo Zhang, Linyang Li, Zhejian Zhou, Yunfan Shao, Zhaoye Fei, Yichuan Ma, Jiawei Hong, Kuikun Liu, Ziyi Wang, et~al.
\newblock Internlm-math: Open math large language models toward verifiable reasoning.
\newblock \emph{arXiv preprint arXiv:2402.06332}, 2024.

\bibitem[Azerbayev et~al.(2024)Azerbayev, Schoelkopf, Paster, Santos, McAleer, Jiang, Deng, Biderman, and Welleck]{azerbayev2024llemma}
Zhangir Azerbayev, Hailey Schoelkopf, Keiran Paster, Marco~Dos Santos, Stephen~Marcus McAleer, Albert~Q. Jiang, Jia Deng, Stella Biderman, and Sean Welleck.
\newblock Llemma: An open language model for mathematics.
\newblock In \emph{The Twelfth International Conference on Learning Representations}, 2024.
\newblock URL \url{https://openreview.net/forum?id=4WnqRR915j}.

\bibitem[Shao et~al.(2024)Shao, Wang, Zhu, Xu, Song, Zhang, Li, Wu, and Guo]{shao2024deepseekmath}
Zhihong Shao, Peiyi Wang, Qihao Zhu, Runxin Xu, Junxiao Song, Mingchuan Zhang, YK~Li, Yu~Wu, and Daya Guo.
\newblock Deepseekmath: Pushing the limits of mathematical reasoning in open language models.
\newblock \emph{arXiv preprint arXiv:2402.03300}, 2024.

\bibitem[Fourrier et~al.(2023)Fourrier, Habib, Wolf, and Tunstall]{lighteval}
Clémentine Fourrier, Nathan Habib, Thomas Wolf, and Lewis Tunstall.
\newblock Lighteval: A lightweight framework for llm evaluation, 2023.
\newblock URL \url{https://github.com/huggingface/lighteval}.

\bibitem[Biderman et~al.(2024)Biderman, Schoelkopf, Sutawika, Gao, Tow, Abbasi, Aji, Ammanamanchi, Black, Clive, DiPofi, Etxaniz, Fattori, Forde, Foster, Jaiswal, Lee, Li, Lovering, Muennighoff, Pavlick, Phang, Skowron, Tan, Tang, Wang, Winata, Yvon, and Zou]{biderman2024lessons}
Stella Biderman, Hailey Schoelkopf, Lintang Sutawika, Leo Gao, Jonathan Tow, Baber Abbasi, Alham~Fikri Aji, Pawan~Sasanka Ammanamanchi, Sidney Black, Jordan Clive, Anthony DiPofi, Julen Etxaniz, Benjamin Fattori, Jessica~Zosa Forde, Charles Foster, Mimansa Jaiswal, Wilson~Y. Lee, Haonan Li, Charles Lovering, Niklas Muennighoff, Ellie Pavlick, Jason Phang, Aviya Skowron, Samson Tan, Xiangru Tang, Kevin~A. Wang, Genta~Indra Winata, François Yvon, and Andy Zou.
\newblock Lessons from the trenches on reproducible evaluation of language models, 2024.

\bibitem[Team(2023)]{team2023internlm}
InternLM Team.
\newblock Internlm: A multilingual language model with progressively enhanced capabilities, 2023.

\bibitem[Touvron et~al.(2023{\natexlab{b}})Touvron, Lavril, Izacard, Martinet, Lachaux, Lacroix, Rozi{\`e}re, Goyal, Hambro, Azhar, et~al.]{touvron2023llama1}
Hugo Touvron, Thibaut Lavril, Gautier Izacard, Xavier Martinet, Marie-Anne Lachaux, Timoth{\'e}e Lacroix, Baptiste Rozi{\`e}re, Naman Goyal, Eric Hambro, Faisal Azhar, et~al.
\newblock Llama: Open and efficient foundation language models.
\newblock \emph{arXiv preprint arXiv:2302.13971}, 2023{\natexlab{b}}.

\bibitem[Smith et~al.(2022)Smith, Patwary, Norick, LeGresley, Rajbhandari, Casper, Liu, Prabhumoye, Zerveas, Korthikanti, et~al.]{smith2022using}
Shaden Smith, Mostofa Patwary, Brandon Norick, Patrick LeGresley, Samyam Rajbhandari, Jared Casper, Zhun Liu, Shrimai Prabhumoye, George Zerveas, Vijay Korthikanti, et~al.
\newblock Using deepspeed and megatron to train megatron-turing nlg 530b, a large-scale generative language model.
\newblock \emph{arXiv preprint arXiv:2201.11990}, 2022.

\bibitem[Dou et~al.(2024)Dou, Liu, Zeng, Guo, Zhou, Lu, and Lin]{dou2024sailor}
Longxu Dou, Qian Liu, Guangtao Zeng, Jia Guo, Jiahui Zhou, Wei Lu, and Min Lin.
\newblock Sailor: Open language models for south-east asia.
\newblock \emph{CoRR}, abs/2404.03608, 2024.
\newblock \doi{10.48550/ARXIV.2404.03608}.
\newblock URL \url{https://doi.org/10.48550/arXiv.2404.03608}.

\bibitem[Qiu et~al.(2024)Qiu, Lv, Jin, Wang, Ning, Yu, Zhang, Chu, Qu, Peng, et~al.]{qiu2024wanjuan}
Jiantao Qiu, Haijun Lv, Zhenjiang Jin, Rui Wang, Wenchang Ning, Jia Yu, ChaoBin Zhang, Pei Chu, Yuan Qu, Runyu Peng, et~al.
\newblock Wanjuan-cc: A safe and high-quality open-sourced english webtext dataset.
\newblock \emph{arXiv preprint arXiv:2402.19282}, 2024.

\bibitem[Broder(1997)]{broder1997resemblance}
Andrei~Z Broder.
\newblock On the resemblance and containment of documents.
\newblock In \emph{Proceedings. Compression and Complexity of SEQUENCES 1997 (Cat. No. 97TB100171)}, pages 21--29. IEEE, 1997.

\bibitem[Penedo et~al.(2024{\natexlab{b}})Penedo, Malartic, Hesslow, Cojocaru, Alobeidli, Cappelli, Pannier, Almazrouei, and Launay]{penedo2023refinedweb}
Guilherme Penedo, Quentin Malartic, Daniel Hesslow, Ruxandra Cojocaru, Hamza Alobeidli, Alessandro Cappelli, Baptiste Pannier, Ebtesam Almazrouei, and Julien Launay.
\newblock The refinedweb dataset for falcon llm: Outperforming curated corpora with web data only.
\newblock \emph{Advances in Neural Information Processing Systems}, 36, 2024{\natexlab{b}}.

\bibitem[Liu et~al.(2024{\natexlab{b}})Liu, Zeng, He, Jiang, and He]{liu2024what}
Wei Liu, Weihao Zeng, Keqing He, Yong Jiang, and Junxian He.
\newblock What makes good data for alignment? a comprehensive study of automatic data selection in instruction tuning.
\newblock In \emph{The Twelfth International Conference on Learning Representations}, 2024{\natexlab{b}}.
\newblock URL \url{https://openreview.net/forum?id=BTKAeLqLMw}.

\bibitem[Ankner et~al.(2024)Ankner, Blakeney, Sreenivasan, Marion, Leavitt, and Paul]{ankner2024perplexed}
Zachary Ankner, Cody Blakeney, Kartik Sreenivasan, Max Marion, Matthew~L Leavitt, and Mansheej Paul.
\newblock Perplexed by perplexity: Perplexity-based pruning with small reference models.
\newblock In \emph{ICLR 2024 Workshop on Navigating and Addressing Data Problems for Foundation Models}, 2024.

\bibitem[Sachdeva et~al.(2024)Sachdeva, Coleman, Kang, Ni, Hong, Chi, Caverlee, McAuley, and Cheng]{sachdeva2024train}
Noveen Sachdeva, Benjamin Coleman, Wang-Cheng Kang, Jianmo Ni, Lichan Hong, Ed~H Chi, James Caverlee, Julian McAuley, and Derek~Zhiyuan Cheng.
\newblock How to train data-efficient llms.
\newblock \emph{arXiv preprint arXiv:2402.09668}, 2024.

\bibitem[Liu et~al.(2024{\natexlab{c}})Liu, Zheng, Muennighoff, Zeng, Dou, Pang, Jiang, and Lin]{liu2024regmix}
Qian Liu, Xiaosen Zheng, Niklas Muennighoff, Guangtao Zeng, Longxu Dou, Tianyu Pang, Jing Jiang, and Min Lin.
\newblock Regmix: Data mixture as regression for language model pre-training.
\newblock \emph{CoRR}, abs/2407.01492, 2024{\natexlab{c}}.
\newblock \doi{10.48550/ARXIV.2407.01492}.
\newblock URL \url{https://doi.org/10.48550/arXiv.2407.01492}.

\bibitem[Soboleva et~al.(2023)Soboleva, Al-Khateeb, Myers, Steeves, Hestness, and Dey]{cerebras2023slimpajama}
Daria Soboleva, Faisal Al-Khateeb, Robert Myers, Jacob~R Steeves, Joel Hestness, and Nolan Dey.
\newblock {SlimPajama: A 627B token cleaned and deduplicated version of RedPajama}, June 2023.
\newblock URL \url{https://huggingface.co/datasets/cerebras/SlimPajama-627B}.

\bibitem[Fan et~al.(2024)Fan, Li, Zou, Li, He, Chern, Hu, and Liu]{fan2024reformatted}
Run-Ze Fan, Xuefeng Li, Haoyang Zou, Junlong Li, Shwai He, Ethan Chern, Jiewen Hu, and Pengfei Liu.
\newblock Reformatted alignment.
\newblock \emph{arXiv preprint arXiv:2402.12219}, 2024.

\bibitem[Yue et~al.(2024)Yue, Zheng, Zhang, and Chen]{yue2024mammoth2}
Xiang Yue, Tuney Zheng, Ge~Zhang, and Wenhu Chen.
\newblock Mammoth2: Scaling instructions from the web.
\newblock \emph{arXiv preprint arXiv:2405.03548}, 2024.

\bibitem[Hassid et~al.(2024)Hassid, Remez, Gehring, Schwartz, and Adi]{hassid2024the}
Michael Hassid, Tal Remez, Jonas Gehring, Roy Schwartz, and Yossi Adi.
\newblock The larger the better? improved {LLM} code-generation via budget reallocation.
\newblock In \emph{First Conference on Language Modeling}, 2024.
\newblock URL \url{https://openreview.net/forum?id=QJvfpWSpWm}.

\bibitem[Brown et~al.(2024)Brown, Juravsky, Ehrlich, Clark, Le, R{\'{e}}, and Mirhoseini]{DBLP:journals/corr/abs-2407-21787-large-language-monkeys-scaling-inference-compute}
Bradley C.~A. Brown, Jordan Juravsky, Ryan~Saul Ehrlich, Ronald Clark, Quoc~V. Le, Christopher R{\'{e}}, and Azalia Mirhoseini.
\newblock Large language monkeys: Scaling inference compute with repeated sampling.
\newblock \emph{CoRR}, abs/2407.21787, 2024.
\newblock \doi{10.48550/ARXIV.2407.21787}.
\newblock URL \url{https://doi.org/10.48550/arXiv.2407.21787}.

\bibitem[Snell et~al.(2024)Snell, Lee, Xu, and Kumar]{DBLP:journals/corr/abs-2408-03314-scaling-llm-test-time-compute}
Charlie Snell, Jaehoon Lee, Kelvin Xu, and Aviral Kumar.
\newblock Scaling {LLM} test-time compute optimally can be more effective than scaling model parameters.
\newblock \emph{CoRR}, abs/2408.03314, 2024.
\newblock \doi{10.48550/ARXIV.2408.03314}.
\newblock URL \url{https://doi.org/10.48550/arXiv.2408.03314}.

\bibitem[Wu et~al.(2024)Wu, Sun, Li, Welleck, and Yang]{DBLP:journals/corr/abs-2408-00724-an-empirical-analysis-compute-optimal-inference}
Yangzhen Wu, Zhiqing Sun, Shanda Li, Sean Welleck, and Yiming Yang.
\newblock An empirical analysis of compute-optimal inference for problem-solving with language models.
\newblock \emph{CoRR}, abs/2408.00724, 2024.
\newblock \doi{10.48550/ARXIV.2408.00724}.
\newblock URL \url{https://doi.org/10.48550/arXiv.2408.00724}.

\bibitem[OpenAI(2024)]{openaio1}
OpenAI.
\newblock Introducing openai o1-preview, 2024.
\newblock URL \url{https://openai.com/index/introducing-openai-o1-preview}.

\bibitem[Luukkonen et~al.(2023)Luukkonen, Komulainen, Luoma, Eskelinen, Kanerva, Kupari, Ginter, Laippala, Muennighoff, Piktus, et~al.]{luukkonen2023fingpt}
Risto Luukkonen, Ville Komulainen, Jouni Luoma, Anni Eskelinen, Jenna Kanerva, Hanna-Mari Kupari, Filip Ginter, Veronika Laippala, Niklas Muennighoff, Aleksandra Piktus, et~al.
\newblock Fingpt: Large generative models for a small language.
\newblock In \emph{Proceedings of the 2023 Conference on Empirical Methods in Natural Language Processing}, pages 2710--2726, 2023.

\bibitem[Zheng et~al.(2024)Zheng, Zhang, Zhang, Ye, and Luo]{zheng2024llamafactory}
Yaowei Zheng, Richong Zhang, Junhao Zhang, Yanhan Ye, and Zheyan Luo.
\newblock Llamafactory: Unified efficient fine-tuning of 100+ language models.
\newblock \emph{arXiv preprint arXiv:2403.13372}, 2024.

\bibitem[Penedo et~al.(2024{\natexlab{c}})Penedo, Kydlíček, Cappelli, Sasko, and Wolf]{penedo2024datatrove}
Guilherme Penedo, Hynek Kydlíček, Alessandro Cappelli, Mario Sasko, and Thomas Wolf.
\newblock Datatrove: large scale data processing, 2024{\natexlab{c}}.
\newblock URL \url{https://github.com/huggingface/datatrove}.

\bibitem[Kwon et~al.(2023)Kwon, Li, Zhuang, Sheng, Zheng, Yu, Gonzalez, Zhang, and Stoica]{kwon2023efficientvllm}
Woosuk Kwon, Zhuohan Li, Siyuan Zhuang, Ying Sheng, Lianmin Zheng, Cody~Hao Yu, Joseph~E. Gonzalez, Hao Zhang, and Ion Stoica.
\newblock Efficient memory management for large language model serving with pagedattention.
\newblock In \emph{Proceedings of the ACM SIGOPS 29th Symposium on Operating Systems Principles}, 2023.

\bibitem[AI(2023)]{litgpt-2023}
Lightning AI.
\newblock Litgpt.
\newblock \url{https://github.com/Lightning-AI/litgpt}, 2023.

\bibitem[Dao(2024)]{dao2023flashattention2}
Tri Dao.
\newblock Flash{A}ttention-2: Faster attention with better parallelism and work partitioning.
\newblock In \emph{International Conference on Learning Representations (ICLR)}, 2024.

\bibitem[Zhao et~al.(2023)Zhao, Gu, Varma, Luo, Huang, Xu, Wright, Shojanazeri, Ott, Shleifer, Desmaison, Balioglu, Damania, Nguyen, Chauhan, Hao, Mathews, and Li]{metafsdp}
Yanli Zhao, Andrew Gu, Rohan Varma, Liang Luo, Chien-Chin Huang, Min Xu, Less Wright, Hamid Shojanazeri, Myle Ott, Sam Shleifer, Alban Desmaison, Can Balioglu, Pritam Damania, Bernard Nguyen, Geeta Chauhan, Yuchen Hao, Ajit Mathews, and Shen Li.
\newblock Pytorch fsdp: Experiences on scaling fully sharded data parallel.
\newblock \emph{Proc. VLDB Endow.}, 16\penalty0 (12):\penalty0 3848–3860, aug 2023.
\newblock ISSN 2150-8097.
\newblock \doi{10.14778/3611540.3611569}.
\newblock URL \url{https://doi.org/10.14778/3611540.3611569}.

\bibitem[Hu et~al.(2024)Hu, Tu, Han, He, Cui, Long, Zheng, Fang, Huang, Zhao, et~al.]{hu2024minicpm}
Shengding Hu, Yuge Tu, Xu~Han, Chaoqun He, Ganqu Cui, Xiang Long, Zhi Zheng, Yewei Fang, Yuxiang Huang, Weilin Zhao, et~al.
\newblock Minicpm: Unveiling the potential of small language models with scalable training strategies.
\newblock \emph{arXiv preprint arXiv:2404.06395}, 2024.

\bibitem[Welbl et~al.(2017)Welbl, Liu, and Gardner]{welbl2017crowdsourcing-sciq}
Johannes Welbl, Nelson~F Liu, and Matt Gardner.
\newblock Crowdsourcing multiple choice science questions.
\newblock \emph{arXiv preprint arXiv:1707.06209}, 2017.

\bibitem[Mehta et~al.(2024)Mehta, Sekhavat, Cao, Horton, Jin, Sun, Mirzadeh, Najibi, Belenko, Zatloukal, et~al.]{mehta2024openelm}
Sachin Mehta, Mohammad~Hossein Sekhavat, Qingqing Cao, Maxwell Horton, Yanzi Jin, Chenfan Sun, Iman Mirzadeh, Mahyar Najibi, Dmitry Belenko, Peter Zatloukal, et~al.
\newblock Openelm: An efficient language model family with open-source training and inference framework.
\newblock \emph{arXiv preprint arXiv:2404.14619}, 2024.

\bibitem[Clark et~al.(2018)Clark, Cowhey, Etzioni, Khot, Sabharwal, Schoenick, and Tafjord]{clark2018arc}
Peter Clark, Isaac Cowhey, Oren Etzioni, Tushar Khot, Ashish Sabharwal, Carissa Schoenick, and Oyvind Tafjord.
\newblock Think you have solved question answering? try arc, the ai2 reasoning challenge.
\newblock \emph{arXiv preprint arXiv:1803.05457}, 2018.

\bibitem[Talmor et~al.(2019)Talmor, Herzig, Lourie, and Berant]{talmor-etal-2019-commonsenseqa}
Alon Talmor, Jonathan Herzig, Nicholas Lourie, and Jonathan Berant.
\newblock {C}ommonsense{QA}: A question answering challenge targeting commonsense knowledge.
\newblock In Jill Burstein, Christy Doran, and Thamar Solorio, editors, \emph{Proceedings of the 2019 Conference of the North {A}merican Chapter of the Association for Computational Linguistics: Human Language Technologies, Volume 1 (Long and Short Papers)}, pages 4149--4158, Minneapolis, Minnesota, June 2019. Association for Computational Linguistics.
\newblock \doi{10.18653/v1/N19-1421}.
\newblock URL \url{https://aclanthology.org/N19-1421}.

\bibitem[Zellers et~al.(2019)Zellers, Holtzman, Bisk, Farhadi, and Choi]{zellers2019hellaswag}
Rowan Zellers, Ari Holtzman, Yonatan Bisk, Ali Farhadi, and Yejin Choi.
\newblock Hellaswag: Can a machine really finish your sentence?
\newblock \emph{arXiv preprint arXiv:1905.07830}, 2019.

\bibitem[Hendrycks et~al.(2021)Hendrycks, Burns, Kadavath, Arora, Basart, Tang, Song, and Steinhardt]{hendrycks2021measuring}
Dan Hendrycks, Collin Burns, Saurav Kadavath, Akul Arora, Steven Basart, Eric Tang, Dawn Song, and Jacob Steinhardt.
\newblock Measuring mathematical problem solving with the math dataset.
\newblock In \emph{Thirty-fifth Conference on Neural Information Processing Systems Datasets and Benchmarks Track (Round 2)}, 2021.

\bibitem[Mihaylov et~al.(2018)Mihaylov, Clark, Khot, and Sabharwal]{mihaylov-etal-2018-suit-openbookqa}
Todor Mihaylov, Peter Clark, Tushar Khot, and Ashish Sabharwal.
\newblock Can a suit of armor conduct electricity? a new dataset for open book question answering.
\newblock In Ellen Riloff, David Chiang, Julia Hockenmaier, and Jun{'}ichi Tsujii, editors, \emph{Proceedings of the 2018 Conference on Empirical Methods in Natural Language Processing}, pages 2381--2391, Brussels, Belgium, October-November 2018. Association for Computational Linguistics.
\newblock \doi{10.18653/v1/D18-1260}.
\newblock URL \url{https://aclanthology.org/D18-1260}.

\bibitem[Bisk et~al.(2020)Bisk, Zellers, Gao, Choi, et~al.]{bisk2020piqa}
Yonatan Bisk, Rowan Zellers, Jianfeng Gao, Yejin Choi, et~al.
\newblock Piqa: Reasoning about physical commonsense in natural language.
\newblock In \emph{Proceedings of the AAAI conference on artificial intelligence}, volume~34, pages 7432--7439, 2020.

\bibitem[Sap et~al.(2019)Sap, Rashkin, Chen, LeBras, and Choi]{sap2019socialiqa}
Maarten Sap, Hannah Rashkin, Derek Chen, Ronan LeBras, and Yejin Choi.
\newblock Socialiqa: Commonsense reasoning about social interactions.
\newblock \emph{arXiv preprint arXiv:1904.09728}, 2019.

\bibitem[Sakaguchi et~al.(2021)Sakaguchi, Bras, Bhagavatula, and Choi]{sakaguchi2021winogrande}
Keisuke Sakaguchi, Ronan~Le Bras, Chandra Bhagavatula, and Yejin Choi.
\newblock Winogrande: An adversarial winograd schema challenge at scale.
\newblock \emph{Communications of the ACM}, 64\penalty0 (9):\penalty0 99--106, 2021.

\bibitem[Liu et~al.(2020)Liu, Cui, Liu, Huang, Wang, and Zhang]{liu2020logiqa}
Jian Liu, Leyang Cui, Hanmeng Liu, Dandan Huang, Yile Wang, and Yue Zhang.
\newblock Logiqa: A challenge dataset for machine reading comprehension with logical reasoning.
\newblock \emph{arXiv preprint arXiv:2007.08124}, 2020.

\bibitem[Clark et~al.(2019)Clark, Lee, Chang, Kwiatkowski, Collins, and Toutanova]{clark2019boolq}
Christopher Clark, Kenton Lee, Ming-Wei Chang, Tom Kwiatkowski, Michael Collins, and Kristina Toutanova.
\newblock Boolq: Exploring the surprising difficulty of natural yes/no questions.
\newblock In \emph{Proceedings of the 2019 Conference of the North American Chapter of the Association for Computational Linguistics: Human Language Technologies, Volume 1 (Long and Short Papers)}, pages 2924--2936, 2019.

\bibitem[Cobbe et~al.(2021)Cobbe, Kosaraju, Bavarian, Chen, Jun, Kaiser, Plappert, Tworek, Hilton, Nakano, et~al.]{cobbe2021gsm8k}
Karl Cobbe, Vineet Kosaraju, Mohammad Bavarian, Mark Chen, Heewoo Jun, Lukasz Kaiser, Matthias Plappert, Jerry Tworek, Jacob Hilton, Reiichiro Nakano, et~al.
\newblock Training verifiers to solve math word problems.
\newblock \emph{arXiv preprint arXiv:2110.14168}, 2021.

\bibitem[Patel et~al.(2021)Patel, Bhattamishra, and Goyal]{patel2021nlp}
Arkil Patel, Satwik Bhattamishra, and Navin Goyal.
\newblock Are nlp models really able to solve simple math word problems?
\newblock In \emph{Proceedings of the 2021 Conference of the North American Chapter of the Association for Computational Linguistics: Human Language Technologies}, pages 2080--2094, 2021.

\bibitem[Miao et~al.(2020)Miao, Liang, and Su]{miao2021diverse}
Shen-Yun Miao, Chao-Chun Liang, and Keh-Yih Su.
\newblock A diverse corpus for evaluating and developing english math word problem solvers.
\newblock In \emph{Proceedings of the 58th Annual Meeting of the Association for Computational Linguistics}, pages 975--984, 2020.

\bibitem[Koncel-Kedziorski et~al.(2016)Koncel-Kedziorski, Roy, Amini, Kushman, and Hajishirzi]{koncel2016mawps}
Rik Koncel-Kedziorski, Subhro Roy, Aida Amini, Nate Kushman, and Hannaneh Hajishirzi.
\newblock Mawps: A math word problem repository.
\newblock In \emph{Proceedings of the 2016 conference of the north american chapter of the association for computational linguistics: human language technologies}, pages 1152--1157, 2016.

\bibitem[Amini et~al.(2019)Amini, Gabriel, Lin, Koncel-Kedziorski, Choi, and Hajishirzi]{amini2019mathqa}
Aida Amini, Saadia Gabriel, Shanchuan Lin, Rik Koncel-Kedziorski, Yejin Choi, and Hannaneh Hajishirzi.
\newblock Mathqa: Towards interpretable math word problem solving with operation-based formalisms.
\newblock In \emph{Proceedings of the 2019 Conference of the North American Chapter of the Association for Computational Linguistics: Human Language Technologies, Volume 1 (Long and Short Papers)}, pages 2357--2367, 2019.

\bibitem[Lu et~al.(2023)Lu, Qiu, Chang, Wu, Zhu, Rajpurohit, Clark, and Kalyan]{lu2023tabmwp}
Pan Lu, Liang Qiu, Kai-Wei Chang, Ying~Nian Wu, Song-Chun Zhu, Tanmay Rajpurohit, Peter Clark, and Ashwin Kalyan.
\newblock Dynamic prompt learning via policy gradient for semi-structured mathematical reasoning.
\newblock In \emph{International Conference on Learning Representations (ICLR)}, 2023.

\bibitem[Kaplan et~al.(2020)Kaplan, McCandlish, Henighan, Brown, Chess, Child, Gray, Radford, Wu, and Amodei]{kaplan2020scaling}
Jared Kaplan, Sam McCandlish, Tom Henighan, Tom~B Brown, Benjamin Chess, Rewon Child, Scott Gray, Alec Radford, Jeffrey Wu, and Dario Amodei.
\newblock Scaling laws for neural language models.
\newblock \emph{arXiv preprint arXiv:2001.08361}, 2020.

\end{thebibliography}
